\documentclass{article}

\PassOptionsToPackage{numbers, compress}{natbib}
\usepackage[main, final]{neurips_2026}

\usepackage[T1]{fontenc}

\usepackage{amsmath}
\usepackage{amsfonts}
\usepackage{nicefrac}

\usepackage{graphicx}

\usepackage{booktabs}
\usepackage{longtable}
\usepackage{tabularx}
\usepackage{multirow}
\usepackage[table]{xcolor}
\usepackage{caption}
\usepackage{subcaption}

\usepackage{microtype}
\usepackage{xspace}

\definecolor{citecolor}{HTML}{0071BC}
\definecolor{linkcolor}{HTML}{ED1C24}
\definecolor{myorange}{RGB}{255,127,0}
\definecolor{myblue}{RGB}{0,114,189}
\definecolor{myred}{RGB}{205,51,51}

\usepackage[
  pagebackref=true,
  breaklinks=true,
  colorlinks,
  bookmarks=false,
  citecolor=citecolor,
  linkcolor=linkcolor,
  urlcolor=gray
]{hyperref}
\usepackage{cleveref}

\title{Learning Generative Dynamics for 3D Molecule Generation via Sequential Neuro-Symbolic Constraints}

\author{%
  Zhuoran Li, Xu Sun, Chang Wen Chen, Wanyu Lin\thanks{Corresponding author.} \\[0.8ex]
  Department of Computing \\ [0.3ex]
  The Hong Kong Polytechnic University,
  Hong Kong SAR, China \\ [0.5ex]
  \texttt{zhuor5.li@connect.polyu.hk, dsai-xu.sun@connect.polyu.hk,} \\
  \texttt{changwen.chen@polyu.edu.hk, wan-yu.lin@polyu.edu.hk} \\
}

\begin{document}

\maketitle

\begin{abstract}

Diffusion models have recently emerged as a powerful paradigm for generating 3D molecules. Despite showing remarkable promise, they learn atomic distributions implicitly from data, without explicitly encoding physicochemical rules into their generative dynamics. Consequently, generated samples can be statistically plausible yet chemically invalid. To address this problem, we propose Sequential Neuro-Symbolic Constrained Diffusion (SensDiff), a framework that integrates symbolic rules into generative modeling, enabling the model to internalize scientific priors during training. Central to SensDiff is the observation that the diffusion trajectory of 3D molecule generation exhibits a coarse-to-fine evolution: valid molecules arise by first mitigating macroscopic geometric conflicts and then refining structures toward microscopic chemical consistency through iterative denoising. Accordingly, SensDiff mirrors this dynamic by sequentially enforcing constraints with adaptive weighting during training, progressively steering generation toward valid molecular geometries. Moreover, SensDiff translates domain knowledge into interpretable generative controls, grounding opaque neural denoising in scientific principles. Experiments on molecular benchmarks and real-world constrained generation tasks confirm that SensDiff consistently improves chemical validity while adhering to physicochemical rules and user-specified targets. The code is publicly available at \url{https://github.com/WanyuGroup/NeurIPS2026_SensDiff}.

\end{abstract}

\section{Introduction}
\label{sec:introduction}

Molecule generation is a central task across various scientific domains, such as drug discovery, materials engineering, and protein design~\cite{zhung20243d, zeni2025generative, watson2023novo}. Since molecules naturally exist in three-dimensional space and their properties are largely governed by spatial configuration, it is essential to capture accurate molecular geometry in generative modeling~\cite{zhang2023learning, zhang2023resgen, uae-3d}. Despite considerable progress, producing 3D structures that are both geometrically realistic and chemically valid remains a critical challenge due to the high-dimensional, constraint-rich nature of molecular space~\cite{geoldm, midi}.

Deep generative models, particularly diffusion models, have been adapted to tackle the challenge of 3D molecule generation~\cite{edm, gcdm}. Specifically, they gradually denoise atomic coordinates and types, enabling stable modeling of molecular distributions and delivering encouraging empirical results~\cite{unigem, next-mol, sldm}. However, learning diffusion models solely from observed data tends to implicitly approximate physicochemical rules rather than internalize them within the generative process, inevitably yielding molecules that conform to the learned distribution but violate fundamental scientific principles.

Motivated by these limitations, recent studies have attempted to equip diffusion models with scientific priors for 3D molecule generation. Among them, equivariant architectures~\cite{edm, EQGAT-diff, end, geoldm} enforce spatial symmetry by construction but acquire chemical feasibility only implicitly from data. Joint discrete-continuous methods~\cite{midi, moldiff, jodo, mudiff} co-denoise atomic features and coordinates in a unified process, yet their learned coupling offers no formal physicochemical assurance. Inference-time guidance~\cite{mudm, eegsde, gcdm, tacs} injects external corrective signals during sampling without incorporating them into the training objective, and thus is not absorbed into the model parameters. Latent-space approaches~\cite{geoldm, uae-3d, ldm-3dg} lower generative complexity via dimensionality reduction, but at the cost of structural fidelity. While these efforts have advanced the field, none systematically embeds scientific priors into the training-time generative dynamics of diffusion models. Consequently, constraint compliance remains an empirical outcome rather than a built-in guarantee.

To this end, we propose SensDiff, a diffusion-based 3D molecule generator that improves the physicochemical plausibility of generated molecules. SensDiff encodes domain knowledge as explicit symbolic constraints, enabling the model to directly learn specified scientific principles rather than inferring them through data fitting. Concretely, we cast symbolic rules governing 3D molecule generation as differentiable energy functions and incorporate them into the training objective. These energy constraints are activated sequentially in a coarse-to-fine curriculum: steric non-overlap first, then bond consistency, and finally valence validity. Such ordering mirrors the iterative structural refinement inherent in diffusion-based 3D molecule generation. In addition, we design a stage-wise weighting strategy to gradually enforce constraints and stabilize training. Specifically, it first linearly ramps up the weight of each newly activated constraint and then scales it in proportion to the degree of constraint satisfaction. We further extend this framework to the sampling stage, enforcing symbolic constraints during the reverse process of a pretrained diffusion model to steer generation toward chemically valid and geometrically realistic molecules without retraining.

To verify the effectiveness of SensDiff, we perform a comprehensive suite of evaluations, including (1) standard 3D molecule generation benchmarks; (2) constrained multi-objective molecule generation (\textit{e.g.,} optimizing quantum-chemical and drug-likeness properties, with up to seven properties satisfied simultaneously); (3) out-of-distribution generalization of the learned scientific rules; (4) extension of the framework to sampling and fine-tuning; (5) transfer of the training strategy to various diffusion backbones; and (6) analysis of failure cases from a physicochemical perspective to guide model improvements. Crucially, we conduct extensive ablations on constraint enforcement mechanisms, weighting strategies, and training- and sampling-time schedules (\textit{e.g.,} constraint activation curricula and epoch schedules), empirically validating each design choice. Our main contributions include: 

\begin{itemize}
    \item We introduce SensDiff, a neuro-symbolic constrained diffusion model for 3D molecule generation that produces molecules with enhanced geometric and physicochemical validity.
    \item We design a training strategy that mirrors the coarse-to-fine nature of diffusion molecular generation, enabling the model to internalize domain knowledge in its generative dynamics.
    \item We evaluate SensDiff through broad experiments, showing improved generation quality and constraint compliance, while ablations and failure case analysis validate each design choice.
\end{itemize}

\section{Preliminaries}
\label{sec:Preliminaries}

This section overviews the preliminaries, with details in Appendix~\ref{sec:detailed_preliminaries} and notations in Appendix~\ref{sec:math_notations}.

\textbf{Molecular Diffusion Models and Symbolic State.}
We adopt GEOLDM~\cite{geoldm} as the diffusion backbone. A molecule with coordinates \(x\) and atom features \(h\) is encoded into a latent state \(z=[z^{(x)} \| z^{(h)}]\), where \(z^{(x)}\) and \(z^{(h)}\) are the coordinate and feature channels. The clean latent state is \(z_0=z_{xh}=[z_{x\mu}\|z_{h\mu}]\). Diffusion defines a corrupted latent \(z_t=\alpha_t z_0+\sigma_t\epsilon\), where \(\epsilon \sim \mathcal N(0,I)\), and the reverse process is parameterized by \(\epsilon_\theta(z_t,t)\). Let \(z_s\) denote the previous reverse-time latent state corresponding to \(z_t\). From \(z_s\), a soft decoder produces decoded coordinates \(R=\{r_i\}_{i=1}^N\) and soft atom-type distributions \(A_i\), while a bond posterior head predicts \(B_{ij}^{(m)}\), where \(m\) denotes the bond type. The symbolic state \((R,A,B)\) is the interface for evaluating and enforcing scientific constraints.

\textbf{Symbolic Constraints.}
SensDiff maps physicochemical constraints over \((R,A,B)\) to differentiable energies via a soft truth value \(\kappa \in (0,1)\) that quantifies constraint satisfaction, with energy \(E=-(1/Z)\sum \omega \log(\kappa+\varepsilon)\), where $\omega$ is a soft antecedent (\textit{i.e.}, precondition satisfaction) weight, \(Z\) is a normalization factor, and \(\varepsilon\) is a numerical stability constant. This yields four energies, \(E_{\text{clash}}, \text{ } E_{\text{bond\_len}}, \text{ } E_{\text{angle}},\) and \(E_{\text{val}}\), corresponding to steric non-overlap, bond-length consistency, bond-angle consistency, and valence validity. During training, they are combined as \(\mathcal L_{\text{symbolic}}(e)=\lambda_c(e)\cdot E_{\text{clash}}+\lambda_b(e)\cdot E_{\text{bond\_len}}+\lambda_a(e)\cdot E_{\text{angle}}+\lambda_v(e)\cdot E_{\text{val}}\), where \(e\) is the epoch and \(\lambda_c(e), \text{ }\lambda_b(e), \text{ }\lambda_a(e), \text{ and }\lambda_v(e)\) are epoch-dependent constraint weights in the diffusion model’s sequential training curriculum, designed to stabilize training when new constraints are introduced.

\section{Method}
\label{sec:method}

Neuro-symbolic learning combines the expressiveness of neural networks with the interpretability of symbolic reasoning~\cite{hitzler2022neuro, feldstein2024mapping, hitzler2022neuro_a}. In this work, symbolic rules encode domain knowledge as differentiable energy constraints whose gradients provide supervisory signals to the denoising network, penalizing chemically invalid intermediate structures during training. We further propose a coarse-to-fine training strategy that sequentially activates the constraints with adaptive weighting. This aligns with the model’s learning progression over training epochs, shaping the generative dynamics with physicochemical principles. As a result, these scientific rules are embedded in the model parameters and can generalize beyond finite training samples, rather than being implicitly memorized from data.

\subsection{Differentiable Molecular Interface}
A key challenge for SensDiff is that physicochemical rules are defined in 3D molecular space, whereas our diffusion evolves in a continuous latent space. To bridge the gap, we construct a differentiable interface that maps intermediate denoising states to 3D molecules. Let \(z=[z^{(x)} \| z^{(h)}] \in \mathbb{R}^{B\times N\times D}\) denote the latent state, where \(z^{(x)}\) and \(z^{(h)}\) represent the coordinate and feature channels, and \(B\), \(N\), and \(D\) denote the batch size, padded number of atoms, and latent dimensionality, respectively. Given a reverse-time state \(z_s\), we decode it into atomic coordinates and soft atom identities
\[
(R,A)=\mathrm{\texttt{decode\_soft}}(z_s),
\qquad
A_i = m_i \cdot \operatorname{\texttt{softmax}}(h_i^{\mathrm{logit}}),
\]
where \(R=\{r_i\}_{i=1}^{N}\) denotes the decoded coordinates, \(A_i\) is the atom-type distribution for atom \(i\), and \(m_i \in \{0,1\}\) is the node mask. For valid atoms with \(m_i=1\), we have \(A_i \in \Delta^{C-1}\), where \(C\) is the number of atom types. For padded atoms with \(m_i=0\), the mask forces \(A_i\) to the zero vector. This soft decoding path is crucial: if atom identities are fixed via a hard \(\arg\max\), the symbolic component becomes constant and yields zero gradients, so symbolic constraints cannot guide the denoising process. Using simplex-valued atom-type probabilities instead preserves uncertainty in atomic identity and keeps the latent-to-symbolic mapping entirely differentiable. Nevertheless, coordinates and soft atom identities alone cannot express bond or valence constraints. We therefore introduce a bond posterior head that infers a soft molecular graph from the decoded state, producing bond posteriors \(B_{ij}^{(m)}\). For each atomic pair \((i,j)\), we form the pair feature \(\phi_{ij} = \left[\left| r_i - r_j \right|_{2},\, A_i,\, A_j\right]\), map it to bond logits \(\ell_{ij}\), symmetrize via \(\tilde{\ell}_{ij}=\tfrac{1}{2}(\ell_{ij}+\ell_{ji})\), and define the bond posteriors as
\[
B_{ij}^{(m)}=
\frac{\exp(\tilde{\ell}_{ij}^{(m)})}
{\sum_{m'=0}^{M-1}\exp(\tilde{\ell}_{ij}^{(m')})},
\]
where \(m \in \{0,\dots,M-1\}\) indexes bond classes, and \(M\) is the number of bond classes. This yields a probabilistic and  differentiable relational representation of molecular bonds. Based on this, the symbolic state \((R,A,B)\) produces the quantities needed to enforce steric, bonding, and valence constraints on intermediate denoising states. To link the symbolic state to empirical chemistry, we precompute statistics on the training data and interpolate them under soft atom identities. The non-bonded distance lower bound for atom types \(a\) and \(b\) is given by
\begin{equation}
d_{\min,ij}=\sum_{a=1}^{C}\sum_{b=1}^{C}A_{ia}A_{jb}d_{\min}^{ab}.
\label{eq:dmin_interp}
\end{equation}
The bond-length statistics (mean \(\mu\) and standard deviation \(\sigma\)) are computed as
\begin{equation}
\mu_{ijm}^{\text{bond}}=\sum_{a,b}A_{ia}A_{jb}\mu_{abm}^{\text{bond}},
\qquad
\sigma_{ijm}^{\text{bond}}=
\max\!\left(\sigma_{\min}^{\text{bond}},\sum_{a,b}A_{ia}A_{jb}\sigma_{abm}^{\text{bond}}\right),
\label{eq:bond_stats_interp}
\end{equation}
and the center-conditioned bond-angle statistics in cosine space are computed as
\begin{equation}
\mu_{j,mn}^{\cos}=\sum_{b=1}^{C}A_{jb}\mu_{b,mn}^{\cos,\mathrm{fb}},
\qquad
\sigma_{j,mn}^{\cos}=
\max\!\left(\sigma_{\min}^{\text{angle}},\sum_{b=1}^{C}A_{jb}\sigma_{b,mn}^{\cos,\mathrm{fb}}\right),
\label{eq:angle_stats_interp}
\end{equation}
where \(\sigma_{\min}^{\text{angle}}=\sigma_{\min}^{\text{bond}}=0.03\) are lower bounds on the interpolated angle and bond-length standard deviations, used to stabilize standardized residuals. \(\mathrm{fb}\) denotes a fallback version of the angle cosine statistics, indexed only by the central atom type and adjacent bond classes, to alleviate sparsity in the fully triplet-conditioned table. Accordingly, these interpolants turn discrete chemistry statistics into differentiable functions of \(z_s\), allowing symbolic constraints to act directly on latent denoising states.

\subsection{Physicochemical Rules as Symbolic Energy Constraints}

The second step is to cast physicochemical rules as differentiable energies. The intuition here is to assign low energy to molecules that satisfy the rule and high energy to those that violate it. Let \(\kappa \in (0,1)\) denote a soft truth value measuring the degree to which a rule is satisfied. We map it to
\begin{equation}
E = -\frac{1}{Z}\sum \omega \log(\kappa+\varepsilon),
\label{eq:energy_def}
\end{equation}
where \(\omega\) is a soft antecedent (\textit{i.e.,} precondition satisfaction) weight, \(Z\) is a normalization factor, and \(\varepsilon=10^{-8}\) ensures numerical stability. This assigns lower energy to strongly satisfied rules and larger penalties for more severe violations. We use the Pseudo-Huber penalty \(\rho(u)=\sqrt{1+u^2}-1\), which is quadratic near zero and linear in the tails~\cite{hartley2003multiple}, to avoid unstable gradients when noisy intermediate states deviate from chemically valid structures. We conduct extensive experiments (see Appendix~\ref{appendix_ablation}) and analyze failure cases from a physicochemical perspective (see Appendix~\ref{fail_case_appendix}) in our preliminary study to determine which constraints to impose during training and in what order to apply them. 

\textbf{Steric Non-Overlap.}
The first rule encodes geometric chemistry, requiring non-bonded atom pairs to avoid spatial overlap. This is a physically fundamental rule, so it becomes reliable earliest during denoising. For pair \((i,j)\), let \(d_{ij}=\|r_i-r_j\|_2\). We define the steric soft truth as
\[
\kappa_{ij}^{\text{clash}}=\sigma\!\left(\frac{d_{ij}-d_{\min,ij}}{\tau_c}\right),
\qquad \tau_c=0.05,
\]
where \(\sigma(\cdot)\) is the sigmoid function. This quantity is close to \(1\) when the distance exceeds the interpolated lower bound and approaches \(0\) as the atoms clash. Since this rule only applies to pairs believed to be non-bonded, its antecedent is the no-bond posterior \(B_{ij}^{(0)}\), yielding energy as
\begin{equation}
E_{\text{clash}}=
-\frac{1}{Z_c}\sum_{i<j}B_{ij}^{(0)}\log(\kappa_{ij}^{\text{clash}}+\varepsilon),
\qquad
Z_c=\sum_{i<j}B_{ij}^{(0)}+\varepsilon.
\label{eq:clash_energy}
\end{equation}
\textbf{Bond Length.}
Bond length captures 1-hop geometry and enforces consistency among distance, atom type, and bond order. For each nonzero bond class \(m \neq 0\), we define the standardized deviation
\[
z_{ijm}^{\text{bond}}=
\frac{d_{ij}-\mu_{ijm}^{\text{bond}}}{\sigma_{ijm}^{\text{bond}}},
\qquad
\kappa_{ijm}^{\text{bond}}=\exp\!\bigl(-\rho(z_{ijm}^{\text{bond}})\bigr),
\]
and the corresponding energy for bond length can be expressed as
\begin{equation}
E_{\text{bond\_len}}=
-\frac{1}{Z_b}\sum_{i<j}\sum_{m\neq 0}B_{ij}^{(m)}\log(\kappa_{ijm}^{\text{bond}}+\varepsilon),
\qquad
Z_b=\sum_{i<j}\sum_{m\neq 0}B_{ij}^{(m)}+\varepsilon.
\label{eq:bond_len_energy}
\end{equation}
\textbf{Bond Angle.}
Bond angle captures 2-hop local coordination geometry and therefore becomes meaningful later than bond length. Directly working with the angle itself would require \(\arccos\), whose derivative becomes unstable near \(0^\circ\) and \(180^\circ\). We instead work in cosine space
\[
c_{ijk}=
\frac{\langle r_i-r_j,\ r_k-r_j\rangle}
{\|r_i-r_j\|_2\,\|r_k-r_j\|_2+\varepsilon},
\]
where \(j\) denotes the center atom, and \(i\) and \(k\) are candidate neighbors of \(j\). Since enumerating all atom triplets scales as $\mathcal{O}(N^3)$, we restrict evaluation to the top-\(K_{\mathrm{nbr}}\) likely neighbors of each center atom, selected according to \(
p_{ji}^{\text{bond}}=\sum_{m\neq 0}B_{ji}^{(m)}.
\)
Let \(\mathcal N_{K_{\mathrm{nbr}}}(j)\) be the corresponding top-\(K_{\mathrm{nbr}}\) set, with \(K_{\mathrm{nbr}}=4\). For \(i,k \in \mathcal N_{K_{\mathrm{nbr}}}(j)\) and \(i<k\), and for nonzero bond classes \(m\) and \(n\), we have
\[
z_{ijkmn}^{\text{angle}}=
\frac{c_{ijk}-\mu_{j,mn}^{\cos}}{\sigma_{j,mn}^{\cos}},
\qquad
\kappa_{ijkmn}^{\text{angle}}=\exp\!\bigl(-\rho(z_{ijkmn}^{\text{angle}})\bigr),
\]
and the corresponding energy for bond angle can be expressed as
\begin{equation}
E_{\text{angle}}=
\frac{
\sum_{j}\sum_{\substack{i<k\\ i,k\in\mathcal N_{K_{\mathrm{nbr}}}(j)}}\sum_{m\neq 0}\sum_{n\neq 0}
B_{ji}^{(m)}B_{jk}^{(n)}\bigl[-\log(\kappa_{ijkmn}^{\text{angle}}+\varepsilon)\bigr]
}{
\sum_{j}\sum_{\substack{i<k\\ i,k\in\mathcal N_{K_{\mathrm{nbr}}}(j)}}\sum_{m\neq 0}\sum_{n\neq 0}
B_{ji}^{(m)}B_{jk}^{(n)}+\varepsilon
}.
\label{eq:angle_energy}
\end{equation}
The product \(B_{ji}^{(m)}B_{jk}^{(n)}\) serves as a two-edge antecedent, so the angle rule is activated only when both incident (neighboring) bonds are plausible. Together, \(E_{\text{bond\_len}}\) and \(E_{\text{angle}}\) enforce bond consistency.

\textbf{Valence Validity.}
The final rule enforces valence validity. Let \(\nu_m\) denote the bond-order weight of bond class \(m\), with \(\nu_0=0\), \(\nu_1=1\), \(\nu_2=2\), \(\nu_3=3\), and \(\nu_{\mathrm{aroma}}=1.5\). The soft valence of atom \(i\) is
\[
v_i^{\text{soft}}=\sum_{j\neq i}\sum_{m=0}^{M-1}\nu_m B_{ij}^{(m)}.
\]
This quantity is the expected total incident bond order under the soft bond posterior. To determine whether it is chemically valid, we compare it against the admissible valence set \(\mathcal V(a)\) for each atom type \(a\), weighted by the empirical prior \(\pi_{a,v}\), and define the valence soft truth as
\[
\kappa_i^{\text{val}}=
\sum_{a=1}^{C}A_{ia}\sum_{v\in\mathcal V(a)}
\pi_{a,v}\exp\!\left(-\frac{(v_i^{\text{soft}}-v)^2}{2\sigma_v^2}\right),
\qquad
\sigma_v=0.15.
\]
Notably, when an atom type permits multiple valid valences, \(\kappa_i^{\text{val}}\) remains high near any admissible value. The corresponding energy for valence validity can be expressed as
\begin{equation}
E_{\text{val}}=
-\frac{1}{N_{\text{valid}}}\sum_i m_i\log(\kappa_i^{\text{val}}+\varepsilon),
\qquad
N_{\text{valid}}=\sum_i m_i.
\label{eq:val_energy}
\end{equation}

\subsection{Coarse-to-Fine Curriculum with Adaptive Loss Weighting}

SensDiff aligns symbolic guidance with the evolution of diffusion-based molecule generation. As structure emerges, bond length becomes meaningful once local connectivity is plausible, and bond angle follows as a 2-hop signal that requires two credible incident bonds. Valence is most downstream, since it aggregates bond order over a neighborhood and amplifies uncertainty in the bond posterior.

Our training pipeline consists of one preparation stage followed by four curriculum stages. Stage~0 precomputes dataset chemistry statistics, trains the variational autoencoder, and pretrains the bond-posterior head. It then freezes both so that subsequent symbolic gradients update only the diffusion model. Stage~1 activates \(E_{\text{clash}}\), Stage~2 adds \(E_{\text{bond\_len}}\), Stage~3 adds \(E_{\text{angle}}\), and Stage~4 adds \(E_{\text{val}}\). This schedule is not merely heuristic, as it matches when intermediate denoising states become reliable enough for each constraint to provide meaningful supervision. To mitigate optimization instability when adding a new constraint, we assign each constraint an adaptive weight \(\lambda_k(e)\) at epoch \(e\). Before activation, \(\lambda_k(e)=0\). After activation at epoch \(e_k^{\text{start}}\), the weight is linearly warmed up over \(R_{\text{ramp}}\) epochs and then further increased based on the current level of rule satisfaction. Formally,
\begin{equation}
\lambda_k(e)=
\begin{cases}
0, & e<e_k^{\text{start}},\\[2mm]
\dfrac{\lambda_k^{\text{final}}}{2}
\min\!\left(1,\dfrac{e-e_k^{\text{start}}}{R_{\text{ramp}}}\right),
& 0\le e-e_k^{\text{start}}<R_{\text{ramp}},\\[4mm]
\dfrac{\lambda_k^{\text{final}}}{2}\Bigl(1+g(\bar E_k)\Bigr),
& e-e_k^{\text{start}}\ge R_{\text{ramp}},
\end{cases}
\label{eq:lambda_schedule}
\end{equation}
where \(e_k^{\text{start}}\) is the activation epoch of constraint \(k\), \(R_{\text{ramp}}\) is the warmup length, \(\lambda_k^{\text{final}}\) is the final weight, and \(\bar E_k\) is the rolling mean of \(E_k\) for smoothing short-term fluctuations. \(\bar E_k^{(0)}\) is the reference energy at the start of the adaptive phase, and \(\tau\) is the threshold for releasing the full weight. Besides,
\[
g(\bar E_k)=
\min\!\left(
1,\
\frac{\max(0,\bar E_k^{(0)}-\bar E_k)}
{\bar E_k^{(0)}-\tau}
\right)
\]
is the energy-adaptive factor. The linear warmup lets the model adapt to the gradients of a newly introduced rule, while the adaptive phase increases the weight only once the rule becomes reliable at the current noise level. Thus, the curriculum synchronizes symbolic supervision with the stability of intermediate denoising states, rather than applying a fixed sequence of auxiliary losses.

\begin{table*}[t]
\caption{Unconditional 3D molecule generation on QM9~\cite{qm9} and GEOM-Drugs~\cite{geom-drug}. \textbf{Best} results are marked in bold. We provide a complete comparison of chemical validity and geometric plausibility on QM9 and GEOM-Drugs in Table~\ref{tab:unconditional_qm9_full} and Table~\ref{tab:unconditional_gd_full}, respectively. All metrics are percentages (\%).}
\label{tab:unconditional_generation}
{\setlength{\heavyrulewidth}{0.03em}
\setlength{\lightrulewidth}{0.03em}
\setlength{\cmidrulewidth}{0.03em}
\setlength{\arrayrulewidth}{0.03em}
\resizebox{1.0\textwidth}{!}{
\centering
\tiny
\renewcommand{\arraystretch}{1.08}
\begin{tabular}{lcccc|cc}
\toprule
\multirow{1}{*}{\raisebox{-2.5ex}{\textsc{{\scriptsize Model}}}} &
\multicolumn{4}{c}{\textsc{QM9}} & \multicolumn{2}{c}{\textsc{GEOM-Drugs}} \\
\cmidrule(lr){2-5}\cmidrule(lr){6-7}
& \textsc{Valid} & \textsc{Atom Sta} & \textsc{Mol Sta} & \textsc{Valid \& Unique} &
\textsc{Valid} & \textsc{Atom Sta} \\
\midrule
EDM~\cite{edm} & 87.9 & 95.5 & 78.4 & 96.0 & 88.8 & 77.5 \\
EEGSDE~\cite{eegsde} & 89.8 & 95.7 & 80.3 & 94.3 & 89.1 & 79.1 \\
GEOLDM~\cite{geoldm} & 89.8 & 94.2 & 86.2 & 95.5 & 95.6 & 80.7 \\
EquiFM~\cite{equifm} & 91.6 & 94.4 & 84.3 & 94.7 & 95.7 & 80.3 \\
GeoBFN~\cite{geobfn} & 90.7 & 94.9 & 86.9 & 93.3 & 88.5 & 81.5 \\
GCDM~\cite{gcdm} & 91.4 & 94.2 & 82.5 & 95.7 & 91.2 & 84.5 \\
RADM~\cite{radm} & 90.0 & 95.9 & 84.7 & 92.2 & 95.7 & 82.1 \\
SLDM~\cite{sldm} & 92.1 & 95.0 & 90.9 & 89.1 & 95.8 & 84.8 \\
UAE-3D~\cite{uae-3d} & 94.4 & 96.2 & 95.4 & 94.4 & 84.4 & 82.5 \\
UniGEM~\cite{unigem} & 92.0 & 95.2 & 87.1 & 93.5 & 94.7 & 82.1 \\
\rowcolor[RGB]{228, 232, 242} SensDiff (\textbf{Ours}) & \textbf{96.9} & \textbf{98.1} & \textbf{95.6} & \textbf{97.2} & \textbf{97.2} & \textbf{87.9} \\
\bottomrule
\end{tabular}
}}
\vspace{-0.1cm}
\end{table*}

\begin{figure*}[t]
    \centering
    \includegraphics[width=1.0\textwidth]{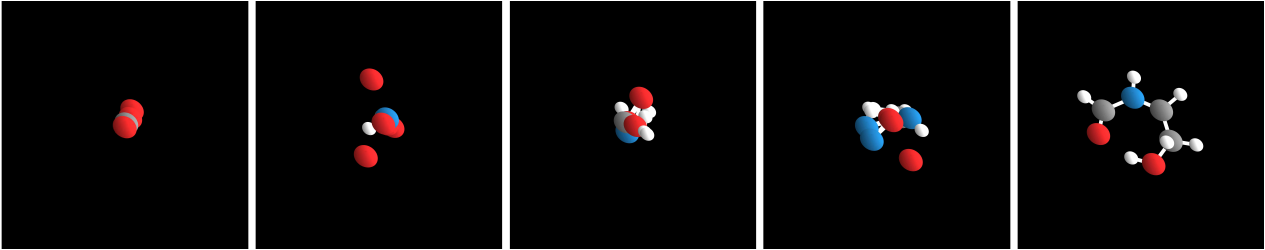}
    \caption{The 3D molecule generation trajectory over intermediate diffusion denoising steps. We also visualize trajectories from multiple baselines for the same molecule in Figures~\ref{fig:mol_tra_004_compressed}, \ref{fig:mol_tra_013_compressed}, and~\ref{fig:mol_tra_026_compressed}.}
    \label{fig:mole_trajectory}
    \vspace{-0.5cm}
\end{figure*}

\subsection{Diffusion Model Training Objective}

\vspace{-0.13cm}

SensDiff incorporates the four symbolic constraints into the training objective. Let \(\mathcal L_{\text{diffusion}}\) denote the native training objective of the diffusion backbone. The symbolic loss is
{\setlength{\abovedisplayskip}{7pt}\setlength{\belowdisplayskip}{0pt}
\begin{equation}
\mathcal L_{\text{symbolic}}(e)=
\lambda_c(e) \cdot E_{\text{clash}}+
\lambda_b(e) \cdot E_{\text{bond\_len}}+
\lambda_a(e) \cdot E_{\text{angle}}+
\lambda_v(e) \cdot E_{\text{val}},
\label{eq:symbolic_loss_final}
\end{equation}}

\noindent and the final training objective combines data fitting with the enforcement of physicochemical rules, ensuring that the model captures complex molecular patterns from data while simultaneously adhering to prescribed scientific principles, and is defined as
{\setlength{\abovedisplayskip}{6pt}\setlength{\belowdisplayskip}{0pt}
\begin{equation}
\mathcal L_{\text{total}}(e)=\mathcal L_{\text{diffusion}}(e)+\mathcal L_{\text{symbolic}}(e).
\label{eq:total_loss}
\end{equation}}

\noindent To make the constraints act on intermediate denoising states rather than only on the final decoded molecule, we evaluate \(\mathcal L_{\text{symbolic}}(e)\) on a one-step reverse surrogate $z_s$. In particular, starting from the clean mean code $z_{xh}$, we sample a random diffusion step $t_{\mathrm{int}}\sim\mathcal U\{1,
\dots,T\}$, construct the noisy latent $z_t=\alpha_t z_{xh}+\sigma_t\epsilon$
with $\epsilon \sim \mathcal N(0,I)$, and then compute one reverse step
{\setlength{\abovedisplayskip}{8pt}\setlength{\belowdisplayskip}{3pt}
\begin{equation}
z_s \sim \mathcal N\!\left(
\frac{z_t}{\alpha_{t|s}} - \frac{\sigma_{t|s}^2}{\alpha_{t|s}\sigma_t}\epsilon_\theta(z_t,t),
\ \left(\frac{\sigma_{t|s}\sigma_s}{\sigma_t}\right)^2 I
\right).
\label{eq:reverse_sampling}
\end{equation}}

\noindent Based on this, the symbolic state \((R,A,B)\) is constructed from \(z_s\), and the resulting symbolic loss is backpropagated through the reverse denoising step to the diffusion model. Because \(E_{\text{clash}}, \text{ } E_{\text{bond\_len}}, \text{ }  E_{\text{angle}},\) and \(E_{\text{val}}\) are differentiable in \((R,A,B)\), and \((R,A,B)\) is differentiable in \(z_s\), each rule yields a gradient \(\nabla_{z_s}E_k\), which then propagates through the reverse kernel to the denoising network. As a result, the model is encouraged not merely to learn the observed molecular distribution, but to organize its generative dynamics around physicochemical rules. 

In summary, SensDiff augments diffusion models with a sequential neuro-symbolic training strategy. It exposes chemically meaningful intermediate states, evaluates physicochemical rules on them, and feeds the resulting constraint signals back into the denoising process. Our design follows the coarse-to-fine behavior of diffusion molecular generation, where global geometry becomes reliable before local bonding and valence. Technically, SensDiff first establishes a differentiable interface from latent states to soft atoms and bonds. It then converts domain knowledge into differentiable energy constraints. These constraints are internalized through a sequential curriculum with epoch-dependent loss weighting. Each constraint is activated only when its required structural signal is stable. Consequently, the model learns from data while being guided by scientific priors, improving chemical validity and rule compliance. We provide implementation details for SensDiff in Appendix~\ref{sec:implementation_details}.

\section{Experiments}
\label{sec:experiments}

\begin{table*}[t]
\caption{Mean absolute error (MAE, \(\downarrow\)) for quantum property conditional generation on QM9~\cite{qm9}.}
\label{tab:quantum_qm9}
\centering
{\setlength{\heavyrulewidth}{0.03em}
\setlength{\lightrulewidth}{0.03em}
\setlength{\cmidrulewidth}{0.03em}
\setlength{\arrayrulewidth}{0.03em}
\setlength{\tabcolsep}{11pt}
\renewcommand{\arraystretch}{1.18}
\small
\resizebox{\textwidth}{!}{%
\begin{tabular}{c c c c c c c}
\toprule
\multicolumn{1}{c}{\textsc{Model}} &
$C_v$ \(\left(\frac{\text{cal}}{\text{mol}}\,\text{K}\right)\) &
$\mu$ (D) &
$\alpha$ (Bohr$^3$) &
$\Delta \varepsilon$ (meV) &
$\varepsilon_{\mathrm{HOMO}}$ (meV) &
$\varepsilon_{\mathrm{LUMO}}$ (meV) \\
\midrule
\textcolor{gray}{U-Bound} & \textcolor{gray}{6.879} & \textcolor{gray}{1.613} & \textcolor{gray}{8.98} & \textcolor{gray}{1464} & \textcolor{gray}{645} & \textcolor{gray}{1457} \\
\textcolor{gray}{L-Bound} & \textcolor{gray}{0.040} & \textcolor{gray}{0.043} & \textcolor{gray}{0.09} & \textcolor{gray}{65} & \textcolor{gray}{39} & \textcolor{gray}{36} \\
EDM~\cite{edm}       & 1.154 & 1.164 & 2.87 & 676 & 368 & 603 \\
EEGSDE~\cite{eegsde} & 0.986 & 0.812 & 2.61 & 506 & 314 & 462 \\
GEOLDM~\cite{geoldm} & 1.072 & 1.156 & 2.46 & 608 & 351 & 541 \\
EquiFM~\cite{equifm} & 1.081 & 1.148 & 2.52 & 614 & 349 & 548 \\
GeoBFN~\cite{geobfn} & 0.992 & 1.041 & 2.43 & 596 & 339 & 532 \\
GCDM~\cite{gcdm}     & 0.723 & 0.879 & 2.06 & 624 & 357 & 495 \\
RADM~\cite{radm}     & 0.907 & 0.851 & 2.07 & 474 & 301 & 397 \\
SLDM~\cite{sldm}     & 0.781 & 0.831 & 1.53 & 456 & 331 & 362 \\
UAE-3D~\cite{uae-3d} & 0.582 & 0.638 & 1.61 & 327 & 224 & 259 \\
UniGEM~\cite{unigem} & 0.754 & 0.801 & 1.84 & 441 & 295 & 349 \\
\rowcolor[RGB]{225, 230, 245} SensDiff (Ours) &
\textbf{0.569} & \textbf{0.618} & \textbf{1.49} & \textbf{314} & \textbf{219} & \textbf{246} \\
\bottomrule
\end{tabular}%
}
}
\end{table*}

\begin{figure*}[t]
    \centering
    \includegraphics[width=1.0\textwidth]{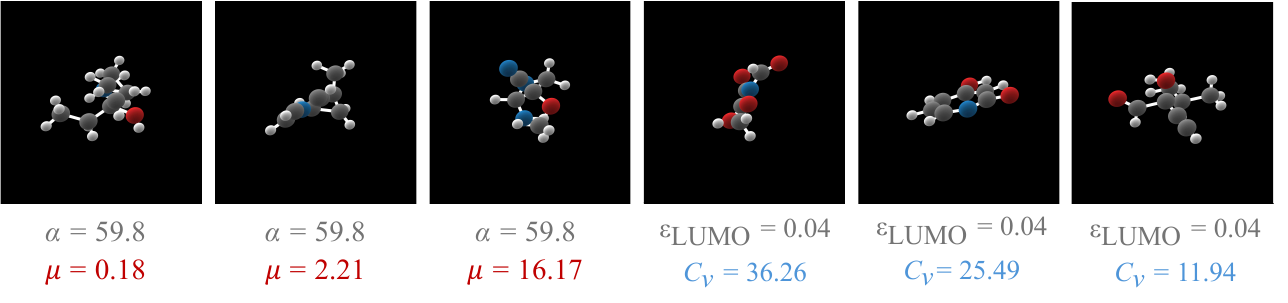}
    \caption{Conditional generation on QM9~\cite{qm9} targeting two quantum-chemical properties simultaneously. Additional results are provided in Figure~\ref{fig:single_appendix} and Figure~\ref{fig:double_appendix}.}
    \label{fig:conditional_double_prop_main}
    \vspace{-0.2cm}
\end{figure*}

We evaluate SensDiff on two standard molecular benchmarks, QM9~\cite{qm9} and GEOM-Drugs~\cite{geom-drug}. QM9 consists of small organic molecules with annotated quantum-chemical properties and is used for both unconditional and conditional generation, while GEOM-Drugs contains larger and more diverse drug-like molecules, providing a more challenging evaluation setting. We compare against a comprehensive set of recent baselines, including EDM~\cite{edm}, EEGSDE~\cite{eegsde}, GEOLDM~\cite{geoldm}, EquiFM~\cite{equifm}, GeoBFN~\cite{geobfn}, GCDM~\cite{gcdm}, RADM~\cite{radm}, SLDM~\cite{sldm}, UAE-3D~\cite{uae-3d}, and UniGEM~\cite{unigem}. All methods are evaluated under a unified pipeline using metrics that cover chemical validity, sample diversity, geometric fidelity, constraint compliance, and conditional controllability. Each experiment is repeated ten times, and the mean results are reported. Further details on the datasets, baselines, and evaluation protocols are provided in Appendix~\ref{sec:experimental_setup}.

\subsection{Unconditional 3D Molecule Generation}
Following prior work~\cite{geoldm, sldm, radm, unigem}, Table~\ref{tab:unconditional_generation} reports results on two benchmarks, with generation trajectories shown in Figure~\ref{fig:mole_trajectory}. SensDiff achieves the best generation quality across all metrics, with the largest gains in stability. It improves upon the strongest baseline by 0.2\% in molecule stability on QM9 and 3.1\% in atom stability on GEOM-Drugs. We attribute this to encoding physicochemical rules as symbolic constraints that directly supervise steric, bonding, and valence properties during training, yielding more chemically coherent structures. The coarse-to-fine curriculum further enforces basic geometric validity before enforcing finer chemical consistency, preventing higher-order constraints from being learned on flawed intermediates. More results are provided in Appendix~\ref{sec:append_unconditional_generation}.

\subsection{Conditional 3D Molecule Generation}

\begin{figure*}[t]
    \centering
    \includegraphics[width=1.0\textwidth]{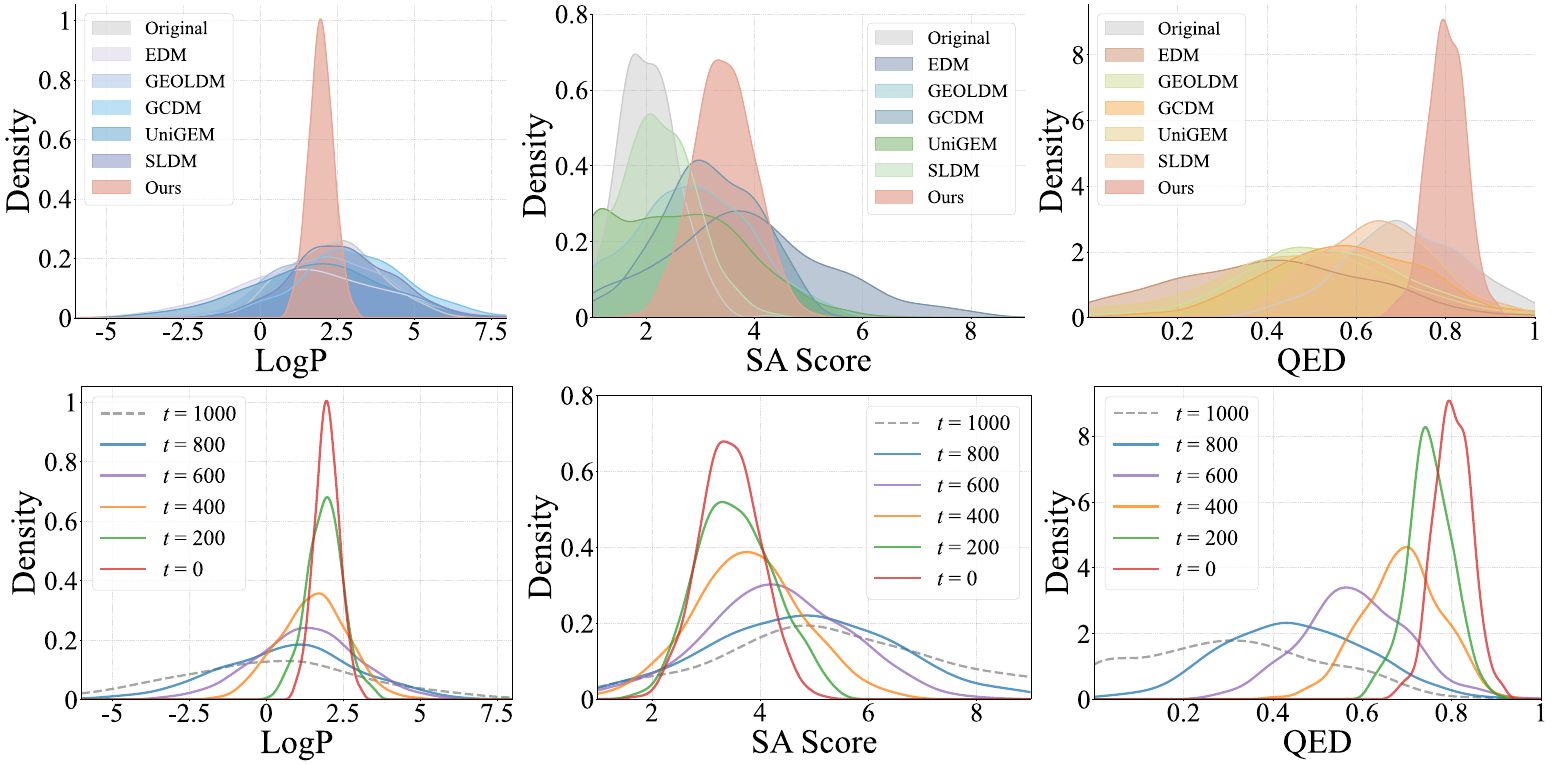}
    \caption{(\textbf{Upper}) The generated molecules precisely satisfy the specified oral drug-design constraints. (\textbf{Bottom}) The molecules converge to the target values over the denoising steps.}
    \label{fig:oral_drug_main}
    \vspace{-0.2cm}
\end{figure*}

\textbf{Targeted Generation with Quantum Properties.}
QM9 annotates molecules with diverse quantum chemical properties, enabling conditional generation to test whether models can produce molecules that match specified targets. Table~\ref{tab:quantum_qm9} reports MAE for property-conditioned generation, where lower is better. SensDiff achieves the lowest MAE across all properties, outperforming the strongest baseline UAE-3D~\cite{uae-3d}. Since these properties are largely determined by 3D geometry, our explicit geometric constraints provide a stronger foundation for conditional generation. Figure~\ref{fig:conditional_double_prop_main} further shows that SensDiff generates structurally diverse molecules that respond coherently to changing conditions, enabling fine-grained control over multiple properties. Additional results are provided in Appendix~\ref{sec:append_conditional_generation_quantum}.

\textbf{Multi-Objective Optimization for Oral Drugs.}
Oral drug design remains challenging because oral bioavailability requires meeting multiple criteria simultaneously, and violating any single criterion can invalidate a candidate~\cite{gomez2005strategies}. Existing methods often struggle to generate 3D molecules that satisfy all requirements at once because these rules are not explicitly learned. We thus apply SensDiff by freezing the GEOM-Drugs model and enforcing constraints only during sampling to preserve structural validity while steering generation toward target ranges. Specifically, we impose seven constraints aligned with Lipinski’s Rule of Five and industry standards~\cite{navia1996design}: \(1<\text{LogP}<3\), \(\text{MW}\le 500\), \(\text{TPSA}\le 140~\text{\AA}^2\), HBD \(\le 5\), HBA \(\le 10\), \(\text{QED}>0.7\), and \(\text{SA}<5\). Figure~\ref{fig:oral_drug_main} visualizes the constraint distributions and convergence toward the targets across denoising steps. More results on oral drug design and Lipinski’s Rule-of-Five--constrained drug design~\cite{lipinski2004lead, walters2012going} are given in Appendix~\ref{con_multi_drug_design}.

\subsection{Ablation Studies}

\begin{figure*}[t]
    \centering
    \includegraphics[width=1.0\textwidth]{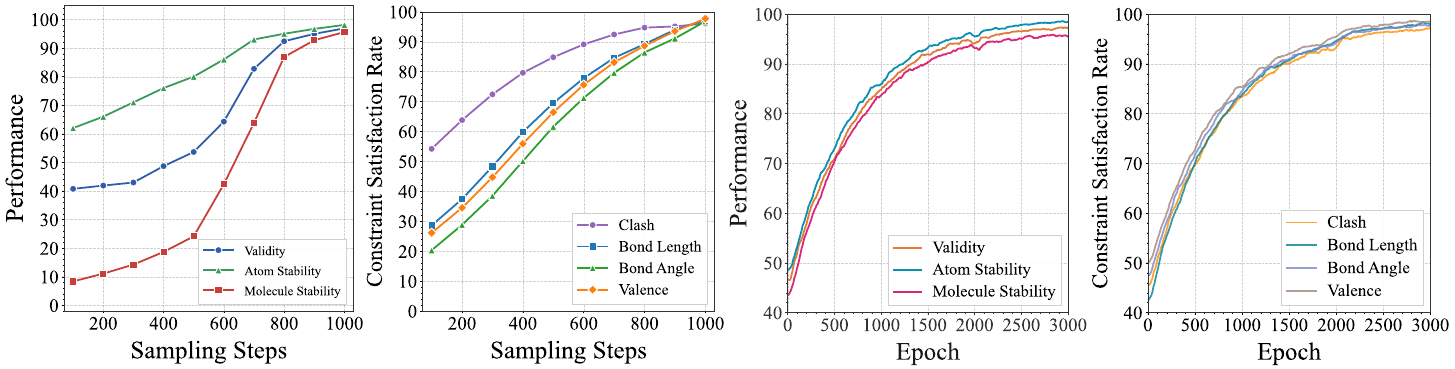}
    \caption{Ablations on sampling steps and training epochs on QM9. We save model checkpoints every 25 epochs (\textit{i.e.,} 120 checkpoints in total). Results on GEOM-Drugs are shown in Figure~\ref{fig:gd_step_epoch}.}
    \label{fig:qm9_step_epoch}
    \vspace{-0.3cm}
\end{figure*}

We conduct a comprehensive set of ablation studies to validate our training strategy. Due to space limitations, we present only the key findings here, with detailed results and analysis in Appendix~\ref{appendix_ablation}.

\textbf{Constraint Enforcement Order.} 
As shown in Table~\ref{tab:constraint_order_ablation}, we study which enforcement order most improves generation quality by sequentially activating constraint subsets. The order matters substantially, with validity ranging from 74.6\% to 90.9\% across permutations. The best ordering (our design) applies global geometry early and local valence later, aligning with the denoising trajectory. In contrast, applying fine-grained local rules too early degrades performance because early geometries are too noisy to satisfy bond-level constraints, yielding misleading gradients that hinder refinement. 

\textbf{Non-Sequential Constraint Enforcement.} 
As reported in Table~\ref{tab:ablation_set_qm9} and Table~\ref{tab:ablation_set_geom_drugs}, we study how sequential enforcement affects the molecule generation. When all constraints are enforced simultaneously, adding more constraints degrades generation quality on both datasets (\textit{e.g.,} on QM9, molecule stability drops from 88.8\% with one constraint to 78.8\% with all four), suggesting that naively combining loss terms introduces competing gradients. This motivates our sequential activation strategy, which introduces rules progressively and lets the model consolidate each prior before adding the next.

\textbf{Weighting Strategies.} 
As shown in Table~\ref{tab:ablation_weight_training} and Table~\ref{tab:ablation_weight_samling}, we investigate the optimal constraint weights for diffusion training and sampling, respectively. Moreover, Table~\ref{tab:ablation_weighting_mechanism} presents how adaptive weighting affects the molecule generation. A hierarchical weighting that prioritizes global geometry over local bonds and valency is essential: flattening or inverting the weights collapses both generation quality and constraint satisfaction, indicating that a dominant structural stabilizer is needed before finer refinements can take effect. Moreover, linear ramping stabilizes training by avoiding abrupt gradients from newly introduced constraints, while energy-adaptive weighting further improves performance by adjusting weights based on constraint satisfaction rather than a fixed schedule.

\textbf{Training and Sampling Schedules.} 
Figure~\ref{fig:qm9_step_epoch} and Figure~\ref{fig:gd_step_epoch} visualize how training epochs and sampling steps improve generation quality and constraint compliance. Table~\ref{tab:ablation_mae} and Table~\ref{tab:ablation_ns_setting} compares constraints applied during training versus sampling only, showing that training-time constraints better improve molecule generation by more thoroughly internalizing scientific priors into the model via iterative learning. Table~\ref{tab:ablation_enforcement_samling} shows that constraints are most effective when applied frequently, since sparse enforcement allows drift to accumulate and reduces both stability and constraint satisfaction.

\subsection{Key Empirical Insights}

\textbf{Observation I}. 
The neuro-symbolic method provides an interpretable lens that informs model design: by analyzing generated molecules, we can pinpoint which rules are violated and devise targeted remedies. We find that failures stem from a strict physicochemical hierarchy, where spatial non-overlap must precede meaningful bond geometry, and valid topology emerges only after local geometry stabilizes. Upstream violations thus cascade into downstream defects, motivating sequential training with adaptive weighting that activates each constraint only when its prerequisite becomes reliable. Notably, this analysis reveals that simply imposing more constraints does not necessarily improve generation, since prematurely activated rules can produce misleading or competing gradients. The key is therefore to enforce the right constraints at the stage when their underlying structural signals become sufficiently reliable. We provide a detailed failure case analysis from a domain-science perspective in Appendix~\ref{fail_case_appendix}.

\textbf{Observation II}. 
SensDiff is trained to encode scientific principles into its generative dynamics, and these learned mechanistic regularities generalize to previously unseen data. This highlights the strong generalizability of our training strategy, which endows the model with science-grounded, rule-consistent reasoning beyond black-box neural denoising. In particular, SensDiff remains effective under cross-dataset transfer and in low-density property regions, where purely data-driven models receive limited distributional support. This suggests that explicit physicochemical rules provide transferable structural inductive biases rather than merely fitting dataset-specific correlations. Details are provided in Appendix~\ref{ood}.

\textbf{Observation III}. 
Our integration of scientific priors enables SensDiff to internalize domain knowledge, improving geometric plausibility, constraint compliance, and fine-grained control (\textit{e.g.,} simultaneously satisfying even conflicting constraints). Importantly, these improvements are complementary rather than mutually exclusive: enforcing scientific rules enhances validity without sacrificing molecular diversity or conditional accuracy. Moreover, the resulting gains extend from directly constrained local structures, such as bond lengths and angles, to higher-order geometry and molecular-level properties. Details are provided in Appendices~\ref{sec:Geometric Plausibility}, \ref{more_metrics}, and~\ref{con_multi_drug_design}.

\textbf{Observation IV}. 
The proposed framework is broadly applicable: we empirically demonstrate that it can be leveraged for sampling-time constraint enforcement and fine-tuning. Moreover, our training strategy generalizes to various diffusion backbones. While sampling-time guidance offers a flexible way to impose new requirements without retraining, training-time integration yields stronger improvements by internalizing the constraints into the model parameters. The consistent gains across backbones further indicate that SensDiff acts as a model-agnostic training principle rather than an architecture-specific modification. Details are provided in Appendices~\ref{appendix_ablation}, \ref{sec:fine-tuning}, and~\ref{generalization_to_backbones}.

\section{Related Work}
\label{sec:related_work}

\textbf{3D Molecule Generation with Diffusion Models.} 
Score-based~\cite{score} and denoising diffusion models~\cite{ddpm} are dominant methods for 3D molecule generation~\cite{mdm, VoxMol, next-mol, sldm, end, uae-3d}. They learn to reverse a Gaussian noising process to recover atomic coordinates and features~\cite{edm, geoldm}, and have been widely applied to molecule generation~\cite{EDM-Bridge, ldm-3dg, unigem}, conformer generation~\cite{GeoDiff, torsional, disco}, optimization~\cite{DiffSBDD, pmdm, decompopt}, and docking~\cite{diffdock, redock, dynamicbind}. Existing approaches primarily improve generation through architectural inductive biases, molecular representations, or diffusion trajectory design. These advances effectively enhance geometric modeling and generation quality, but chemical feasibility is still learned largely through statistical correlations in the training data. However, purely data-driven training does not ensure adherence to scientific priors unless they are explicitly specified, so statistically plausible samples may still violate fundamental physicochemical rules. This limitation becomes more pronounced for complex molecular structures and generation beyond the well-represented training distribution. We thus formulate symbolic rules as differentiable energies and incorporate them into the training loss to shape the generative dynamics toward chemically valid structures.

\textbf{Constrained Molecule Generation.}
Imposing task-specific constraints on molecular generators is vital for scientific discovery~\cite{qiang2023bridging, gebauer2022inverse, chen2023shapeconditioned}. Large language models (LLMs) represent molecules as string sequences~\cite{smiles, selfies} and satisfy user-specified objectives through instruction tuning, chemical feedback loops, or representational manipulation~\cite{edwards2022translation, pei2023biot5, fang2024molinstructions, li2026controllable}. Flow-based models learn transport maps and use equivariance or conditional vector fields to steer generation toward constraint-satisfying structures~\cite{geobfn, equifm, zhou2025energybased}. Diffusion models strike a balance between training stability and generation quality, making them particularly well suited for 3D molecule generation~\cite{berrada2024improved, karras2024analyzing}. Nevertheless, many constrained generation methods introduce corrective signals only during sampling, which limits the model's ability to internalize the underlying scientific principles. Moreover, applying multiple constraints simultaneously may introduce competing guidance, especially when intermediate molecular structures remain noisy and unreliable. Building on the strengths of diffusion models, we propose a symbolic-rule-constrained training strategy that sequentially integrates physicochemical constraints into the denoising process according to their structural dependencies. Our strategy is compatible with different molecular diffusion backbones and improves physicochemical feasibility while preserving generation quality and conditional controllability.

\section{Conclusion}
\label{sec:conclusion}

In this work, we propose SensDiff, a diffusion-based 3D molecule generator for physicochemically valid molecule generation. SensDiff internalizes scientific priors via sequential, adaptively weighted symbolic constraints aligned with diffusion’s generative dynamics. Experiments show high generation quality and strong constraint compliance, while ablation studies validate our key design choices.

%%%%%%%%%%%%%%%%%%%%%%%%%%%%%%%%%%%%%%%%%%%%%%%%%%%%%%%%%%%%

\newpage
\section*{Acknowledgements and Disclosures}

This work was supported in part by the Hong Kong Polytechnic University Internal Research Fund (P0057774, Young Innovative Researcher Scheme; P0063303, RIAIoT) and by the Research Grants Council of Hong Kong through the General Research Fund (Ref. No. 15208725). The authors received no other third-party funding or support related to this work during the past 36 months and declare that they have no competing interests.

\vspace{0.8cm}
\bibliographystyle{plain}
\bibliography{reference}

\clearpage
\section*{NeurIPS Paper Checklist}

%%% BEGIN INSTRUCTIONS %%%
The checklist is designed to encourage best practices for responsible machine learning research, addressing issues of reproducibility, transparency, research ethics, and societal impact. Do not remove the checklist: {\bf The papers not including the checklist will be desk rejected.} The checklist should follow the references and follow the (optional) supplemental material.  The checklist does NOT count towards the page
limit. 

Please read the checklist guidelines carefully for information on how to answer these questions. For each question in the checklist:
\begin{itemize}
    \item You should answer \answerYes{}, \answerNo{}, or \answerNA{}.
    \item \answerNA{} means either that the question is Not Applicable for that particular paper or the relevant information is Not Available.
    \item Please provide a short (1--2 sentence) justification right after your answer (even for \answerNA). 
   % \item {\bf The papers not including the checklist will be desk rejected.}
\end{itemize}

{\bf The checklist answers are an integral part of your paper submission.} They are visible to the reviewers, area chairs, senior area chairs, and ethics reviewers. You will also be asked to include it (after eventual revisions) with the final version of your paper, and its final version will be published with the paper.

The reviewers of your paper will be asked to use the checklist as one of the factors in their evaluation. While \answerYes{} is generally preferable to \answerNo{}, it is perfectly acceptable to answer \answerNo{} provided a proper justification is given (e.g., error bars are not reported because it would be too computationally expensive'' or ``we were unable to find the license for the dataset we used''). In general, answering \answerNo{} or \answerNA{} is not grounds for rejection. While the questions are phrased in a binary way, we acknowledge that the true answer is often more nuanced, so please just use your best judgment and write a justification to elaborate. All supporting evidence can appear either in the main paper or the supplemental material, provided in appendix. If you answer \answerYes{} to a question, in the justification please point to the section(s) where related material for the question can be found.

IMPORTANT, please:
\begin{itemize}
    \item {\bf Delete this instruction block, but keep the section heading ``NeurIPS Paper Checklist"},
    \item  {\bf Keep the checklist subsection headings, questions/answers and guidelines below.}
    \item {\bf Do not modify the questions and only use the provided macros for your answers}.
\end{itemize}

%%% END INSTRUCTIONS %%%

\begin{enumerate}

\item {\bf Claims}
    \item[] Question: Do the main claims made in the abstract and introduction accurately reflect the paper's contributions and scope?
    \item[] Answer: \answerYes{} % Replace by \answerYes{}, \answerNo{}, or \answerNA{}.
    \item[] Justification: Please refer to the contributions in Section~\ref{sec:introduction}.

\item {\bf Limitations}
    \item[] Question: Does the paper discuss the limitations of the work performed by the authors?
    \item[] Answer: \answerYes{} % Replace by \answerYes{}, \answerNo{}, or \answerNA{}.
    \item[] Justification: Please refer to the limitation and future work in Section~\ref{limitation}.

\item {\bf Theory assumptions and proofs}
    \item[] Question: For each theoretical result, does the paper provide the full set of assumptions and a complete (and correct) proof?
    \item[] Answer: \answerNA{} % Replace by \answerYes{}, \answerNo{}, or \answerNA{}.
    \item[] Justification: This work is primarily empirical, with the mathematical formulation presented in Section~\ref{sec:method}.

    \item {\bf Experimental result reproducibility}
    \item[] Question: Does the paper fully disclose all the information needed to reproduce the main experimental results of the paper to the extent that it affects the main claims and/or conclusions of the paper (regardless of whether the code and data are provided or not)?
    \item[] Answer: \answerYes{} % Replace by \answerYes{}, \answerNo{}, or \answerNA{}.
    \item[] Justification: Detailed experimental setup and implementation details are provided in Sections~\ref{sec:experimental_setup} and~\ref{sec:implementation_details}, and the official code is publicly available at \url{https://github.com/WanyuGroup/NeurIPS2026_SensDiff}.

\item {\bf Open access to data and code}
    \item[] Question: Does the paper provide open access to the data and code, with sufficient instructions to faithfully reproduce the main experimental results, as described in supplemental material?
    \item[] Answer: \answerYes{} % Replace by \answerYes{}, \answerNo{}, or \answerNA{}.
    \item[] Justification:We release the code, together with detailed usage instructions, at \url{https://github.com/WanyuGroup/NeurIPS2026_SensDiff}.

\item {\bf Experimental setting/details}
    \item[] Question: Does the paper specify all the training and test details (e.g., data splits, hyperparameters, how they were chosen, type of optimizer) necessary to understand the results?
    \item[] Answer: \answerYes{} % Replace by \answerYes{}, \answerNo{}, or \answerNA{}.
    \item[] Justification: We provide detailed experimental setup and implementation details in Section~\ref{sec:experimental_setup} and \ref{sec:implementation_details}, respectively, which are necessary to understand the results.

\item {\bf Experiment statistical significance}
    \item[] Question: Does the paper report error bars suitably and correctly defined or other appropriate information about the statistical significance of the experiments?
    \item[] Answer: \answerNo{} % Replace by \answerYes{}, \answerNo{}, or \answerNA{}.
    \item[] Justification: Instead, we run each experiment ten times and report the mean.

\item {\bf Experiments compute resources}
    \item[] Question: For each experiment, does the paper provide sufficient information on the computer resources (type of compute workers, memory, time of execution) needed to reproduce the experiments?
    \item[] Answer: \answerYes{} % Replace by \answerYes{}, \answerNo{}, or \answerNA{}.
    \item[] Justification: Please refer to Section~\ref{sec:implementation_details} for compute resources.
    
\item {\bf Code of ethics}
    \item[] Question: Does the research conducted in the paper conform, in every respect, with the NeurIPS Code of Ethics \url{https://neurips.cc/public/EthicsGuidelines}?
    \item[] Answer: \answerYes{} % Replace by \answerYes{}, \answerNo{}, or \answerNA{}.
    \item[] Justification: We have read and followed the Code of Ethics.

\item {\bf Broader impacts}
    \item[] Question: Does the paper discuss both potential positive societal impacts and negative societal impacts of the work performed?
    \item[] Answer: \answerYes{} % Replace by \answerYes{}, \answerNo{}, or \answerNA{}.
    \item[] Justification: Please refer to Section~\ref{Broader Impacts}.
    
\item {\bf Safeguards}
    \item[] Question: Does the paper describe safeguards that have been put in place for responsible release of data or models that have a high risk for misuse (e.g., pre-trained language models, image generators, or scraped datasets)?
    \item[] Answer: \answerNo{} % Replace by \answerYes{}, \answerNo{}, or \answerNA{}.
    \item[] Justification: This work does not involve data or models that pose a high risk for misuse.

\item {\bf Licenses for existing assets}
    \item[] Question: Are the creators or original owners of assets (e.g., code, data, models), used in the paper, properly credited and are the license and terms of use explicitly mentioned and properly respected?
    \item[] Answer: \answerYes{} % Replace by \answerYes{}, \answerNo{}, or \answerNA{}.
    \item[] Justification: All assets used in this paper are properly credited, and their licenses and terms of use are explicitly stated and respected. Please refer to Section~\ref{sec:experiments} and Section~\ref{sec:experimental_setup}.

\item {\bf New assets}
    \item[] Question: Are new assets introduced in the paper well documented and is the documentation provided alongside the assets?
    \item[] Answer: \answerYes{} % Replace by \answerYes{}, \answerNo{}, or \answerNA{}.
    \item[] Justification: We release the code for implementing the proposed method at \url{https://github.com/WanyuGroup/NeurIPS2026_SensDiff}.

\item {\bf Crowdsourcing and research with human subjects}
    \item[] Question: For crowdsourcing experiments and research with human subjects, does the paper include the full text of instructions given to participants and screenshots, if applicable, as well as details about compensation (if any)? 
    \item[] Answer: \answerNA{} % Replace by \answerYes{}, \answerNo{}, or \answerNA{}.
    \item[] Justification: The paper does not involve crowdsourcing nor research with human subjects.

\item {\bf Institutional review board (IRB) approvals or equivalent for research with human subjects}
    \item[] Question: Does the paper describe potential risks incurred by study participants, whether such risks were disclosed to the subjects, and whether Institutional Review Board (IRB) approvals (or an equivalent approval/review based on the requirements of your country or institution) were obtained?
    \item[] Answer: \answerNA{} % Replace by \answerYes{}, \answerNo{}, or \answerNA{}.
    \item[] Justification: The paper does not involve crowdsourcing nor research with human subjects.

\item {\bf Declaration of LLM usage}
    \item[] Question: Does the paper describe the usage of LLMs if it is an important, original, or non-standard component of the core methods in this research? Note that if the LLM is used only for writing, editing, or formatting purposes and does \emph{not} impact the core methodology, scientific rigor, or originality of the research, declaration is not required.
    %this research? 
    \item[] Answer: \answerNA{} % Replace by \answerYes{}, \answerNo{}, or \answerNA{}.
    \item[] Justification: The LLM is used only for grammar checking, formatting, and writing assistance.

\end{enumerate}

\newpage
\tableofcontents

\newpage
\appendix
\newpage
\section{Notations}
\label{sec:math_notations}

We organize the notation used in our method into four groups: Table~\ref{tab:notation_diffusion} summarizes the notation for molecular representation and diffusion models, Table~\ref{tab:notation_symbolic} summarizes the notation for the differentiable symbolic state and chemistry statistics, Table~\ref{tab:notation_constraints} summarizes the notation used in the four symbolic constraints, and Table~\ref{tab:notation_training} summarizes the notation used in the training curriculum and objective.

\vspace{1\baselineskip}

\begin{table*}[h]
\centering
\caption{Notation for molecular representation and diffusion models.}
\label{tab:notation_diffusion}
\small
\setlength{\tabcolsep}{8pt}
\renewcommand{\arraystretch}{1.4}
\begin{tabular}{p{0.25\textwidth}p{0.65\textwidth}}
\toprule
\textbf{Notation} & \textbf{Meaning} \\
\midrule
\(B\) & Batch size. \\
\(N\) & Padded number of atoms per molecule. \\
\(C\) & Number of atom types. \\
\(M\) & Number of bond classes. \\
\(D\) & Latent dimensionality. \\
\(x\) & Molecular coordinates. \\
\(h\) & Molecular atom features. \\
\(h^{\mathrm{cat}}\) & Categorical atom features. \\
\(h^{\mathrm{int}}\) & Integer-valued atom features. \\
\(m_i\) & Node mask for atom slot \(i\). \\
\(i,j,k\) & Atom indices; in angle expressions, \(j\) is the center atom and \(i,k\) are its neighbors. \\
\(z\) & Latent molecular state. \\
\(z^{(x)}\) & Coordinate channel of the latent state. \\
\(z^{(h)}\) & Feature channel of the latent state. \\
\(z_{x\mu}, z_{x\sigma}\) & Encoder mean and standard deviation for the coordinate latent code. \\
\(z_{h\mu}, z_{h\sigma}\) & Encoder mean and standard deviation for the feature latent code. \\
\(z_{xh}\) & Mean latent code used as the clean latent state for diffusion. \\
\(z_0\) & Clean latent state. \\
\(z_t\) & Corrupted latent state at diffusion step \(t\). \\
\(z_s\) & Previous-step latent state in the reverse process. \\
\(z_T\) & Terminal Gaussian-noise latent state. \\
\(t\) & Normalized diffusion time. \\
\(s\) & Previous normalized diffusion time. \\
\(t_{\mathrm{int}}\) & Integer diffusion step. \\
\(T\) & Total number of diffusion steps. \\
\(\alpha_t\) & Signal coefficient in the forward diffusion process. \\
\(\sigma_t\) & Noise coefficient in the forward diffusion process. \\
\(\alpha_{t|s}\) & Conditional signal coefficient for the reverse step from \(t\) to \(s\). \\
\(\sigma_{t|s}\) & Conditional noise coefficient for the reverse step from \(t\) to \(s\). \\
\(\epsilon\) & Gaussian noise. \\
\(\theta\) & Parameters of the reverse diffusion model. \\
\(\epsilon_\theta\) & Noise prediction network. \\
\(p_\theta(z_s\mid z_t)\) & Reverse transition kernel. \\
\(\mu\) & Mean of the Gaussian reverse kernel. \\
\(\sigma\) & Standard deviation of the Gaussian reverse kernel. \\
\bottomrule
\end{tabular}
\end{table*}

\begin{table*}[h]
\centering
\caption{Notation for the differentiable symbolic state and chemistry statistics.}
\label{tab:notation_symbolic}
\small
\setlength{\tabcolsep}{9pt}
\renewcommand{\arraystretch}{1.6}
\begin{tabular}{p{0.25\textwidth}p{0.65\textwidth}}
\toprule
\textbf{Notation} & \textbf{Meaning} \\
\midrule
\(R\) & Decoded coordinates from the soft decoder. \\
\(r_i\) & Coordinate of atom \(i\). \\
\(A_i\) & Soft atom-type distribution for atom \(i\). \\
\(A_{ia}\) & Probability that atom \(i\) has atom type \(a\). \\
\(h_i^{\mathrm{logit}}\) & Pre-softmax atom-type logits for atom \(i\). \\
\(\Delta^{C-1}\) & Probability simplex for valid atom-type distributions. \\
\(B_{ij}^{(m)}\) & Bond posterior for atom pair \((i,j)\) and bond class \(m\). \\
\(m\) & Bond-class index. \\
\(n\) & A second bond-class index used in the angle rule. \\
\(v\) & Valence value used in the admissible valence set. \\
\(\phi_{ij}\) & Pair feature used by the bond posterior head. \\
\(\ell_{ij}\) & Bond logits for pair \((i,j)\). \\
\(\tilde{\ell}_{ij}\) & Symmetrized bond logits for pair \((i,j)\). \\
\(a,b\) & Atom-type indices used in chemistry statistics. \\
\(d_{\min}^{ab}\) & Non-bonded distance lower bound for atom types \(a\) and \(b\). \\
\(d_{\min,ij}\) & Softly interpolated non-bonded distance lower bound for pair \((i,j)\). \\
\(\mu_{abm}^{\text{bond}}\) & Bond-length mean conditioned on atom types \((a,b)\) and bond class \(m\). \\
\(\sigma_{abm}^{\text{bond}}\) & Bond-length standard deviation conditioned on atom types \((a,b)\) and bond class \(m\). \\
\(\mu_{ijm}^{\text{bond}}\) & Interpolated bond-length mean for atom pair \((i,j)\). \\
\(\sigma_{ijm}^{\text{bond}}\) & Interpolated bond-length standard deviation for atom pair \((i,j)\). \\
\(\mu_{b,mn}^{\cos,\mathrm{fb}}\) & Fallback angle-cosine mean conditioned on center atom type \(b\) and bond classes \((m,n)\). \\
\(\sigma_{b,mn}^{\cos,\mathrm{fb}}\) & Fallback angle-cosine standard deviation conditioned on center atom type \(b\) and bond classes \((m,n)\). \\
\(\mu_{j,mn}^{\cos}\) & Interpolated angle-cosine mean for center atom \(j\). \\
\(\sigma_{j,mn}^{\cos}\) & Interpolated angle-cosine standard deviation for center atom \(j\). \\
\(\sigma_{\min}^{\text{bond}}\) & Lower bound for interpolated bond-length standard deviation. \\
\(\sigma_{\min}^{\text{angle}}\) & Lower bound for interpolated angle standard deviation. \\
\(\nu_m\) & Bond-order weight of bond class \(m\). \\
\(\mathcal V(a)\) & Admissible valence set for atom type \(a\). \\
\(\pi_{a,v}\) & Empirical prior for valence \(v\) of atom type \(a\). \\
\(v_i^{\text{soft}}\) & Soft valence of atom \(i\). \\
\bottomrule
\end{tabular}
\end{table*}

\begin{table*}[t]
\centering
\caption{Notation used in the four symbolic constraints.}
\label{tab:notation_constraints}
\footnotesize
\setlength{\tabcolsep}{5pt}
\renewcommand{\arraystretch}{1.22}
\begin{tabular}{p{0.25\textwidth}p{0.65\textwidth}}
\toprule
\textbf{Notation} & \textbf{Meaning} \\
\midrule
\(E\) & Generic differentiable energy. \\
\(\kappa\) & Soft truth value of a symbolic constraint. \\
\(\omega\) & Soft antecedent weight. \\
\(Z\) & Normalization factor in the generic energy definition. \\
\(\varepsilon\) & Numerical-stability constant. \\
\(\sigma(\cdot)\) & Sigmoid function. \\
\(\rho\) & Pseudo-Huber penalty. \\
\(d_{ij}\) & Euclidean distance between atoms \(i\) and \(j\). \\
\(\|\cdot\|_2\) & Euclidean norm. \\
\(\langle \cdot,\cdot \rangle\) & Inner product. \\
\(\kappa_{ij}^{\text{clash}}\) & Soft truth for the steric non-overlap rule. \\
\(\tau_c\) & Temperature parameter for the clash rule. \\
\(E_{\text{clash}}\) & Steric-clash energy. \\
\(Z_c\) & Normalization factor for the clash energy. \\
\(z_{ijm}^{\text{bond}}\) & Standardized bond-length residual. \\
\(\kappa_{ijm}^{\text{bond}}\) & Soft truth for the bond-length rule. \\
\(E_{\text{bond\_len}}\) & Bond-length energy. \\
\(Z_b\) & Normalization factor for the bond-length energy. \\
\(c_{ijk}\) & Cosine of the angle formed by atoms \(i\)-\(j\)-\(k\). \\
\(p_{ji}^{\text{bond}}\) & Total nonzero bond probability between atoms \(j\) and \(i\). \\
\(K_{\mathrm{nbr}}\) & Number of likely neighbors retained for bond-angle evaluation. \\
\(\mathcal N_{K_{\mathrm{nbr}}}(j)\) & Top-\(K_{\mathrm{nbr}}\) likely neighbors of center atom \(j\). \\
\(z_{ijkmn}^{\text{angle}}\) & Standardized angle residual. \\
\(\kappa_{ijkmn}^{\text{angle}}\) & Soft truth for the bond-angle rule. \\
\(E_{\text{angle}}\) & Bond-angle energy. \\
\(\kappa_i^{\text{val}}\) & Soft truth for the valence-validity rule. \\
\(\sigma_v\) & Bandwidth parameter in the valence kernel. \\
\(N_{\text{valid}}\) & Number of valid atoms in the molecule or batch item. \\
\(E_{\text{val}}\) & Valence energy. \\
\bottomrule
\end{tabular}
\end{table*}

\begin{table*}[t]
\centering
\caption{Notation for the training curriculum and objective.}
\label{tab:notation_training}
\footnotesize
\setlength{\tabcolsep}{5pt}
\renewcommand{\arraystretch}{1.22}
\begin{tabular}{p{0.25\textwidth}p{0.65\textwidth}}
\toprule
\textbf{Notation} & \textbf{Meaning} \\
\midrule
\(k\) & Index of a symbolic constraint in the curriculum. \\
\(e\) & Training epoch. \\
\(\lambda_k(e)\) & Epoch-dependent weight of constraint \(k\). \\
\(e_k^{\text{start}}\) & Activation epoch of constraint \(k\). \\
\(R_{\text{ramp}}\) & Warmup length for a newly activated constraint. \\
\(\lambda_k^{\text{final}}\) & Final weight of constraint \(k\). \\
\(\bar E_k\) & Rolling mean of the energy of constraint \(k\). \\
\(\bar E_k^{(0)}\) & Reference rolling energy at the start of the adaptive phase. \\
\(\tau\) & Threshold used to release the full constraint weight. \\
\(g(\bar E_k)\) & Energy-adaptive factor used in the weighting schedule. \\
\(\lambda_c(e)\) & Weight of the clash energy. \\
\(\lambda_b(e)\) & Weight of the bond-length energy. \\
\(\lambda_a(e)\) & Weight of the bond-angle energy. \\
\(\lambda_v(e)\) & Weight of the valence energy. \\
\(\mathcal L_{\text{diffusion}}\) & Native diffusion loss of the backbone. \\
\(\mathcal L_{\text{symbolic}}(e)\) & Weighted symbolic loss at epoch \(e\). \\
\(\mathcal L_{\text{total}}(e)\) & Final training loss at epoch \(e\). \\
\(\nabla_{z_s}E\) & Gradient of an energy with respect to the intermediate latent state \(z_s\). \\
\(\nabla_{z_s}E_k\) & Gradient of constraint energy \(E_k\) with respect to \(z_s\). \\
\bottomrule
\end{tabular}
\end{table*}

\clearpage
\section{Extended Preliminaries}
\label{sec:detailed_preliminaries}

\textbf{Overview}. 
We consider a batch of molecules with coordinates \(x \in \mathbb{R}^{B\times N\times 3}\), atom features \(h=\{h^{\mathrm{cat}},h^{\mathrm{int}}\}\) (\textit{e.g.,} cat and int denote categorical and integer, respectively), and node masks \(m_i \in \{0,1\}\), where \(B\) is the batch size, \(N\) is the padded number of atoms, \(C\) is the number of atom types, \(M\) is the number of bond classes, and \(D\) is the latent dimensionality. Here, \(N\) denotes the maximum number of atoms after padding. For example, if molecules in a batch contain different numbers of atoms, they are padded to the same size \(N\). The quantity \(C\) denotes the number of possible atom types. For example, in QM9~\cite{qm9} we typically have \(C=5\), corresponding to H, C, N, O, and F. The quantity \(M\) denotes the number of bond classes. For example, one may use \(M=5\) to represent no bond, single, double, triple, and aromatic bonds. 

The latent molecular state is \(z=[z^{(x)} \| z^{(h)}] \in \mathbb{R}^{B\times N\times D}\), where \(z^{(x)}\) and \(z^{(h)}\) denote the coordinate and feature channels. At diffusion step \(t=t_{\mathrm{int}}/T\), where \(T\) is the total number of diffusion steps, we denote the corrupted latent by \(z_t\) and the previous-step latent by \(z_s\), with \(s=(t_{\mathrm{int}}-1)/T\). We use \(\alpha_t\) and \(\sigma_t\) for the signal and noise coefficients, \(\epsilon_\theta(z_t,t)\) for the noise predictor, and \(z_{xh}=[z_{x\mu}\|z_{h\mu}]\) for the mean latent code produced by the encoder. Soft decoding yields coordinates \(R=\{r_i\}_{i=1}^N\), atom-type distributions \(A_i\), and bond posteriors \(B_{ij}^{(m)}\), where \(m=0\) denotes the no-bond class. These variables define the symbolic state used for energy functions: \(E_{\text{clash}}, E_{\text{bond\_len}}, E_{\text{angle}},\) and \(E_{\text{val}}\).

\textbf{Diffusion Model and Molecular Representation.}
As discussed above, we use GEOLDM~\cite{geoldm} as our molecular diffusion backbone. The molecular state \((x,h)\) is encoded into a latent representation \(z\), and the clean mean code \(z_{xh}=[z_{x\mu}\|z_{h\mu}]\) is used as the latent state modeled by diffusion. The forward process is defined as
\[
z_t=\alpha_t z_0+\sigma_t\epsilon,\qquad \epsilon \sim \mathcal N(0,I),
\]
so \(z_t\) gradually interpolates between clean molecular structure and Gaussian noise. The corresponding reverse transition \(p_\theta(z_s\mid z_t)\) is parameterized by \(\epsilon_\theta(z_t,t)\). For the reverse step used here, the Gaussian kernel is expressed as
\[
\mu = \frac{z_t}{\alpha_{t|s}} - \frac{\sigma_{t|s}^2}{\alpha_{t|s}\sigma_t}\epsilon_\theta(z_t,t),
\qquad
\sigma = \frac{\sigma_{t|s}\sigma_s}{\sigma_t},
\]
which yields \(z_s \sim \mathcal N(\mu,\sigma^2 I)\). This latent-space formulation is important because it exposes a continuous denoising state \(z_s\) on which symbolic constraints can be applied.

\textbf{3D Molecular Variational Autoencoder.}
Since GEOLDM operates in latent space, we require a molecular variational autoencoder to encode molecular inputs into a latent 3D representation. A 3D molecular variational autoencoder provides a continuous latent representation for molecular geometry and atom features. Specifically, it aims to map the structured molecular input \((x,h)\) into a latent state \(z\) that can be modeled more smoothly by diffusion models, and to reconstruct molecular variables from latent states during molecule generation. In our setting, the encoder outputs channel-wise latent statistics \((z_{x\mu},z_{x\sigma},z_{h\mu},z_{h\sigma})\), and the mean code \(z_{xh}=[z_{x\mu}\|z_{h\mu}]\) is used as the clean latent representation passed to the diffusion model. Because the underlying object is a 3D molecule, the encoder and decoder are designed to preserve geometric structure while jointly modeling node features~\cite{egnn}. The decoder thus provides the latent-to-molecule mapping needed for both 3D molecule generation and evaluating symbolic constraint satisfaction.

\textbf{Latent Diffusion Model.}
Latent diffusion applies the diffusion process in the learned latent space \(z\) rather than directly on molecular state \((x,h)\). This separation is useful because the variational autoencoder first compresses molecular geometry and node attributes into a continuous representation, after which diffusion only models the distribution of latent states. In practice, one starts from either an encoded clean latent \(z_0\) during training or a noisy latent \(z_T \sim \mathcal N(0,I)\) during sampling, applies the reverse dynamics \(z_t \rightarrow z_s \rightarrow \cdots \rightarrow z_0\), and finally decodes the recovered latent into a molecule. The key to SensDiff is that latent diffusion exposes a continuous state \(z_s\) on which one can define gradients of symbolic energy constraints. Since \(R\), \(A\), and \(B\) are differentiable functions of \(z_s\), a constraint energy \(E(R,A,B)\) induces gradients \(\nabla_{z_s}E\), making latent space the natural level at which neural denoising and symbolic reasoning can be effectively integrated.

\textbf{Differentiable Molecular Interface.}
The method requires a differentiable map from latent states to chemically meaningful variables. A standard hard decoding path is unsuitable because discretization destroys the gradients needed for training-time evaluation of symbolic constraint satisfaction. Therefore, the soft decoding path maps \(z_s\) to
\[
(R,A)=\mathrm{\texttt{decode\_soft}}(z_s),
\qquad
A_i=\operatorname{\texttt{softmax}}(h_i^{\mathrm{logit}})\cdot m_i,
\]
so that, for valid nodes, \(A_i \in \Delta^{C-1}\) remains a continuous probability vector rather than collapsing to a one-hot atom type. This is crucial because the symbolic branch must evaluate constraints on intermediate denoising states, where atomic identity is still uncertain.

The decoded geometry and soft atom identities are then lifted to a relational molecular representation through a bond posterior head. For each pair \((i,j)\), the pair feature is
\[
\phi_{ij} = \left[\left| r_i - r_j \right|_{2},\, A_i,\, A_j\right]
\]
which is mapped to bond logits \(\ell_{ij}\), symmetrized as \(\tilde{\ell}_{ij}=\tfrac{1}{2}(\ell_{ij}+\ell_{ji})\), and normalized by
\[
B_{ij}^{(m)} =
\frac{\exp(\tilde{\ell}_{ij}^{(m)})}
{\sum_{m'=0}^{M-1}\exp(\tilde{\ell}_{ij}^{(m')})}.
\]
The resulting \(B_{ij}^{(m)}\) is symmetric in \(i\) and \(j\). This molecular interface is essential because bond-length, bond-angle, and valence constraints are not defined on 3D coordinates alone. They require a probabilistic and differentiable relational representation derived from the latent state.

\textbf{Neuro-Symbolic Constraints as Differentiable Energies.}
Our neuro-symbolic formulation evaluates symbolic physicochemical rules on the symbolic state \((R,A,B)\) and maps each rule to a differentiable energy, which can be expressed as
\[
E = -\frac{1}{Z}\sum \omega \log(\kappa + \varepsilon),
\]
where \(\kappa \in (0,1)\) is a soft truth, \(\omega\) is a soft antecedent weight, \(Z\) is a normalization factor, and \(\varepsilon=10^{-8}\). Because atom identities are represented by soft distributions, discrete chemistry statistics must be interpolated continuously. For example, the non-bonded distance lower bound is 
\[
d_{\min,ij}=\sum_{a=1}^{C}\sum_{b=1}^{C}A_{ia}A_{jb}d_{\min}^{ab}.
\]
The bond-length statistics are 
\[
\mu_{ijm}^{\text{bond}}=\sum_{a,b}A_{ia}A_{jb}\mu_{abm}^{\text{bond}},
\qquad
\sigma_{ijm}^{\text{bond}}=\max(\sigma_{\min}^{\text{bond}},\sum_{a,b}A_{ia}A_{jb}\sigma_{abm}^{\text{bond}}),
\]
and the angle statistics are 
\[
\mu_{j,mn}^{\cos}=\sum_{b=1}^{C}A_{jb}\mu_{b,mn}^{\cos,\mathrm{fb}},
\qquad
\sigma_{j,mn}^{\cos}=\max(\sigma_{\min}^{\text{angle}},\sum_{b=1}^{C}A_{jb}\sigma_{b,mn}^{\cos,\mathrm{fb}}),
\]
where \(\sigma_{\min}^{\text{bond}}\) and \(\sigma_{\min}^{\text{angle}}\) are lower bounds used to stabilize the interpolated standard deviations. Likewise, the soft valence of atom \(i\) is defined as
\[
v_i^{\text{soft}}=\sum_{j\neq i}\sum_{m=0}^{M-1}\nu_m B_{ij}^{(m)},
\]
where \(\nu_m\) denotes the bond order weight of bond class \(m\). This quantity is the expected total incident bond order of atom \(i\) under the soft bond posterior \(B_{ij}^{(m)}\), and therefore provides a differentiable relaxation of discrete chemical valence. It is then evaluated against the admissible valence set \(\mathcal{V}(a)\) of atom type \(a\), weighted by the corresponding empirical prior \(\pi_{a,v}\), so that the valence constraint favors values of \(v_i^{\text{soft}}\) that are chemically valid for plausible atom identities. These quantities define the four energy constraints \(E_{\text{clash}}, E_{\text{bond\_len}}, E_{\text{angle}},\) and \(E_{\text{val}}\), corresponding to steric non-overlap, bond-length consistency, bond-angle consistency, and valence validity, respectively.

\newpage
\section{Experimental Setup}
\label{sec:experimental_setup}

\textbf{Datasets.}
We evaluate SensDiff on QM9~\cite{qm9} and GEOM-Drugs~\cite{geom-drug}. QM9 contains 134k small organic molecules with annotated quantum-chemical properties and is a standard benchmark for assessing whether a model can generate chemically plausible 3D molecules while controlling molecular properties. In our setting, the atom vocabulary is \(\{\text{H},\text{C},\text{N},\text{O},\text{F}\}\), and each molecule has at most nine heavy atoms. We use QM9 for both unconditional and conditional molecule generation. GEOM-Drugs is a larger and more challenging benchmark of drug-like molecules with diverse 3D conformers. Compared with QM9, it features larger structures, richer topologies, and broader geometric variability, offering a more demanding test of scalability and robustness. We use GEOM-Drugs to evaluate whether the proposed constraints remain effective beyond the small-molecule regime. We split each dataset into 80\%, 10\%, and 10\% for training, validation, and testing, respectively.

\textbf{Baselines.}
We compare SensDiff against a broad set of recent 3D molecular generators chosen to cover the main modeling paradigms used in the literature. The first group consists of strong geometry-aware generative backbones. EDM~\cite{edm} is the canonical equivariant diffusion baseline for 3D molecule generation and serves as the standard reference for direct diffusion in Euclidean space. GCDM~\cite{gcdm} is another diffusion-based model that aims to model molecular geometry more completely, making it a useful comparator for testing whether better geometric realism can be achieved by geometry modeling alone, without explicit symbolic constraints. SLDM~\cite{sldm} emphasizes a simplified straight-line diffusion trajectory and represents methods that improve generation through trajectory design and efficiency rather than through domain constraints. RADM~\cite{radm} provides a complementary comparison because it handles rotations through explicit alignment rather than built-in equivariance, allowing us to assess whether SensDiff remains competitive against strong models with a different inductive bias. The second group captures methods that inject extra information during generation. EEGSDE~\cite{eegsde} is especially relevant because it uses energy-guided sampling. This makes it a natural comparator for SensDiff: both methods leverage scientific signals, but EEGSDE applies guidance primarily at inference time, whereas SensDiff integrates symbolic constraints into the training objective itself. The third group focuses on latent-space and representation-centric generation. GEOLDM~\cite{geoldm} is the most direct backbone baseline because SensDiff is built on top of it. This comparison isolates the contribution of the proposed neuro-symbolic component from the underlying latent diffusion architecture. UAE-3D~\cite{uae-3d} is another strong latent-space generator that emphasizes a more unified and information-preserving 3D representation. Finally, we compare against alternative generative paradigms and property-aware models. EquiFM~\cite{equifm} represents equivariant flow matching, which learns a continuous probability transport process instead of a denoising diffusion chain. GeoBFN~\cite{geobfn} represents Bayesian flow networks and provides a non-diffusion comparison point while targeting the same 3D molecule generation problem. UniGEM~\cite{unigem} jointly models molecule generation and property prediction, making it particularly relevant for conditional generation experiments. In summary, these baselines cover direct diffusion, guided diffusion, latent diffusion, flow matching, Bayesian flow networks, and unified generation-property modeling. All metrics are computed by us for all baselines under a unified evaluation pipeline, and the values may differ slightly from those reported in the original papers.

\textbf{Evaluation Metrics.}
For generation quality, we report validity, validity \& uniqueness, and novelty. Validity is the fraction of generated samples that are chemically valid under the standard benchmark pipeline. Validity \& uniqueness is the fraction of valid molecules that are non-duplicated, preventing inflated performance from repeated valid samples. Novelty is the fraction of generated molecules absent from the training set. To assess 3D plausibility, we report atom and molecule stability, along with bond-length, bond-angle, and dihedral-angle statistics. Atom and molecule stability measure whether local environments and whole molecules pass standard stability checks in prior 3D benchmarks, while the internal-coordinate statistics quantify deviations from the empirical geometry distribution, where smaller deviation is better. For QM9 quantum properties, we evaluate six target properties, namely $\alpha$, $\Delta \varepsilon$, $\varepsilon_{\mathrm{HOMO}}$, $\varepsilon_{\mathrm{LUMO}}$, $\mu$, and $C_v$, with units Bohr$^3$, meV, meV, meV, D, and \(\frac{\text{cal}}{\text{mol}}\,\text{K}\), respectively. For all six properties, we report mean absolute error (MAE) between the requested target value and the achieved value of the generated molecule. MAE provides a direct measure of controllability: lower MAE means the model follows the desired scientific target more accurately. For constrained generation in drug-like molecules, we report the achieved values of LogP~\cite{logp}, Molecular Weight (MW)~\cite{mw}, Hydrogen-Bond Acceptors (HBA)~\cite{hbahbd}, Hydrogen-Bond Donors (HBD)~\cite{hbahbd}, Topological Polar Surface Area (TPSA)~\cite{tpsa}, Quantitative Estimate of Drug-likeness (QED)~\cite{qed}, and Synthetic Accessibility (SA) Score~\cite{sascore}. These descriptors cover hydrophobicity, molecular size, hydrogen-bonding capacity, polarity, overall drug-likeness, and synthetic accessibility. We report the raw descriptor values because the goal is not to optimize every descriptor in the same direction, but to place generated molecules inside a user-specified target range of chemical property space. To measure whether all requested conditions are satisfied, we report the constraint satisfaction rate (CSR), defined as the fraction of generated molecules that satisfy the specified constraints. CSR is important in multi-objective settings because property-wise averages can be misleading. A molecule may be close to one target yet conflict with several others. When similarity preservation is part of the task, we report Tanimoto similarity, a standard fingerprint-based overlap score in $[0,1]$ where larger values indicate greater structural similarity to the reference molecule or scaffold~\cite{tanimoto}. We repeat each experiment ten times and report the mean results.

\vspace{6\baselineskip}
\section{Implementation Details}
\label{sec:implementation_details}

\textbf{Overview.}
Our implementation follows the pipeline described in the main text, and we provide additional details here. We use QM9 with the standard GEOLDM preprocessing pipeline, keep integer charges in the node representation, and retain the default normalization factors \([1,4,10]\). Molecules are represented by coordinates \(x \in \mathbb{R}^{B\times N\times 3}\), atom features \(h=\{h^{\mathrm{cat}},h^{\mathrm{int}}\}\), and node masks \(m_i\), with \(N=29\) after padding on QM9, \(C=5\) atom types, and \(M=5\) bond classes. The latent representation has \(\mathrm{latent\_nf}=1\), so the latent dimensionality is \(D=4\). The encoder, decoder, and diffusion denoiser all inherit the EGNN configuration with 9 layers, hidden size 256, attention, and tanh gating. Diffusion model uses \(T=1000\) steps, the polynomial-2 noise schedule, noise precision \(10^{-5}\), and the standard \(\ell_2\) denoising objective. The role of SensDiff is to expose a differentiable symbolic state around the reverse-time latent variable \(z_s\), so that the same denoising trajectory can be supervised by the four physicochemical constraints defined in Section~\ref{sec:method}, rather than to replace the GEOLDM generative backbone itself.

\textbf{Preparation of the symbolic branch.} Before training the constrained diffusion model, we first make the symbolic branch fully operational. The first step is an offline pass over the QM9 training split, where RDKit is used to reconstruct bond graphs from the reference 3D geometries. From this pass, we accumulate the atom-pair non-bonded distance lower bounds \(d_{\min}^{ab}\), the bond-length statistics \((\mu_{abm}^{\text{bond}},\sigma_{abm}^{\text{bond}})\), the center-conditioned fallback angle-cosine statistics \((\mu_{b,mn}^{\cos,\mathrm{fb}},\sigma_{b,mn}^{\cos,\mathrm{fb}})\), and the admissible valence sets \(\mathcal V(a)\) together with their empirical priors \(\pi_{a,v}\). These tables are not used as hard lookups during training. Instead, they are converted into differentiable quantities through the soft interpolation rules in Equation~\ref{eq:dmin_interp}, Equation~\ref{eq:bond_stats_interp}, and Equation~\ref{eq:angle_stats_interp}. In practice, although the full triplet-conditioned angle table is precomputed, the implementation uniformly uses the center-conditioned fallback table in Equation~\ref{eq:angle_stats_interp} for the vectorized angle kernel, because many fully conditioned triplets are sparse on QM9 and the fallback version is both more stable and computationally easier. We then train the GEOLDM VAE for 300 epochs with batch size 64, learning rate \(10^{-4}\), KL weight 0.01, and EMA decay 0.9999, and we separately pretrain the bond posterior head with RDKit-derived bond orders as five-class supervision. Once Stage 0 is complete, both the VAE and the bond posterior head are frozen so that all subsequent symbolic gradients reshape only the diffusion dynamics.

\textbf{Differentiable symbolic interface.} The implementation uses the same soft molecular interface described in Section~\ref{sec:method}, but we make the engineering role of each component explicit here. First, the decoder exposes a dedicated \texttt{decode\_soft} path, because a hard \(\arg\max\) over atom-type logits would destroy the gradients needed to optimize the symbolic energies. Given a reverse-time latent \(z_s\), \texttt{decode\_soft} returns recentered coordinates \(R\) and simplex-valued atom identities \(A_i\), where padding positions are suppressed by the node mask. Second, a \texttt{BondPosteriorHead} maps each pair feature \(\phi_{ij} = \left[\left| r_i - r_j \right|_{2},\, A_i,\, A_j\right]\) to logits over the five bond classes and then converts them into \(B_{ij}^{(m)}\). This head uses two hidden layers of width 128 with SiLU activations, directly symmetrizes the logits so that \(B_{ij}^{(m)}=B_{ji}^{(m)}\), and suppresses self-loops and padded pairs by adding large negative penalties before softmax, which ensures that the no-bond class dominates on the diagonal and at masked positions. The resulting state \((R,A,B)\) is the implementation-level realization of the symbolic state used in Section~\ref{sec:method}. It is also the critical bridge that makes the general energy definition in Equation~\ref{eq:energy_def} differentiable with respect to the latent variable.

\textbf{Constraint evaluation.} The four symbolic energies are implemented exactly according to the manuscript: steric non-overlap in Equation~\ref{eq:clash_energy}, bond-length consistency in Equation~\ref{eq:bond_len_energy}, bond-angle consistency in Equation~\ref{eq:angle_energy}, and valence validity in Equation~\ref{eq:val_energy}. The implementation directly follows the soft-truth construction in Equation~\ref{eq:energy_def}, so each rule is evaluated as a normalized negative log soft satisfaction score. For \(E_{\text{clash}}\), the temperature is fixed to \(\tau_c=0.05\). For \(E_{\text{bond\_len}}\) and \(E_{\text{angle}}\), standardized residuals are regularized with the Pseudo-Huber penalty \(\rho(u)=\sqrt{1+u^2}-1\)~\cite{hartley2003multiple}, which avoids the severe underflow and vanishing-gradient behavior that would arise from a pure Gaussian penalty on highly corrupted denoising states. For numerical stability, we impose lower bounds of \(0.03\) on the interpolated bond length and angle standard deviations, use \(\varepsilon=10^{-8}\) in all logarithmic and normalization terms, clamp cosine values to a safe interval before evaluating the angle term, and restrict the angle computation to the top-\(K_{\mathrm{nbr}}\) neighbors with \(K_{\mathrm{nbr}}=4\), where neighbors are ranked by total bonding probability. For valence, the implementation uses the bond-order weights \(\nu_0=0\), \(\nu_1=1\), \(\nu_2=2\), \(\nu_3=3\), and \(\nu_{\mathrm{aroma}}=1.5\), and the RBF bandwidth is fixed to \(\sigma_v=0.15\). All energy computations remain fully vectorized and mask-aware, so padded atoms do not contribute either to the energy values or to their gradients.

\textbf{Training objective and one-step reverse surrogate.} The training objective follows the exact decomposition already stated in Equation~\ref{eq:symbolic_loss_final} and Equation~\ref{eq:total_loss}. In every iteration, we first load the batch with non-blocking GPU transfers, recenter the coordinates with \texttt{remove\_mean\_with\_mask}, optionally apply Gaussian data augmentation, and compute the native GEOLDM objective \(\mathcal L_{\text{diffusion}}\). We then run the frozen encoder to obtain the clean mean latent code \(z_{xh}=[z_{x\mu}\|z_{h\mu}]\), immediately detach this tensor so that symbolic gradients do not flow back into the encoder, sample a random timestep \(t_{\mathrm{int}}\sim\mathcal U\{1,\dots,T\}\), and corrupt the latent using the same forward diffusion process as the backbone. Rather than evaluating the symbolic branch only on the final decoded molecule, we use the one-step reverse surrogate in Equation~\ref{eq:reverse_sampling}. Concretely, we form \(z_t\), compute one reverse transition \(z_t \rightarrow z_s\) through the GEOLDM reverse kernel, decode \(z_s\) into \((R,A)\), infer \(B\), and then evaluate whichever of the four energies are active at the current training stage, while the inactive terms are set to zero by the stage scheduler. This choice is a key engineering compromise as it keeps the symbolic supervision attached to intermediate denoising states, exactly as required by the method, while avoiding the prohibitive cost of unrolling a full \(T\)-step reverse chain inside each training iteration. Once the energies are computed, we assemble \(\mathcal L_{\text{symbolic}}(e)\) from Equation~\ref{eq:symbolic_loss_final}, add it to the diffusion objective as in Equation~\ref{eq:total_loss}, and backpropagate through the one-step reverse kernel so that only the diffusion network is updated. No additional connectivity penalty, loss term, or extra KL term is introduced beyond the native GEOLDM objective and the four symbolic energies.

\textbf{Curriculum and adaptive weighting.} The activation schedule matches the coarse-to-fine nature of diffusion molecular generation. Stage 1 covers epochs 1--500 and activates only \(E_{\text{clash}}\), Stage 2 covers epochs 501--1400 and adds \(E_{\text{bond\_len}}\), Stage 3 covers epochs 1401--2200 and adds \(E_{\text{angle}}\), and Stage 4 covers epochs 2201--3000 and adds \(E_{\text{val}}\). The final weights are \(\lambda_c=4.0\), \(\lambda_b=3.0\), \(\lambda_a=2.0\), and \(\lambda_v=1.0\). Their epoch-dependent values are controlled by Equation~\ref{eq:lambda_schedule}. Each newly activated term is first linearly ramped for 100 epochs until it reaches half of its target weight, after which the remaining half is released according to the rolling energy decrease. The implementation maintains a 20-epoch rolling mean for each constraint, records the ramp-end snapshot \(\bar E_k^{(0)}\), uses the target threshold \(\tau=0.03\), and applies the same adaptive factor that appears in Equation~\ref{eq:lambda_schedule}. To prevent abrupt destabilization when a newly introduced rule yields unusually large gradients, the scheduler includes a safeguard: if the instantaneous energy exceeds twice its rolling mean, the constraint is frozen at half strength for 50 epochs before adaptive release resumes. This mechanism is not a separate modeling component. It implements the curriculum and is crucial for stabilizing the stage-wise introduction of the four constraints during training.

\textbf{Optimization and validation protocol.} Training uses learning rate \(10^{-4}\), EMA decay 0.9999, and adaptive queue-based gradient clipping rule inherited from the implementation notes. We enable TF32 matrix multiplication on GPU. Mixed precision is handled by using BF16 autocast for the forward and backward passes. Otherwise, the code falls back to FP16 with GradScaler. Gradients are unscaled before clipping, after which the optimizer step and EMA update are performed. A dedicated logger records the epoch, iteration, stage, individual symbolic energies, diffusion loss, symbolic loss, and total loss throughout training. Each epoch ends with a stability check, and every 20 epochs we run validation, compute NLL-style statistics, generate 1000 molecules with EMA parameters, and select the best checkpoint. These details matter because the symbolic branch increases computational cost and gradient variance. Mixed precision, adaptive clipping, EMA, and stage-wise logging are therefore key to making the training recipe reproducible rather than merely well defined.

\begin{figure*}[t]
    \centering
    \includegraphics[width=0.9\textwidth]{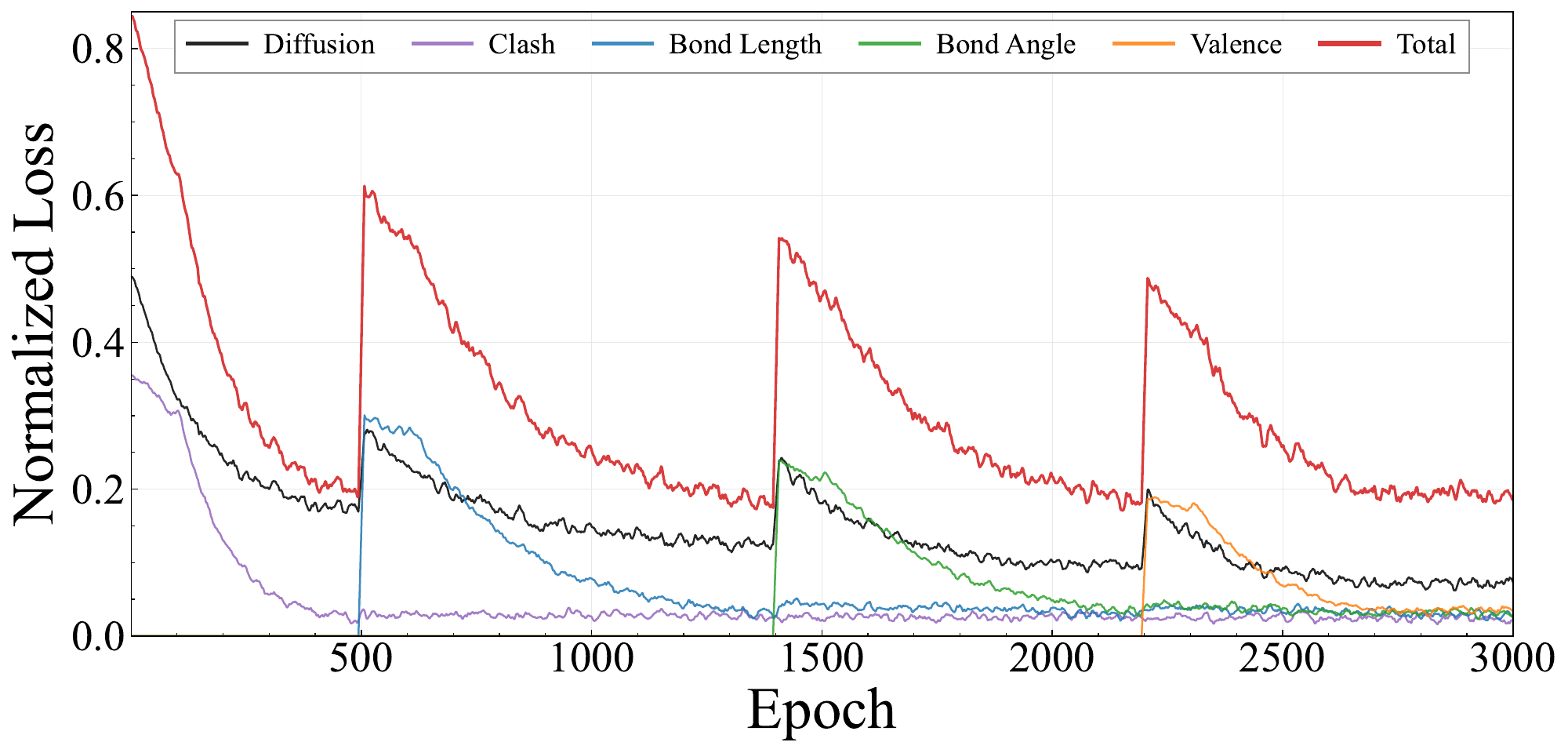}
    \caption{\textbf{Training loss on QM9}~\cite{qm9}. Symbolic energy constraints are activated sequentially in a coarse-to-fine curriculum. Each triggers a transient spike in total loss that is quickly absorbed as adaptive weighting ramps up the newly introduced constraint.}
    \label{fig:loss_qm9}
\end{figure*}

\textbf{Inference-time latent refinement.} Sampling keeps the standard GEOLDM reverse chain but inserts symbolic constraints after every reverse step. We first sample \(z_T\) with center-of-gravity-zero Gaussian noise for the coordinate channels and standard Gaussian noise for the feature channels, and then run the reverse kernel from \(t=T\) to \(t=1\), using the same reverse transition as Equation~\ref{eq:reverse_sampling}. At inference, the stage is fixed to 4, so all four constraints are active regardless of which training stage was used to produce the checkpoint. After each reverse step, we perform \(K=3\) refinement steps on the current latent state. In each refinement step, a detached float32 copy of the latent is marked as requiring gradients, decoded to \((R,A)\), converted to \(B\), and evaluated with the full weighted symbolic energy corresponding to the final form of Equation~\ref{eq:symbolic_loss_final}. We then compute the gradient with respect to the latent variable only, replace non-finite values with zero if necessary, mask out padded positions, and normalize the gradient independently for each molecule in the batch. This per-sample normalization matters because otherwise a single molecule with large energy could dominate the batch update direction. We apply the normalized gradient with step size \(\eta_t=\alpha_t/K\), where \(\alpha_t=\alpha_{\max}(1-t/T)^2\) and \(\alpha_{\max}=1.0\). This quadratic schedule makes refinement negligible at the noise-dominated start of the reverse chain and strongest near the end, when structure is sufficiently formed for constraints to be informative. After each step, we recenter the coordinate channels and remask the feature channels. If the symbolic energy is NaN or Inf, we skip refinement and keep the original reverse sample.

\textbf{Hardware and efficiency}.  
Figure~\ref{fig:loss_qm9} visualizes SensDiff’s training loss on QM9. We adopt a sequential training strategy that progressively introduces constraints on top of the diffusion loss, as reflected by the periodic spikes in the total loss curve whenever a new term is activated. Despite these transient increases, all individual loss components converge rapidly and stabilize, demonstrating that our curriculum-based weight scheduling enables the model to effectively internalize each symbolic constraint without disrupting previously learned objectives. Training is run on an NVIDIA H200 GPU, while sampling is run on an NVIDIA RTX 4090 GPU. We use QM9 as an example and report diffusion hyperparameters in Table~\ref{tab:hyperparameters}. On average, training takes 62.358 minutes on QM9 and 74.263 minutes on GEOM-Drugs. Sampling takes 420 milliseconds per molecule on average, close to our backbone GEOLDM (417 milliseconds). SensDiff is faster than many baselines, such as EDM (528 milliseconds), EEGSDE (731 milliseconds), GCDM (571 milliseconds), and UniGEM (587 milliseconds). With symbolic constraints enabled at sampling time, the cost increases slightly to 472 milliseconds per molecule. In summary, our training and sampling efficiency are comparable to those of popular molecular diffusion backbones.

\clearpage
\begin{center}
\footnotesize
\renewcommand{\arraystretch}{1.05}
\setlength{\tabcolsep}{6pt}
\begin{longtable}{p{0.60\textwidth}p{0.22\textwidth}}
\caption{Hyperparameters used in the QM9 implementation of SensDiff.}
\label{tab:hyperparameters} \\
\toprule
\textbf{Hyperparameter} & \textbf{Value} \\
\midrule
\endfirsthead

\toprule
\textbf{Hyperparameter} & \textbf{Value} \\
\midrule
\endhead

\bottomrule
\endfoot

\multicolumn{2}{c}{\textbf{Data}} \\
\midrule
atom types \(C\) & 5 (H, C, N, O, F) \\
normalize\_factors & [1, 4, 10] \\
include\_charges & True \\
\midrule

\multicolumn{2}{c}{\textbf{3D Molecular Variational Autoencoder}} \\
\midrule
latent\_nf & 1 \\
kl\_weight & 0.01 \\
n\_epochs & 300 \\
batch\_size & 64 \\
learning rate & \(10^{-4}\) \\
ema\_decay & 0.9999 \\
\midrule

\multicolumn{2}{c}{\textbf{Diffusion Model}} \\
\midrule
diffusion\_steps \(T\) & 1000 \\
noise\_schedule & \texttt{polynomial\_2} \\
noise\_precision & \(10^{-5}\) \\
loss\_type & \(\ell_2\) \\
n\_epochs & 3000 \\
batch\_size & 64 \\
learning rate & \(10^{-4}\) \\
ema\_decay & 0.9999 \\
ode\_regularization & \(10^{-3}\) \\
gradient clipping & adaptive queue \\
test\_epochs & 20 \\
n\_stability\_samples & 1000 \\
\midrule

\multicolumn{2}{c}{\textbf{EGNN Backbone}} \\
\midrule
n\_layers & 9 \\
nf (hidden) & 256 \\
attention & True \\
tanh gating & True \\
\midrule

\multicolumn{2}{c}{\textbf{Bond Posterior Head}} \\
\midrule
hidden\_nf & 128 \\
n\_bond\_classes \(M\) & 5 \\
activation & SiLU~\cite{elfwing2018sigmoid} \\
\midrule

\multicolumn{2}{c}{\textbf{Symbolic Energies}} \\
\midrule
\(\tau_c\) & 0.05 \AA \\
\(\delta\) (Pseudo-Huber) & 1.0 \\
\(\sigma_{\min}^{\text{bond}}\) & 0.03 \AA \\
\(\sigma_{\min}^{\text{angle}}\) & 0.03 \\
\(\sigma_v\) & 0.15 \\
\(K_{\mathrm{nbr}}\) & 4 \\
\(\varepsilon\) & \(10^{-8}\) \\
\midrule

\multicolumn{2}{c}{\textbf{Training Curriculum}} \\
\midrule
stage\_epochs & (500, 1400, 2200) \\
ramp\_epochs & 100 \\
\(\lambda_c\) & 4.0 \\
\(\lambda_b\) & 3.0 \\
\(\lambda_a\) & 2.0 \\
\(\lambda_v\) & 1.0 \\
freeze spike ratio & \(2.0\times\) rolling mean \\
freeze duration & 50 epochs \\
\midrule

\multicolumn{2}{c}{\textbf{Diffusion Sampling}} \\
\midrule
\(\alpha_{\max}\) & 1.0 \\
\(K\) & 3 \\
\end{longtable}
\end{center}

\section{Geometric Plausibility of SensDiff}
\label{sec:Geometric Plausibility}

\begin{figure*}[t]
    \centering
    \includegraphics[width=0.95\textwidth]{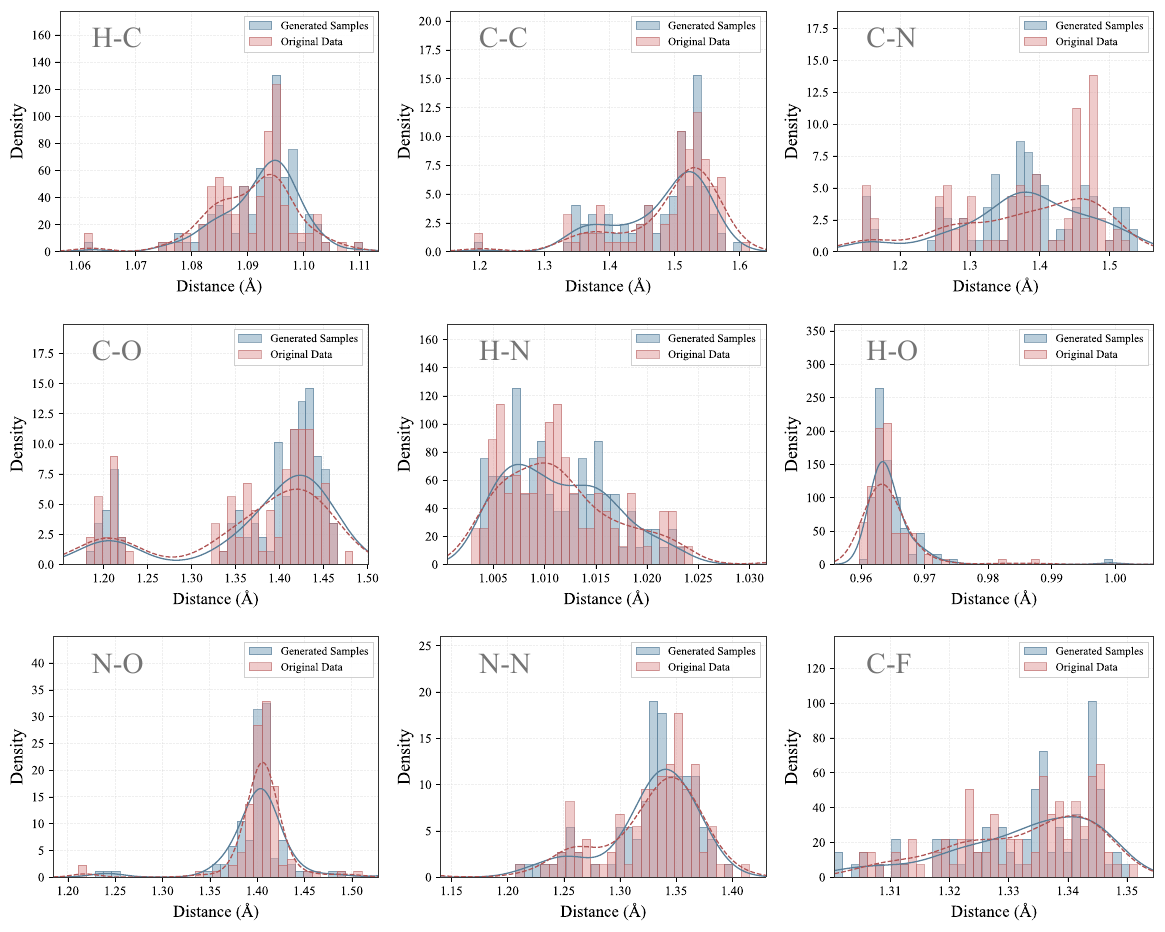}
    \caption{Common bond-length distribution in QM9~\cite{qm9}.}
    \label{fig:qm9_bond_distribution}
\end{figure*}

\begin{figure*}[t]
    \centering
    \includegraphics[width=0.95\textwidth]{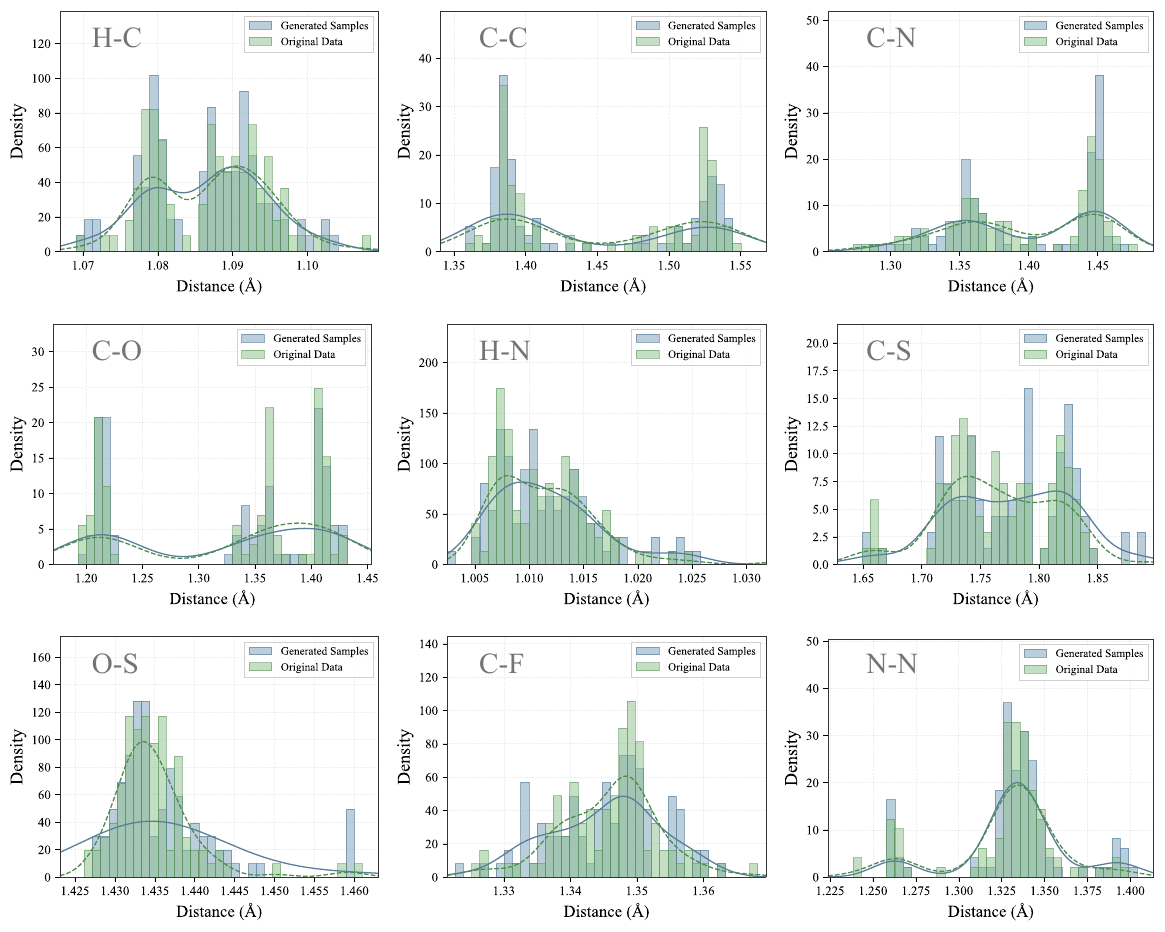}
    \caption{Common bond-length distribution in GEOM-Drugs~\cite{geom-drug}.}
    \label{fig:gd_bond_distribution}
\end{figure*}

\textbf{Empirical Bond-Length Distribution.}
The geometric structure of a molecule is important because it largely determines its physicochemical properties and functional behavior~\cite{NEURIPS2025_3792ddbf, an2026mmcpgen, zeyu2026discrete, hong2026geometric}. As presented in Figure~\ref{fig:qm9_bond_distribution} and Figure~\ref{fig:gd_bond_distribution}, this study evaluates geometric compliance at the level of bond-specific empirical distributions. This analysis is motivated by the role of \(E_{\text{bond\_len}}\) in our method. Since \(E_{\text{bond\_len}}\) explicitly constrains 1-hop geometry by matching interatomic distances to chemically plausible bond-length statistics, a natural question is whether the generated molecules recover the same bond-length distributions as the original data, rather than merely achieving higher validity or stability scores. Therefore, we compare the empirical bond-length distributions of common bond types between the reference molecules and the molecules generated by SensDiff on both QM9 and GEOM-Drugs. This study directly tests whether the symbolic bond-length constraint is reflected in the local geometry of the generated samples. The results show that generated bond-length distributions closely track those of the original data across common bond types. On QM9, the agreement is strong in both peak location and spread, indicating that SensDiff preserves the dominant local bond geometry of the training distribution. On GEOM-Drugs, where the bond environments are more diverse and several bond types exhibit broader or multi-modal structure, the generated distributions still follow the main modes of the reference data well. This supports the central claim of the paper that the symbolic bond-length constraint improves geometric compliance at the local structural level, rather than only improving assessment metrics.

\begin{figure*}[t]
    \centering
    \includegraphics[width=0.95\textwidth]{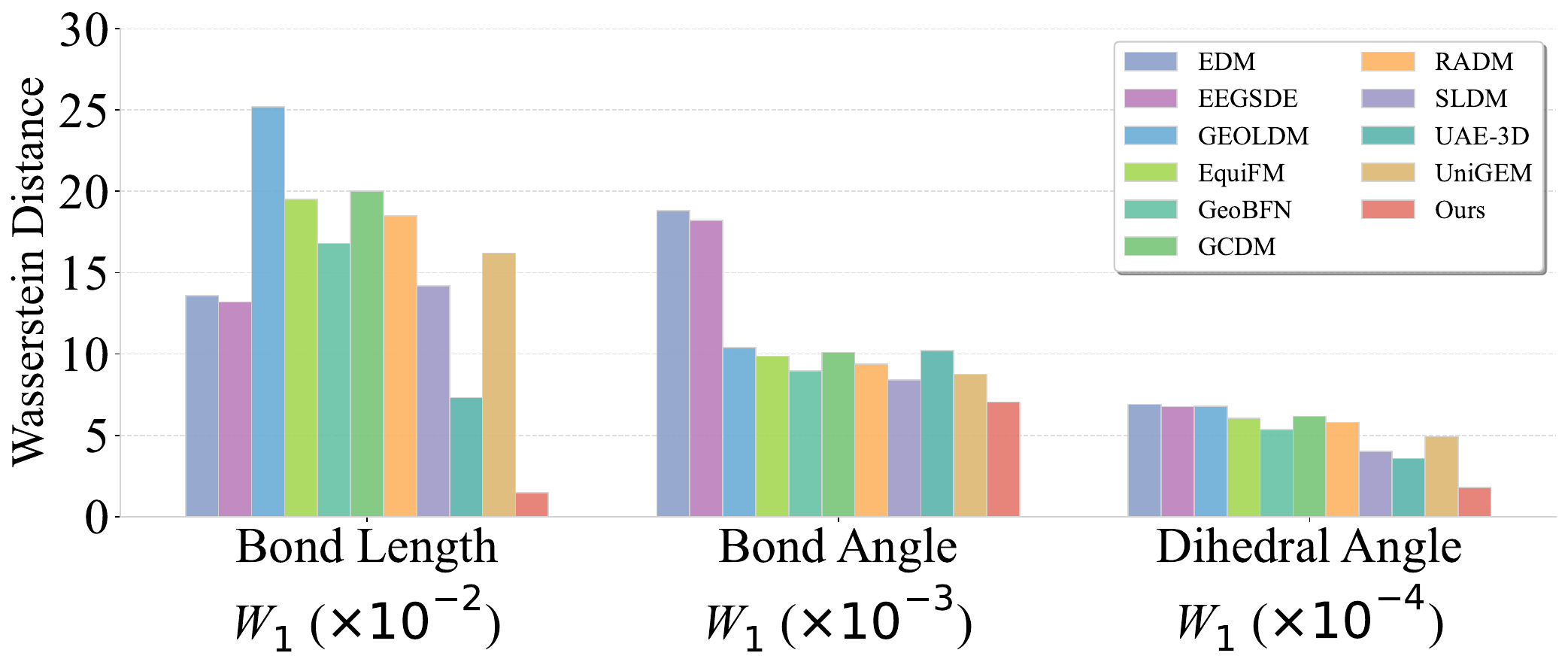}
    \caption{Distributional geometric compliance measured by Wasserstein distance on QM9~\cite{qm9}. Lower values indicate a better match between the generated and reference geometry distributions.}
    \label{fig:wasserstein_geometry}
\end{figure*}

\textbf{Distributional Geometric Compliance Across Methods.}
Figure~\ref{fig:wasserstein_geometry} compares different methods through a unified distributional metric. While the empirical bond-length distribution provide a fine-grained view for individual bond types, they do not directly summarize how well a method matches the overall geometry distribution relative to competing baselines. Consequently, we measure geometric compliance using the Wasserstein distance between the generated and reference distributions. Importantly, we report this discrepancy not only for bond length, but also for bond angle and dihedral angle. This gives a comprehensive evaluation across 1-hop, 2-hop, and higher-order local geometry, and therefore tests whether improvements in bond-level structure propagate to richer geometric structure. The Wasserstein analysis further supports our main claim. SensDiff achieves the lowest distributional discrepancy among all compared methods on bond length, bond angle, and dihedral angle. The strongest gain appears in bond length, which is consistent with our design: bond length is the first chemically meaningful local signal enforced by the coarse-to-fine curriculum. At the same time, the advantage extends to bond angle and dihedral angle, showing that better 1-hop geometric consistency also improves higher-order local structure. Accordingly, these results indicate that SensDiff improves geometric compliance not only at the level of individual bond statistics, but also at the level of the overall geometry distribution.

\clearpage
\section{Unconditional 3D Molecule Generation}
\label{sec:append_unconditional_generation}

As presented in Table~\ref{tab:unconditional_qm9_full} and Table~\ref{tab:unconditional_gd_full}, SensDiff achieves the best performance across nearly all metrics on both QM9 and GEOM-Drugs. A particularly noteworthy observation is that existing methods tend to exhibit a trade-off between chemical validity and geometric accuracy: UAE-3D~\cite{uae-3d} attains low bond-length and bond-angle errors but suffers a significant drop in validity on GEOM-Drugs, whereas high-validity methods such as SLDM~\cite{sldm} and GEOLDM~\cite{geoldm} show considerably larger geometric deviations. SensDiff resolves this tension by simultaneously achieving 97.2\% validity and geometric errors comparable to or lower than UAE-3D~\cite{uae-3d}, which we attribute to the coarse-to-fine curriculum that first establishes correct spatial geometry through steric constraints before enforcing finer chemical rules. The advantage of explicit constraint supervision becomes more pronounced on GEOM-Drugs, where molecular complexity is substantially higher. SensDiff surpasses the strongest baseline by 12.6\% in molecule stability, suggesting that purely data-driven methods struggle to implicitly capture all relevant physicochemical regularities as molecular size grows. Meanwhile, SensDiff maintains the highest validity \& uniqueness and competitive novelty, indicating that the symbolic constraints do not overly restrict generative diversity but rather guide the model to explore chemically plausible regions of the molecular space more effectively. These results collectively demonstrate that structured domain knowledge and data-driven generation are complementary. The former provides reliable inductive biases for fundamental physicochemical properties, while the latter captures complex distributional patterns that are difficult to encode symbolically.

\begin{figure*}[t]
    \centering
    \includegraphics[width=1.0\textwidth]{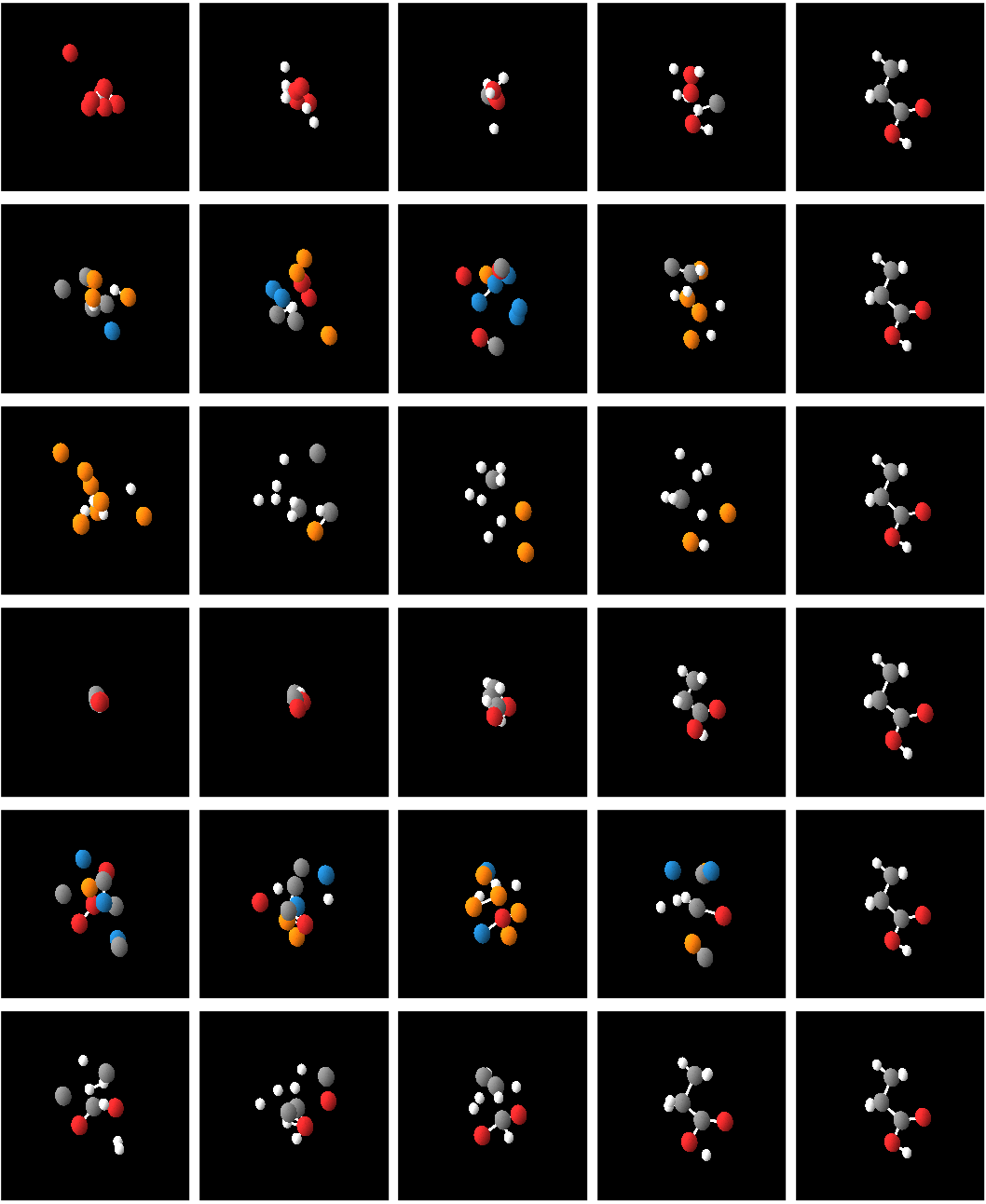}
    \caption{Comparison of molecule generation trajectories for the same sample (ID: 004) across methods. From top to bottom, we show samples generated by SensDiff (ours), GEOLDM~\cite{geoldm}, RADM~\cite{radm}, SLDM~\cite{sldm}, UAE-3D~\cite{uae-3d}, and UniGEM~\cite{unigem}.}
    \label{fig:mol_tra_004_compressed}
\end{figure*}

\begin{figure*}[t]
    \centering
    \includegraphics[width=1.0\textwidth]{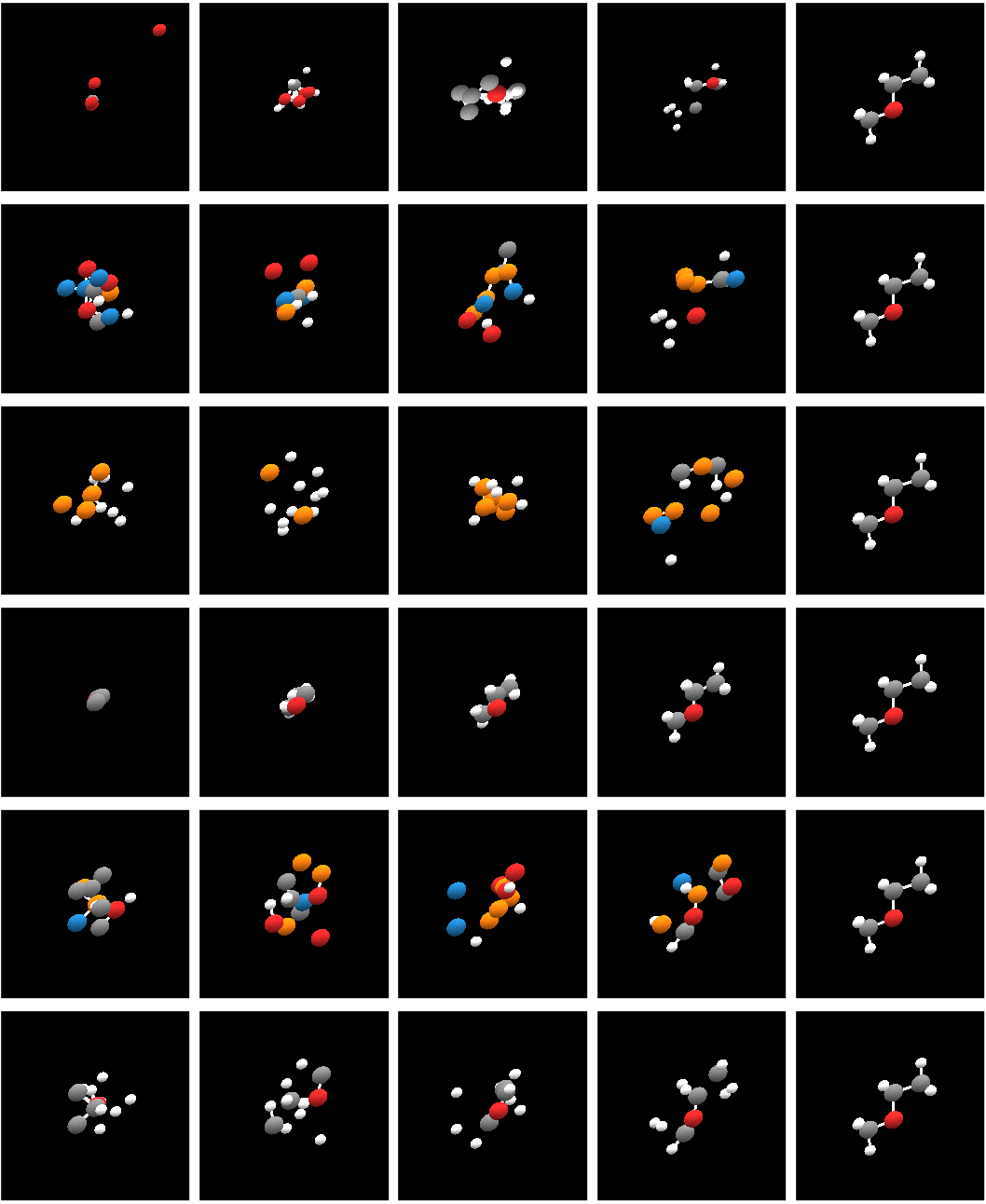}
    \caption{Comparison of molecule generation trajectories for the same sample (ID: 013) across methods. From top to bottom, we show samples generated by SensDiff (ours), GEOLDM~\cite{geoldm}, RADM~\cite{radm}, SLDM~\cite{sldm}, UAE-3D~\cite{uae-3d}, and UniGEM~\cite{unigem}.}
    \label{fig:mol_tra_013_compressed}
\end{figure*}

\begin{figure*}[t]
    \centering
    \includegraphics[width=1.0\textwidth]{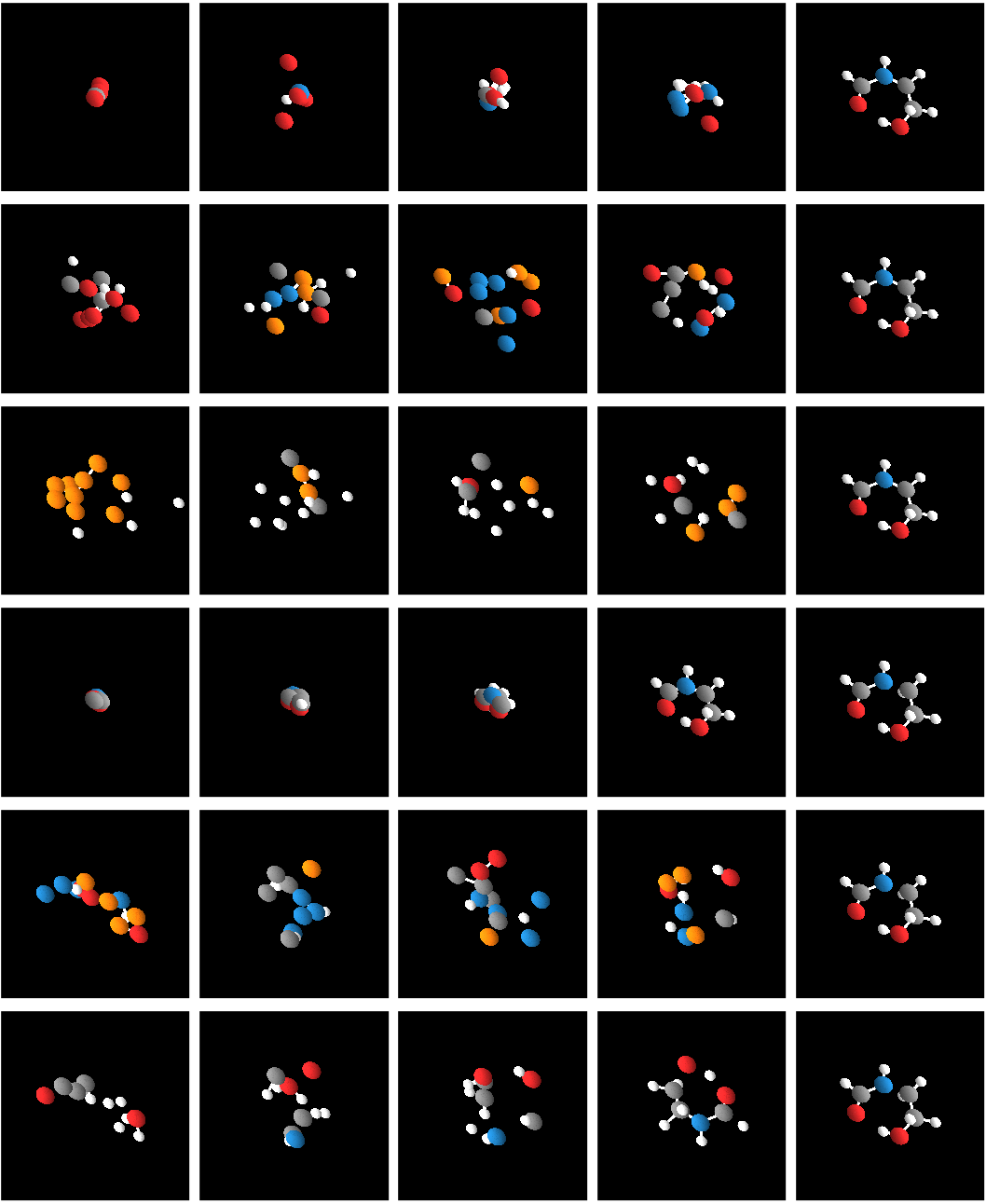}
    \caption{Comparison of molecule generation trajectories for the same sample (ID: 026) across methods. From top to bottom, we show samples generated by SensDiff (ours), GEOLDM~\cite{geoldm}, RADM~\cite{radm}, SLDM~\cite{sldm}, UAE-3D~\cite{uae-3d}, and UniGEM~\cite{unigem}.}
    \label{fig:mol_tra_026_compressed}
\end{figure*}

\begin{figure*}[t]
    \centering
    \includegraphics[width=1.0\textwidth]{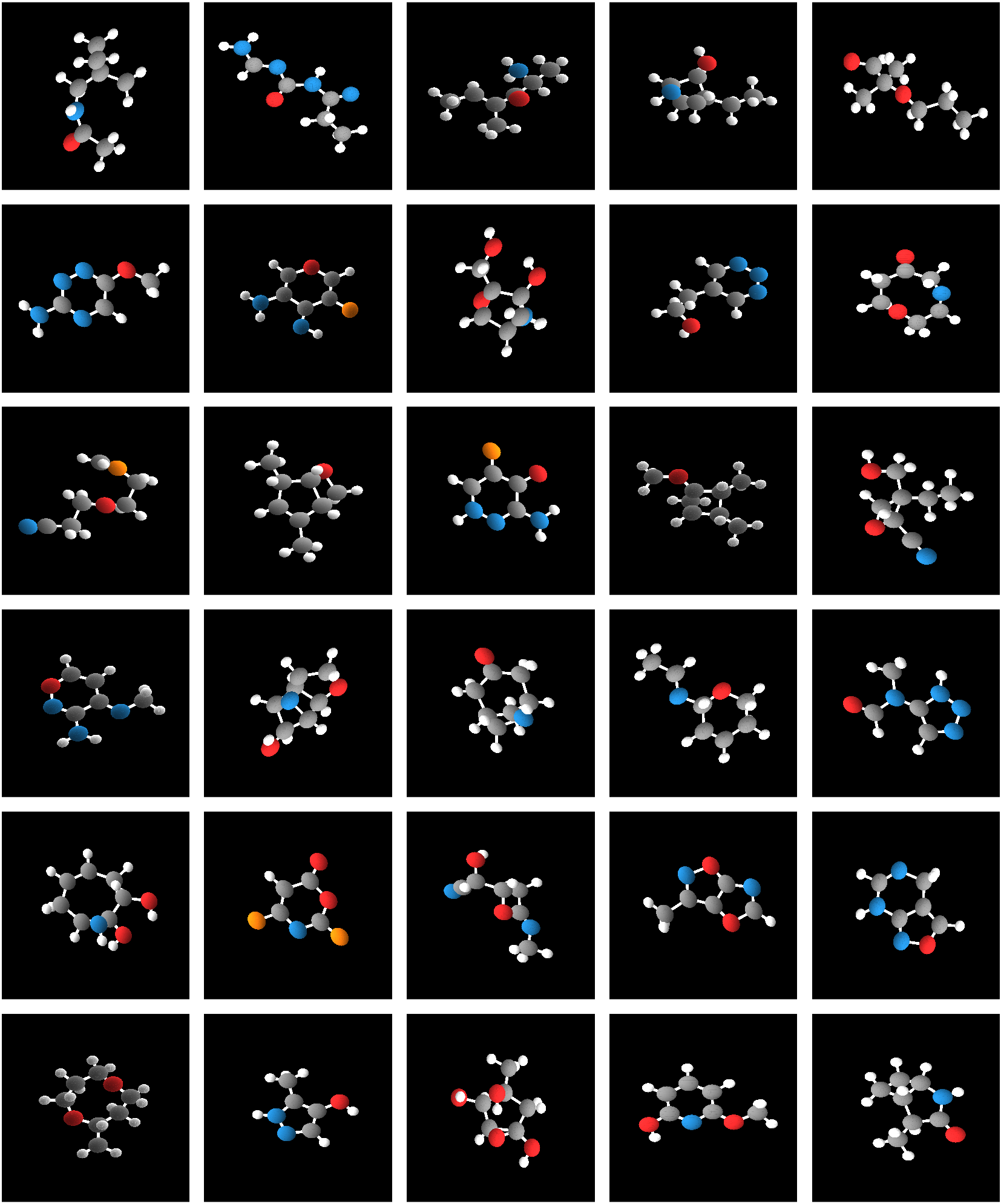}
    \caption{Random samples from QM9~\cite{qm9} generations.}
    \label{fig:qm9_samples}
\end{figure*}

\begin{figure*}[t]
    \centering
    \includegraphics[width=1.0\textwidth]{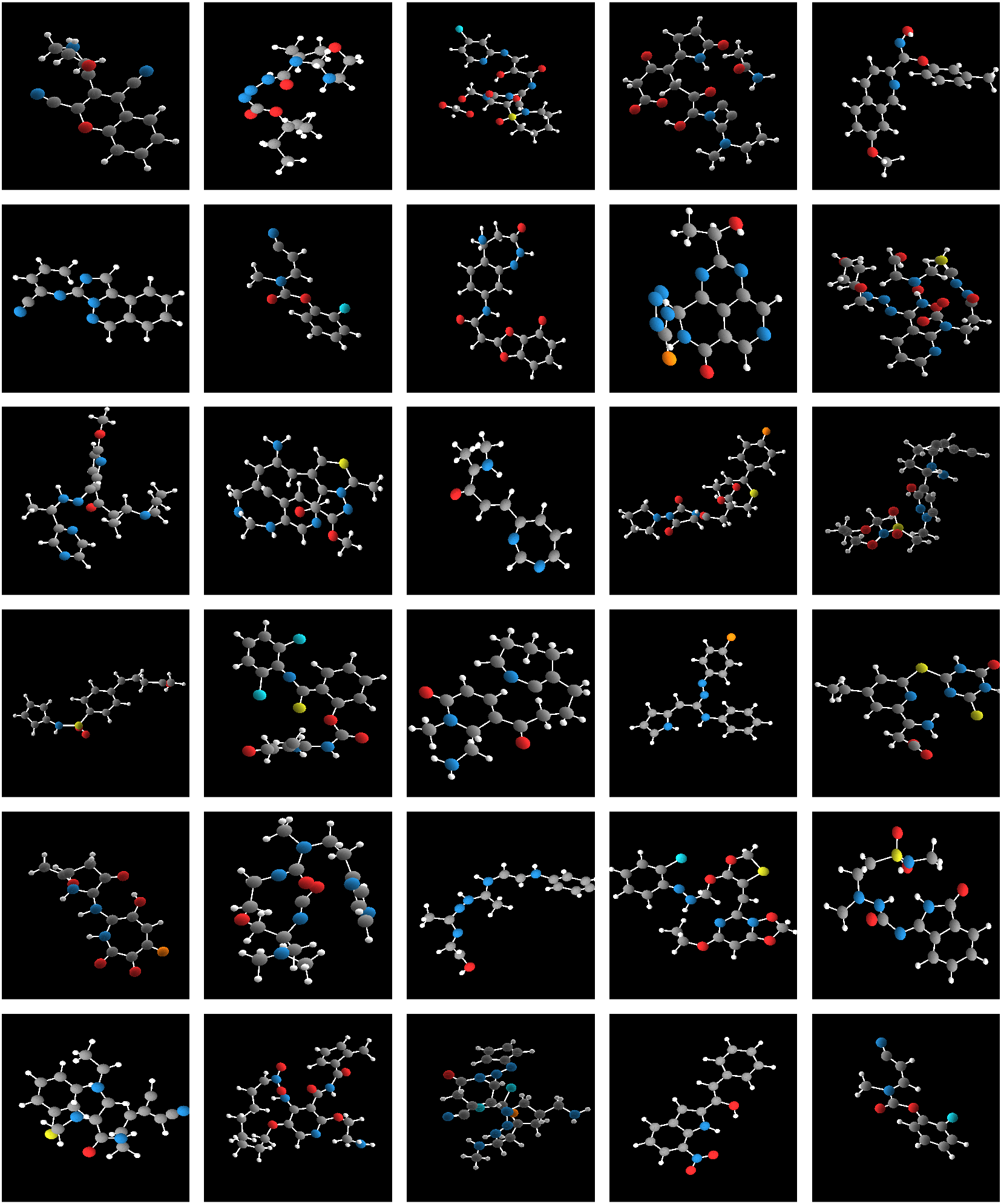}
    \caption{Random samples from GEOM-Drugs~\cite{geom-drug} generations.}
    \label{fig:gd_samples}
\end{figure*}

As shown in Figure~\ref{fig:mol_tra_004_compressed}, Figure~\ref{fig:mol_tra_013_compressed}, and Figure~\ref{fig:mol_tra_026_compressed}, SensDiff exhibits a more structured denoising trajectory: atoms gradually coalesce into chemically valid arrangements, with bonds separating from early steps onward. In contrast, several baselines remain cluttered or exhibit overlapping atoms at intermediate steps, followed by abrupt rearrangements near the final step. We also show 30 randomly picked molecules from QM9 and GEOM-Drugs generations in Figure~\ref{fig:qm9_samples} and Figure~\ref{fig:gd_samples}, respectively.

\vspace{1\baselineskip}

\begin{table*}[h]
\caption{Unconditional 3D molecule generation results on QM9~\cite{qm9}. Validity, uniqueness, novelty, and stability are reported as percentages (\%). Geometric metrics (\textit{e.g.,} bond length, bond angle, and dihedral angle) are reported as errors. \textbf{Best} results are in bold.}
\label{tab:unconditional_qm9_full}
{\setlength{\heavyrulewidth}{0.03em}
\setlength{\lightrulewidth}{0.03em}
\setlength{\cmidrulewidth}{0.03em}
\setlength{\arrayrulewidth}{0.03em}
\resizebox{1.0\textwidth}{!}{
\centering
\large
\renewcommand{\arraystretch}{1.10}
\begin{tabular}{lcccccccc}
\toprule
\multirow{2}{*}{\raisebox{-0.6ex}{\textsc{Model}}} &
\multicolumn{5}{c}{\textsc{Generation Quality} ($\uparrow$)} &
\multicolumn{3}{c}{\textsc{Geometric Fidelity} ($\downarrow$)} \\
\cmidrule(lr){2-6}\cmidrule(lr){7-9}
& \textsc{Valid} & \textsc{Valid \& Unique} & \textsc{Novelty} & \textsc{Atom Sta} & \textsc{Mol Sta} &
\textsc{Bond Length} & \textsc{Bond Angle} & \textsc{Dihedral Angle} \\
\midrule
EDM~\cite{edm}       & 87.9 & 96.0 & 63.0 & 95.5 & 78.4 & 0.13600 & 0.01880 & 0.00069 \\
EEGSDE~\cite{eegsde} & 89.8 & 94.3 & 61.3 & 95.7 & 80.3 & 0.13200 & 0.01820 & 0.00068 \\
GEOLDM~\cite{geoldm} & 89.8 & 95.5 & 54.4 & 94.2 & 86.2 & 0.25200 & 0.01040 & 0.00068 \\
EquiFM~\cite{equifm} & 91.6 & 94.7 & 58.8 & 94.4 & 84.3 & 0.19500 & 0.00986 & 0.00060 \\
GeoBFN~\cite{geobfn} & 90.7 & 93.3 & 63.6 & 94.9 & 86.9 & 0.16800 & 0.00897 & 0.00054 \\
GCDM~\cite{gcdm}     & 91.4 & 95.7 & 56.8 & 94.2 & 82.5 & 0.20000 & 0.01010 & 0.00062 \\
RADM~\cite{radm}     & 90.0 & 92.2 & 59.9 & 95.9 & 84.7 & 0.18500 & 0.00939 & 0.00058 \\
SLDM~\cite{sldm}     & 92.1 & 89.1 & 71.3 & 95.0 & 90.9 & 0.14200 & 0.00840 & 0.00040 \\
UAE-3D~\cite{uae-3d} & 94.4 & 94.4 & 92.5 & 96.2 & 95.4 & 0.07320 & 0.01020 & 0.00036 \\
UniGEM~\cite{unigem} & 92.0 & 93.5 & 67.0 & 95.2 & 87.1 & 0.16200 & 0.00875 &  0.00049 \\
\rowcolor[RGB]{225, 230, 245} SensDiff (Ours) &
\textbf{96.9} & \textbf{97.2} & \textbf{92.6} & \textbf{98.1} & \textbf{95.6} &
\textbf{0.01470} & \textbf{0.00705} & \textbf{0.00018} \\
\bottomrule
\end{tabular}
}}
\vspace{-1cm}
\end{table*}

\vspace{3\baselineskip}

\begin{table*}[h]
\caption{Unconditional 3D molecule generation results on GEOM-Drugs~\cite{geom-drug}. Validity, uniqueness, novelty, and stability are reported as percentages (\%). Geometric metrics (\textit{e.g.,} bond length, bond angle, and dihedral angle) are reported as errors. \textbf{Best} results are in bold.}
\label{tab:unconditional_gd_full}
{\setlength{\heavyrulewidth}{0.03em}
\setlength{\lightrulewidth}{0.03em}
\setlength{\cmidrulewidth}{0.03em}
\setlength{\arrayrulewidth}{0.03em}
\resizebox{1.0\textwidth}{!}{
\centering
\large
\renewcommand{\arraystretch}{1.10}
\begin{tabular}{lcccccccc}
\toprule
\multirow{2}{*}{\raisebox{-0.4ex}{\textsc{Model}}} &
\multicolumn{5}{c}{\textsc{Generation Quality} ($\uparrow$)} &
\multicolumn{3}{c}{\textsc{Geometric Fidelity} ($\downarrow$)} \\
\cmidrule(lr){2-6}\cmidrule(lr){7-9}
& \textsc{Valid} & \textsc{Valid \& Unique} & \textsc{Novelty} & \textsc{Atom Sta} & \textsc{Mol Sta} &
\textsc{Bond Length} & \textsc{Bond Angle} & \textsc{Dihedral Angle} \\
\midrule
EDM~\cite{edm}       & 88.8 & 96.6 & 89.0 & 77.5 & 41.5 & 0.44700 & 0.51200 & 0.01520 \\
EEGSDE~\cite{eegsde} & 89.1 & 94.0 & 87.2 & 79.1 & 46.5 & 0.42700 & 0.49300 & 0.01440 \\
GEOLDM~\cite{geoldm} & 95.6 & 95.6 & 90.5 & 80.7 & 49.3 & 0.40700 & 0.43500 & 0.01750 \\
EquiFM~\cite{equifm} & 95.7 & 94.9 & 88.5 & 80.3 & 47.5 & 0.41700 & 0.45600 & 0.01640 \\
GeoBFN~\cite{geobfn} & 88.5 & 96.4 & 88.1 & 81.5 & 52.2 & 0.38300 & 0.41300 & 0.01370 \\
GCDM~\cite{gcdm}     & 91.2 & 96.3 & 91.7 & 84.5 & 58.8 & 0.30000 & 0.29200 & 0.00884 \\
RADM~\cite{radm}     & 95.7 & 94.9 & 89.6 & 82.1 & 53.2 & 0.36600 & 0.40000 & 0.01310 \\
SLDM~\cite{sldm}     & 95.8 & 95.3 & 89.7 & 84.8 & 62.0 & 0.24900 & 0.23500 & 0.00678 \\
UAE-3D~\cite{uae-3d} & 84.4 & 94.4 & 92.8 & 82.5 & 66.8 & \textbf{0.01040} & 0.00529 & \textbf{0.00018} \\
UniGEM~\cite{unigem} & 94.7 & 95.1 & 90.4 & 82.1 & 54.2 & 0.35300 & 0.39000 & 0.01180 \\
\rowcolor[RGB]{225, 230, 245} SensDiff (Ours) &
\textbf{97.2} & \textbf{98.8} & \textbf{91.8} & \textbf{87.9} & \textbf{79.4} &
0.01470 & \textbf{0.00505} & \textbf{0.00018} \\
\bottomrule
\end{tabular}
}}
\vspace{-1cm}
\end{table*}

\newpage

\newpage

\newpage

\newpage

\newpage

\clearpage
\section{Conditional 3D Molecule Generation with Quantum-Chemical Properties}
\label{sec:append_conditional_generation_quantum}

As reported in Table~\ref{tab:quantum_val_qm9}, SensDiff achieves the highest molecule stability and validity across all six conditional generation tasks. The gains are especially significant in molecule stability, where SensDiff surpasses the strongest baseline SLDM~\cite{sldm} by up to 2.1\% on $\mu$ and 1.9\% on $\Delta\varepsilon$, suggesting that the explicit physicochemical constraints facilitate the model to avoid generating chemically implausible structures. These results confirm that enforcing geometric rules during training not only improves property-level accuracy but also enhances the overall structural quality of the generated molecules.

\begin{figure*}[t]
    \centering
    \includegraphics[width=1.0\textwidth]{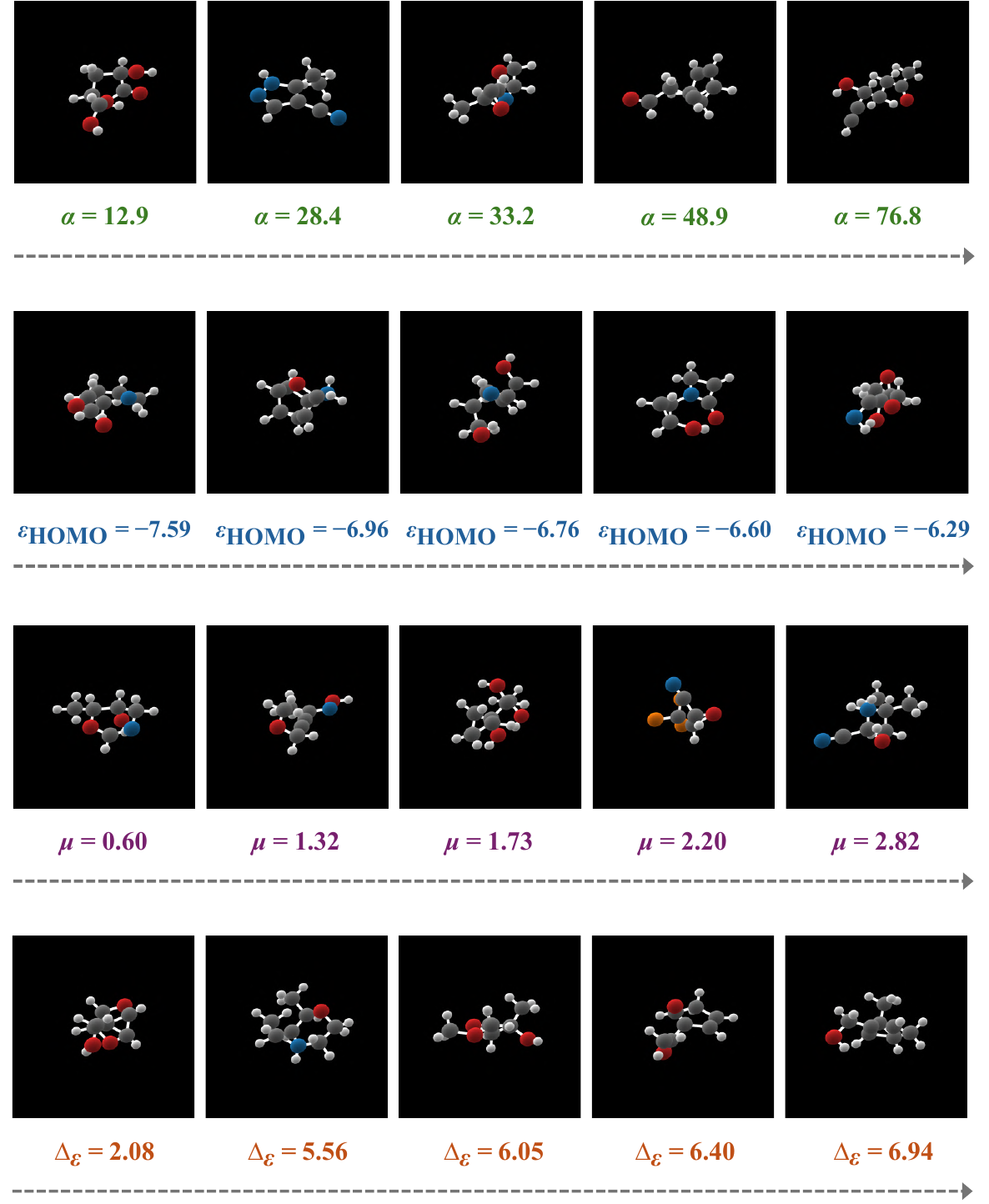}
    \caption{Conditional generation on QM9~\cite{qm9} targeting one quantum property.}
    \label{fig:single_appendix}
\end{figure*}

\begin{figure*}[t]
    \centering
    \includegraphics[width=1.0\textwidth]{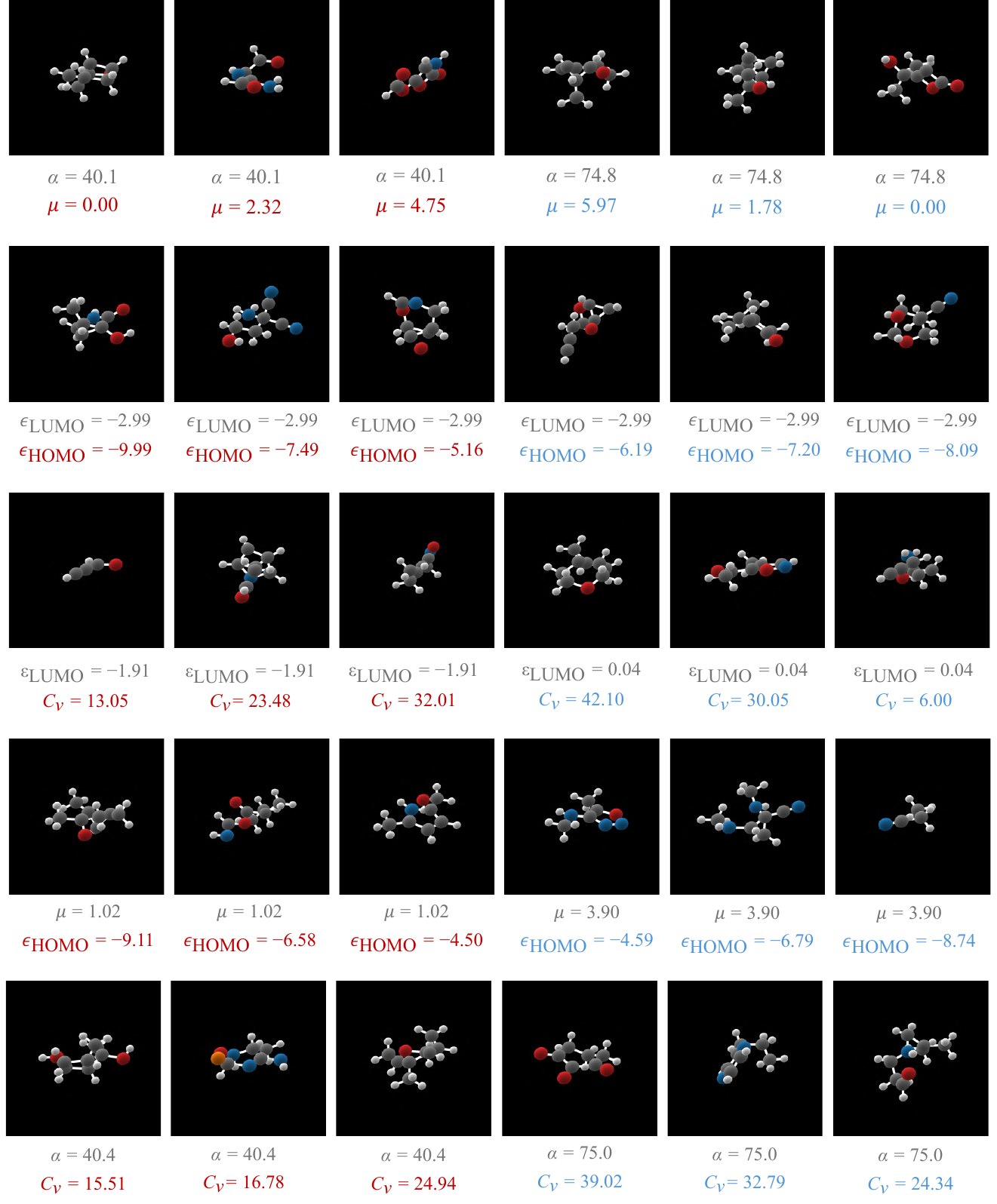}
    \caption{Conditional generation on QM9~\cite{qm9} targeting two quantum properties simultaneously (increasing or decreasing each).}
    \label{fig:double_appendix}
\end{figure*}

As shown in Table~\ref{tab:quantum_two_prop_edm}--\ref{tab:quantum_two_prop_sensdiff}, SensDiff achieves the lowest MAE on every property pair while simultaneously obtaining the highest molecule stability and validity, confirming that physicochemical plausibility and property accuracy are complementary rather than competing objectives. The improvements are most pronounced for geometry-sensitive electronic properties such as $\varepsilon_{\mathrm{LUMO}}$ and $\Delta\varepsilon$, where MAE drops by over 30\% relative to the strongest baseline (Table~\ref{tab:quantum_two_prop_unigem}), suggesting that explicit bond-length and valence constraints yield more physically realistic geometries from which accurate electronic structure naturally follows. Under dual-property conditioning, a more challenging setting that tempts the model to distort geometry to satisfy two targets simultaneously, SensDiff still maintains validity above 0.925 across all pairs (Table~\ref{tab:quantum_two_prop_sensdiff}), whereas baselines frequently fall below 0.90 (Table~\ref{tab:quantum_two_prop_edm} and~\ref{tab:quantum_two_prop_eegsde}), indicating that symbolic constraints effectively regularize against invalid structural deformation. The consistent gains over EEGSDE~\cite{eegsde} (Table~\ref{tab:quantum_two_prop_eegsde}), which also leverages constraints during sampling, further demonstrate that embedding domain knowledge into training is more effective than purely post-hoc gradient guidance, as the model internalizes geometric priors rather than correcting an already imperfect score function. Finally, the uniform improvements across thermodynamic ($C_v$), electronic ($\varepsilon_{\mathrm{HOMO}}$, $\varepsilon_{\mathrm{LUMO}}$), and electrostatic ($\mu$) properties indicate that the benefit is not property-specific but stems from a fundamentally improved molecular representation. We also visualize molecules generated under quantum-property control, with single-property results in Figure~\ref{fig:single_appendix} and double-property results in Figure~\ref{fig:double_appendix}.

\begin{table*}[h]
\caption{Validity and molecule stability (\%, $\uparrow$) for quantum-property conditional generation on QM9~\cite{qm9}. \textbf{Best} results are marked in bold.}
\label{tab:quantum_val_qm9}
{\setlength{\heavyrulewidth}{0.03em}
\setlength{\lightrulewidth}{0.03em}
\setlength{\cmidrulewidth}{0.03em}
\setlength{\arrayrulewidth}{0.03em}
\resizebox{1.0\textwidth}{!}{
\centering
\Large
\renewcommand{\arraystretch}{1.3}
\begin{tabular}{lcccccccccccc}
\toprule
\multirow{2}{*}{\textsc{Model}} &
\multicolumn{2}{c}{$C_v$ \(\left(\frac{\text{cal}}{\text{mol}}\,\text{K}\right)\)} &
\multicolumn{2}{c}{$\mu$ (D)} &
\multicolumn{2}{c}{$\alpha$ (Bohr$^3$)} &
\multicolumn{2}{c}{$\Delta \varepsilon$ (meV)} &
\multicolumn{2}{c}{$\varepsilon_{\mathrm{HOMO}}$ (meV)} &
\multicolumn{2}{c}{$\varepsilon_{\mathrm{LUMO}}$ (meV)} \\
\cmidrule(lr){2-3}\cmidrule(lr){4-5}\cmidrule(lr){6-7}\cmidrule(lr){8-9}\cmidrule(lr){10-11}\cmidrule(lr){12-13}
& \textsc{Mol Sta} & \textsc{Valid}
& \textsc{Mol Sta} & \textsc{Valid}
& \textsc{Mol Sta} & \textsc{Valid}
& \textsc{Mol Sta} & \textsc{Valid}
& \textsc{Mol Sta} & \textsc{Valid}
& \textsc{Mol Sta} & \textsc{Valid} \\
\midrule
EDM~\cite{edm}       & 79.6 & 86.4 & 83.4 & 88.9 & 80.3 & 87.1 & 82.7 & 89.2 & 77.8 & 84.8 & 81.9 & 88.3 \\
EEGSDE~\cite{eegsde} & 81.1 & 87.8 & 84.7 & 90.1 & 82.4 & 88.4 & 84.1 & 89.6 & 79.9 & 86.9 & 83.6 & 89.2 \\
GEOLDM~\cite{geoldm} & 82.5 & 88.1 & 85.9 & 91.0 & 83.7 & 89.4 & 85.3 & 90.2 & 80.9 & 87.4 & 84.4 & 89.8 \\
EquiFM~\cite{equifm} & 82.2 & 87.9 & 85.6 & 90.7 & 83.4 & 89.1 & 84.9 & 89.9 & 80.6 & 87.2 & 84.1 & 89.6 \\
GeoBFN~\cite{geobfn} & 81.8 & 87.6 & 85.1 & 90.2 & 82.9 & 88.8 & 84.6 & 89.6 & 80.2 & 87.1 & 83.9 & 89.1 \\
GCDM~\cite{gcdm}     & 80.9 & 87.3 & 84.2 & 89.3 & 81.9 & 87.9 & 83.7 & 89.1 & 79.6 & 86.8 & 82.9 & 88.6 \\
RADM~\cite{radm}     & 81.9 & 87.7 & 84.9 & 89.9 & 82.8 & 88.7 & 84.6 & 89.9 & 80.4 & 87.2 & 83.7 & 89.4 \\
SLDM~\cite{sldm}     & 83.4 & 88.9 & 86.7 & 91.2 & 84.6 & 89.8 & 86.2 & 90.9 & 81.6 & 87.9 & \textbf{85.4} & 90.3 \\
UAE-3D~\cite{uae-3d} & 82.8 & \textbf{90.2} & 87.4 & 91.8 & 86.2 & 88.2 & 86.5 & 89.4	 & 83.0 & 86.7	& 84.9 & \textbf{90.6} \\
UniGEM~\cite{unigem} & 82.9 & 88.3 & 86.2 & 90.8 & 84.1 & 89.2 & 85.6 & 90.1 & 81.2 & 87.6 & 84.8 & 89.7 \\
\rowcolor[RGB]{225, 230, 245} SensDiff (Ours) &
\textbf{83.6} & 89.8 & \textbf{88.8} & \textbf{92.1} & \textbf{86.5} & \textbf{89.7} & \textbf{88.1} & \textbf{91.8} & \textbf{83.0} & \textbf{88.4} & 85.3 & 90.5 \\
\bottomrule
\end{tabular}
}}
\end{table*}

\vspace{1\baselineskip}

\begin{table*}[h]
\caption{Double-property conditional generation on QM9~\cite{qm9} with EDM~\cite{edm}.}
\label{tab:quantum_two_prop_edm}
\centering
\scriptsize
\setlength{\tabcolsep}{6pt}
\renewcommand{\arraystretch}{1.4}
\begin{tabular*}{\textwidth}{@{\extracolsep{\fill}} l c c c l c c c}
\toprule
\multicolumn{4}{c}{\textsc{Property 1}} & \multicolumn{4}{c}{\textsc{Property 2}} \\
\cmidrule(lr){1-4}\cmidrule(lr){5-8}
\textsc{Metric} & \textsc{MAE} $\downarrow$ & \textsc{Mol Sta} $\uparrow$ & \textsc{Valid} $\uparrow$ &
\textsc{Metric} & \textsc{MAE} $\downarrow$ & \textsc{Mol Sta} $\uparrow$ & \textsc{Valid} $\uparrow$ \\
\midrule
$C_v$ \(\left(\frac{\text{cal}}{\text{mol}}\,\text{K}\right)\) & 1.079 & 0.768 & 0.898 & $\mu$ (D) & 1.156 & 0.768 & 0.898 \\
$\alpha$ (Bohr$^3$) & 2.760 & 0.772 & 0.899 & $\mu$ (D) & 1.158 & 0.772 & 0.899 \\
$\Delta \varepsilon$ (meV) & 683 & 0.759 & 0.896 & $\mu$ (D) & 1.130 & 0.759 & 0.896 \\
$\varepsilon_{\mathrm{HOMO}}$ (meV) & 371 & 0.760 & 0.895 & $\mu$ (D) & 1.041 & 0.760 & 0.895 \\
$\varepsilon_{\mathrm{HOMO}}$ (meV) & 364 & 0.748 & 0.892 & $\varepsilon_{\mathrm{LUMO}}$ (meV) & 612 & 0.748 & 0.892 \\
$\varepsilon_{\mathrm{HOMO}}$ (meV) & 382 & 0.745 & 0.888 & $C_v$ \(\left(\frac{\text{cal}}{\text{mol}}\,\text{K}\right)\) & 1.118 & 0.745 & 0.888 \\
$\varepsilon_{\mathrm{HOMO}}$ (meV) & 376 & 0.749 & 0.890 & $\alpha$ (Bohr$^3$) & 2.880 & 0.749 & 0.890 \\
$\Delta \varepsilon$ (meV) & 702 & 0.754 & 0.894 & $\varepsilon_{\mathrm{LUMO}}$ (meV) & 505 & 0.754 & 0.894 \\
\bottomrule
\end{tabular*}
\end{table*}

\begin{table*}[h]
\caption{Double-property conditional generation on QM9~\cite{qm9} with UniGEM~\cite{unigem}.}
\label{tab:quantum_two_prop_unigem}
\centering
\scriptsize
\setlength{\tabcolsep}{6pt}
\renewcommand{\arraystretch}{1.6}
\begin{tabular*}{\textwidth}{@{\extracolsep{\fill}} l c c c l c c c}
\toprule
\multicolumn{4}{c}{\textsc{Property 1}} & \multicolumn{4}{c}{\textsc{Property 2}} \\
\cmidrule(lr){1-4}\cmidrule(lr){5-8}
\textsc{Metric} & \textsc{MAE} $\downarrow$ & \textsc{Mol Sta} $\uparrow$ & \textsc{Valid} $\uparrow$ &
\textsc{Metric} & \textsc{MAE} $\downarrow$ & \textsc{Mol Sta} $\uparrow$ & \textsc{Valid} $\uparrow$ \\
\midrule
$C_v$ \(\left(\frac{\text{cal}}{\text{mol}}\,\text{K}\right)\) & 0.952 & 0.811 & 0.907 & $\mu$ (D) & 0.876 & 0.811 & 0.907 \\
$\alpha$ (Bohr$^3$) & 2.430 & 0.818 & 0.911 & $\mu$ (D) & 0.838 & 0.818 & 0.911 \\
$\Delta \varepsilon$ (meV) & 544 & 0.796 & 0.903 & $\mu$ (D) & 0.856 & 0.796 & 0.903 \\
$\varepsilon_{\mathrm{HOMO}}$ (meV) & 246 & 0.829 & 0.909 & $\mu$ (D) & 0.848 & 0.829 & 0.909 \\
$\varepsilon_{\mathrm{HOMO}}$ (meV) & 252 & 0.822 & 0.905 & $\varepsilon_{\mathrm{LUMO}}$ (meV) & 556 & 0.822 & 0.905 \\
$\varepsilon_{\mathrm{HOMO}}$ (meV) & 264 & 0.808 & 0.898 & $C_v$ \(\left(\frac{\text{cal}}{\text{mol}}\,\text{K}\right)\) & 0.926 & 0.808 & 0.898 \\
$\varepsilon_{\mathrm{HOMO}}$ (meV) & 259 & 0.814 & 0.901 & $\alpha$ (Bohr$^3$) & 2.490 & 0.814 & 0.901 \\
$\Delta \varepsilon$ (meV) & 548 & 0.801 & 0.900 & $\varepsilon_{\mathrm{LUMO}}$ (meV) & 482 & 0.801 & 0.900 \\
\bottomrule
\end{tabular*}
\end{table*}

\begin{table*}[h]
\caption{Double-property conditional generation on QM9~\cite{qm9} with EEGSDE~\cite{eegsde}.}
\label{tab:quantum_two_prop_eegsde}
\centering
\scriptsize
\setlength{\tabcolsep}{6pt}
\renewcommand{\arraystretch}{1.6}
\begin{tabular*}{\textwidth}{@{\extracolsep{\fill}} l c c c l c c c}
\toprule
\multicolumn{4}{c}{\textsc{Property 1}} & \multicolumn{4}{c}{\textsc{Property 2}} \\
\cmidrule(lr){1-4}\cmidrule(lr){5-8}
\textsc{Metric} & \textsc{MAE} $\downarrow$ & \textsc{Mol Sta} $\uparrow$ & \textsc{Valid} $\uparrow$ &
\textsc{Metric} & \textsc{MAE} $\downarrow$ & \textsc{Mol Sta} $\uparrow$ & \textsc{Valid} $\uparrow$ \\
\midrule
$C_v$ \(\left(\frac{\text{cal}}{\text{mol}}\,\text{K}\right)\) & 0.981 & 0.754 & 0.892 & $\mu$ (D) & 0.912 & 0.754 & 0.892 \\
$\alpha$ (Bohr$^3$) & 2.610 & 0.770 & 0.897 & $\mu$ (D) & 0.855 & 0.770 & 0.897 \\
$\Delta \varepsilon$ (meV) & 563 & 0.727 & 0.890 & $\mu$ (D) & 0.866 & 0.727 & 0.890 \\
$\varepsilon_{\mathrm{HOMO}}$ (meV) & 328 & 0.752 & 0.889 & $\mu$ (D) & 0.894 & 0.752 & 0.889 \\
$\varepsilon_{\mathrm{HOMO}}$ (meV) & 319 & 0.741 & 0.887 & $\varepsilon_{\mathrm{LUMO}}$ (meV) & 536 & 0.741 & 0.887 \\
$\varepsilon_{\mathrm{HOMO}}$ (meV) & 341 & 0.739 & 0.882 & $C_v$ \(\left(\frac{\text{cal}}{\text{mol}}\,\text{K}\right)\) & 1.006 & 0.739 & 0.882 \\
$\varepsilon_{\mathrm{HOMO}}$ (meV) & 333 & 0.745 & 0.885 & $\alpha$ (Bohr$^3$) & 2.670 & 0.745 & 0.885 \\
$\Delta \varepsilon$ (meV) & 587 & 0.736 & 0.888 & $\varepsilon_{\mathrm{LUMO}}$ (meV) & 458 & 0.736 & 0.888 \\
\bottomrule
\end{tabular*}
\end{table*}

\begin{table*}[h]
\caption{Double-property conditional generation on QM9~\cite{qm9} with SensDiff (Ours).}
\label{tab:quantum_two_prop_sensdiff}
\centering
\scriptsize
\setlength{\tabcolsep}{6pt}
\renewcommand{\arraystretch}{1.6}
\begin{tabular*}{\textwidth}{@{\extracolsep{\fill}} l c c c l c c c}
\toprule
\multicolumn{4}{c}{\textsc{Property 1}} & \multicolumn{4}{c}{\textsc{Property 2}} \\
\cmidrule(lr){1-4}\cmidrule(lr){5-8}
\textsc{Metric} & \textsc{MAE} $\downarrow$ & \textsc{Mol Sta} $\uparrow$ & \textsc{Valid} $\uparrow$ &
\textsc{Metric} & \textsc{MAE} $\downarrow$ & \textsc{Mol Sta} $\uparrow$ & \textsc{Valid} $\uparrow$ \\
\midrule
$C_v$ \(\left(\frac{\text{cal}}{\text{mol}}\,\text{K}\right)\) & 0.672 & 0.846 & 0.934 & $\mu$ (D) & 0.721 & 0.846 & 0.934 \\
$\alpha$ (Bohr$^3$) & 1.920 & 0.855 & 0.938 & $\mu$ (D) & 0.748 & 0.855 & 0.938 \\
$\Delta \varepsilon$ (meV) & 449 & 0.832 & 0.929 & $\mu$ (D) & 0.757 & 0.832 & 0.929 \\
$\varepsilon_{\mathrm{HOMO}}$ (meV) & 219 & 0.864 & 0.936 & $\mu$ (D) & 0.782 & 0.864 & 0.936 \\
$\varepsilon_{\mathrm{HOMO}}$ (meV) & 226 & 0.857 & 0.932 & $\varepsilon_{\mathrm{LUMO}}$ (meV) & 323 & 0.857 & 0.932 \\
$\varepsilon_{\mathrm{HOMO}}$ (meV) & 238 & 0.841 & 0.925 & $C_v$ \(\left(\frac{\text{cal}}{\text{mol}}\,\text{K}\right)\) & 0.752 & 0.841 & 0.925 \\
$\varepsilon_{\mathrm{HOMO}}$ (meV) & 234 & 0.849 & 0.929 & $\alpha$ (Bohr$^3$) & 1.930 & 0.849 & 0.929 \\
$\Delta \varepsilon$ (meV) & 437 & 0.836 & 0.927 & $\varepsilon_{\mathrm{LUMO}}$ (meV) & 368 & 0.836 & 0.927 \\
\bottomrule
\end{tabular*}
\end{table*}

\newpage

\newpage

\clearpage
\section{Conditional 3D Molecule Generation for Multi-Objective Drug Design}
\label{con_multi_drug_design}

\begin{figure*}[h]
    \centering
    \includegraphics[width=1.0\textwidth]{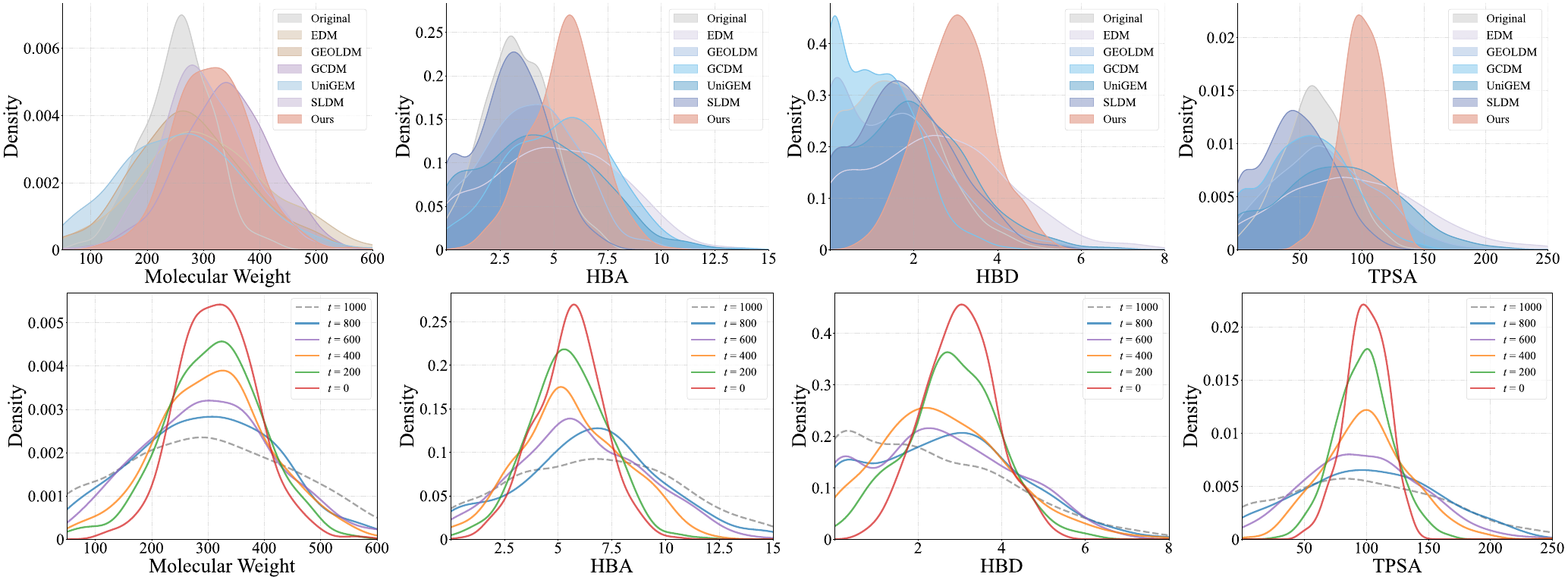}
    \caption{(\textbf{Upper}) The generated molecules precisely satisfy the specified oral drug design constraints (\textit{e.g.,} MW, HBA, HBD, and TPSA). (\textbf{Bottom}) Molecules converge toward the target values across denoising steps.}
    \label{fig:oral_drug_append}
\end{figure*}

\begin{figure*}[h]
    \centering
    \includegraphics[width=1.0\textwidth]{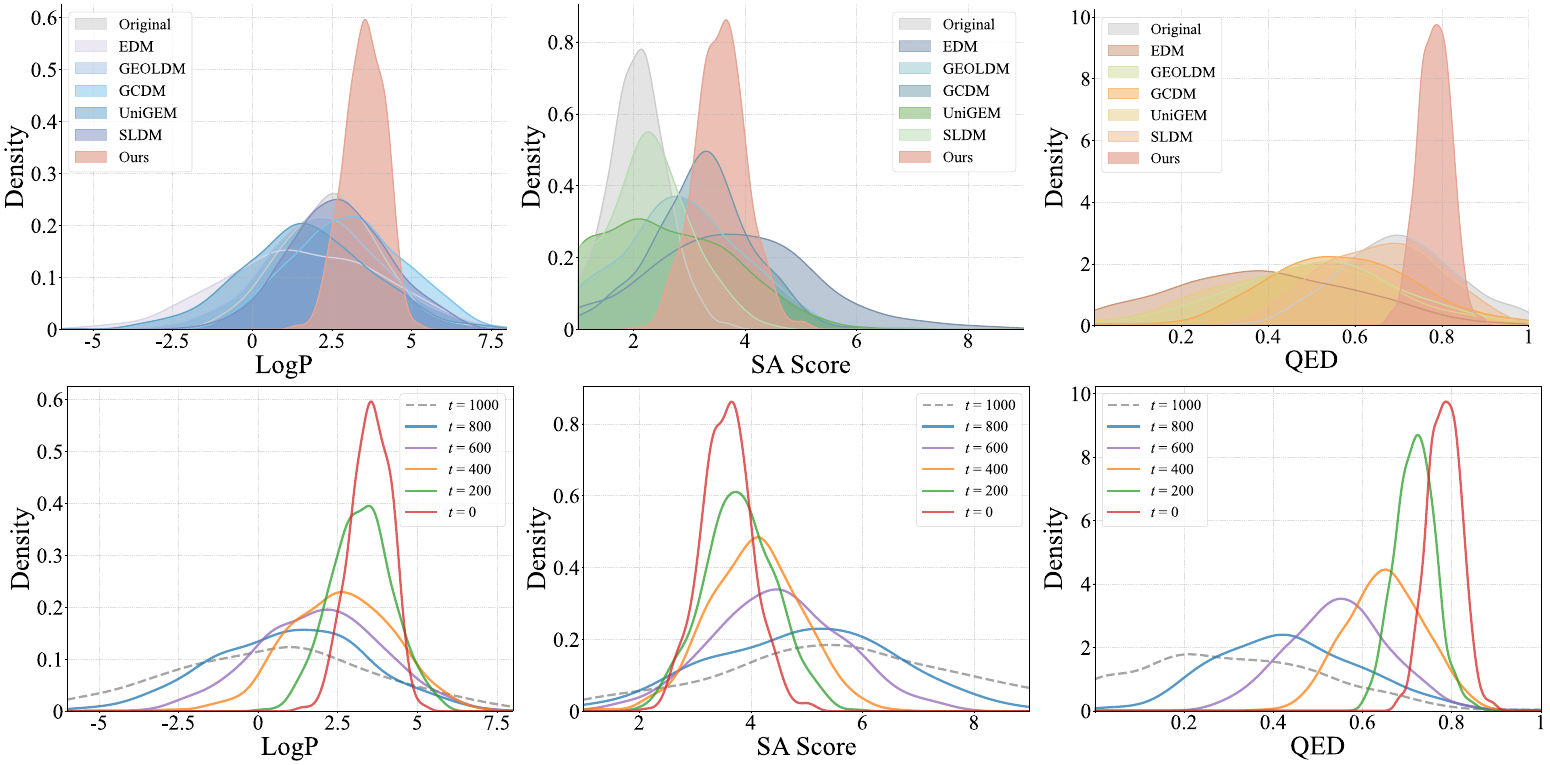}
    \caption{(\textbf{Upper}) The generated molecules precisely satisfy the specified Lipinski’s Rule-of-Five constraints (\textit{e.g.,} LogP, SA Score, and QED). (\textbf{Bottom}) Molecules converge toward the target values across denoising steps.}
    \label{fig:lipinski1_append}
\end{figure*}

\begin{figure*}[h]
    \centering
    \includegraphics[width=1.0\textwidth]{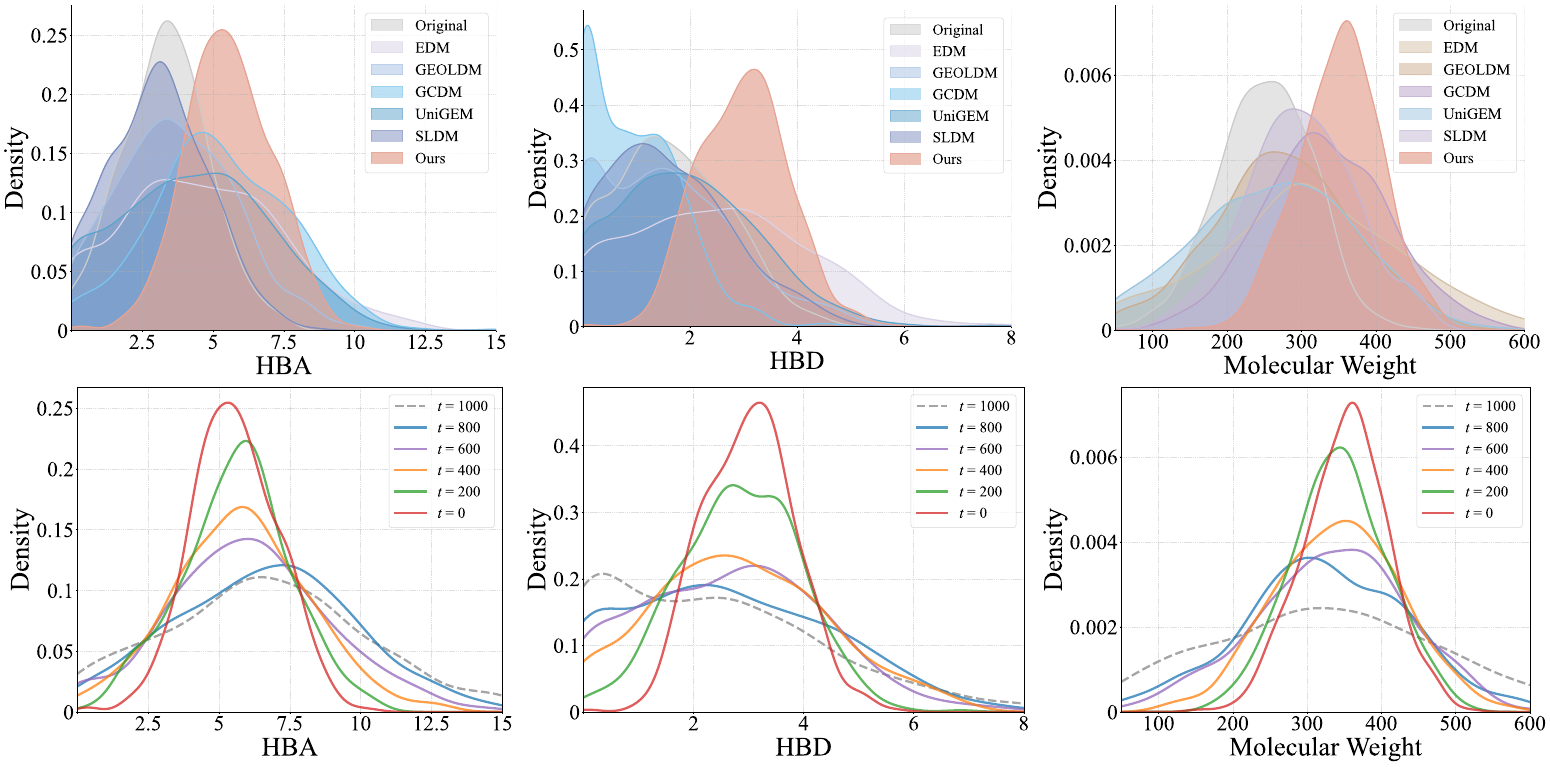}
    \caption{(\textbf{Upper}) The generated molecules precisely satisfy the specified Lipinski’s Rule-of-Five constraint (\textit{e.g.,} HBA, HBD, and MW). (\textbf{Bottom}) Molecules converge toward the target values across denoising steps.}
    \label{fig:lipinski2_append}
\end{figure*}

\begin{figure*}[h]
    \centering
    \includegraphics[width=1.0\textwidth]{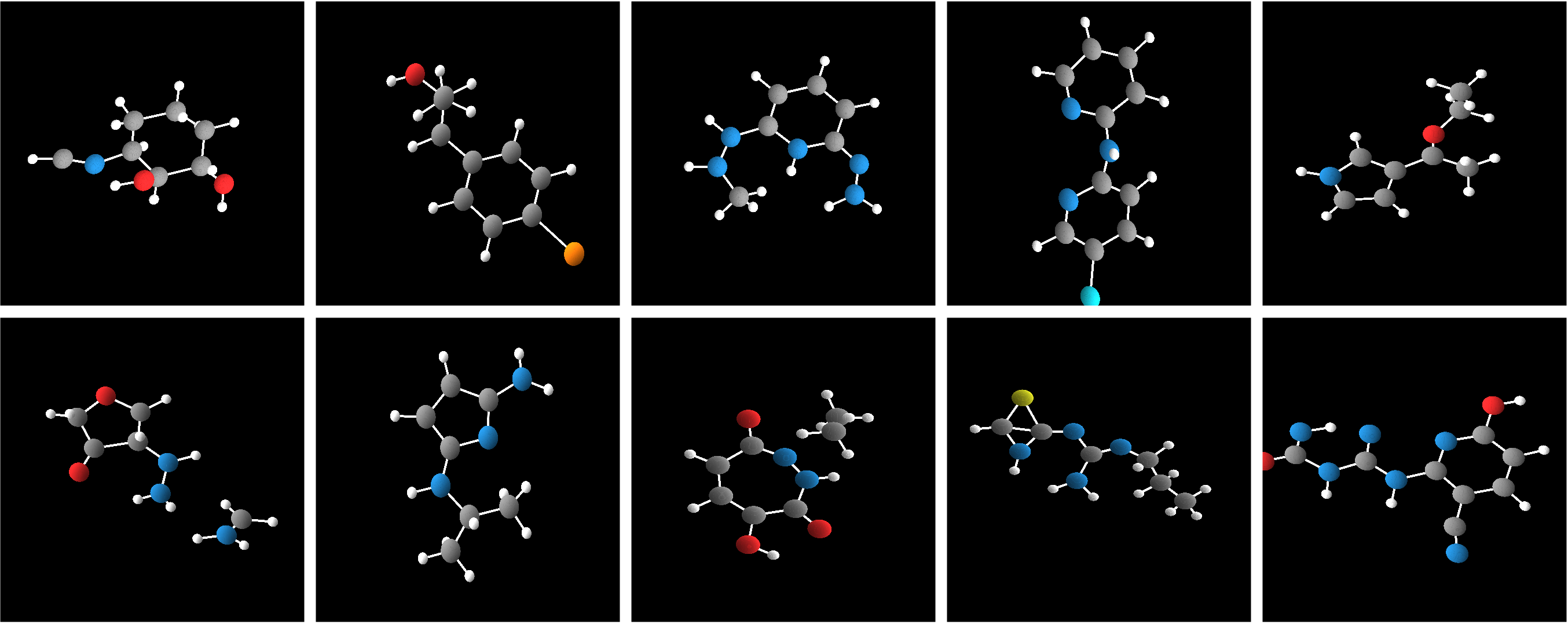}
    \caption{Random samples of molecules satisfying the oral drug design constraints.}
    \label{fig:oral_drug_mol}
\end{figure*}

\begin{figure*}[h]
    \centering
    \includegraphics[width=1.0\textwidth]{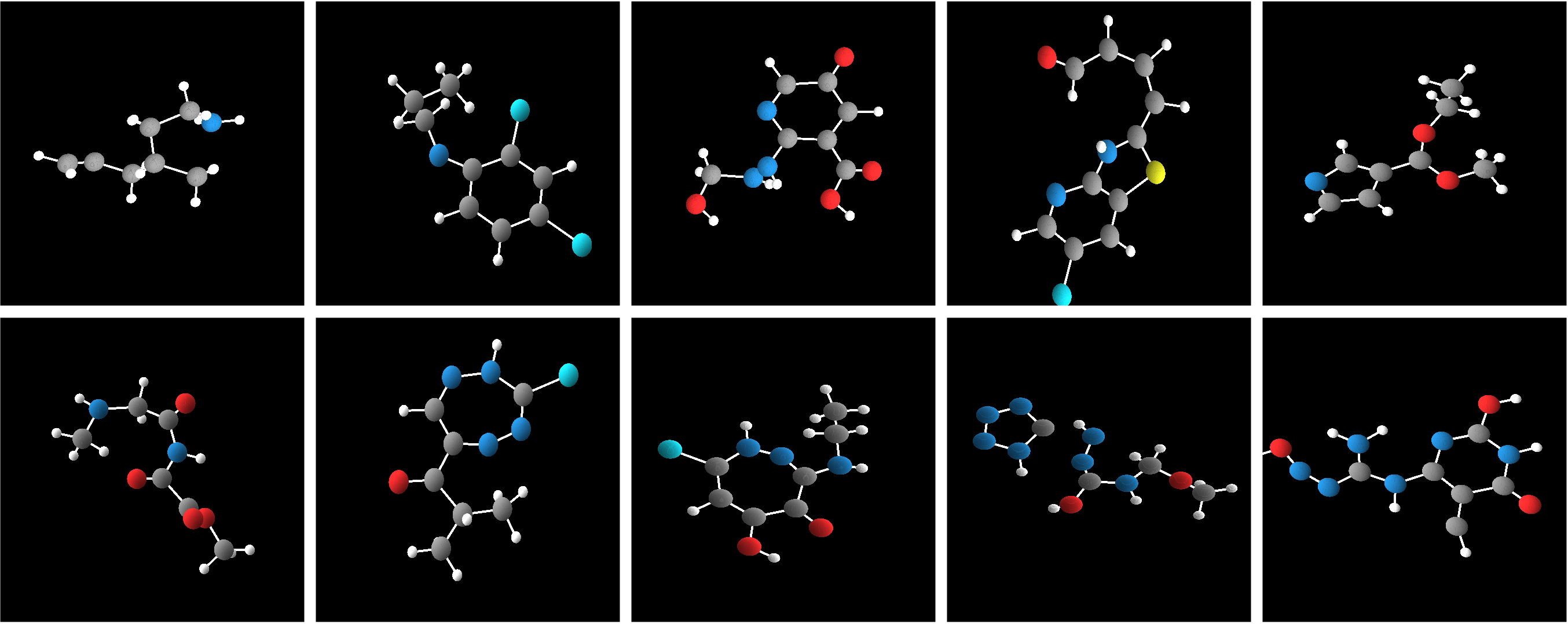}
    \caption{Random samples of molecules satisfying Lipinski’s Rule-of-Five constraints.}
    \label{fig:lipinski_mol}
\end{figure*}

In this task, we consider two settings: (1) oral drug design (detailed in Section~\ref{sec:experiments}) and (2) designing drug under Lipinski’s Rule-of-Five constraints, requiring \(\text{HBA}\le 10\), \(\text{HBD}\le 5\), \(\text{MW}\le 500\,\text{Da}\), \(\text{LogP}\le 5\), \(\text{SA Score}<5\), and \(\text{QED}>0.7\)~\cite{lipinski2004lead, walters2012going}. Figure~\ref{fig:oral_drug_append} shows the distributions and denoising trajectory of MW, HBA, HBD, and TPSA for oral drug design. 
In this study, we impose these constraints only during sampling by formulating them as symbolic energy constraints and applying them to the pretrained model at sampling time. Figure~\ref{fig:lipinski1_append} and Figure~\ref{fig:lipinski2_append} show the distributions and denoising trajectory of LogP/SA Score/QED and HBA/HBD/MW, respectively, for Lipinski-constrained drug design. The property distributions of molecules generated by SensDiff exhibit the sharpest peaks and the closest alignment with the target distributions across all descriptors, whereas baselines such as UniGEM~\cite{unigem} and SLDM~\cite{sldm} produce noticeably broader and shifted distributions. The bottom rows of each figure further reveal that the generated property distributions progressively sharpen from an uniform spread at $t{=}1000$ to a concentrated mode at $t{=}0$, suggesting that the denoising process smoothly steers molecules toward the desired property targets. These results collectively showcase that SensDiff achieves fine-grained property control under both tasks. We also visualize representative generated molecules for both settings in Figures~\ref{fig:oral_drug_mol} and~\ref{fig:lipinski_mol}, respectively.

\clearpage
\section{Additional Ablation Studies}
\label{appendix_ablation}

In this section, we systematically ablate each component of SensDiff. Notably, ``$/$'' indicates the enforcement order (\textit{e.g.,} \(E_{3}/E_{1}\) enforces \(E_{3}\) followed by \(E_{1}\)). Due to space limitations, we denote the four constraints \(E_{\text{clash}}\) (\textit{i.e.,} steric non-overlap), \(E_{\text{bond\_len}}\) (\textit{i.e.,} bond length), \(E_{\text{angle}}\) (\textit{i.e.,} bond angle), and \(E_{\text{val}}\) (\textit{i.e.,} valence validity) as \(E_{1}\), \(E_{2}\), \(E_{3}\), and \(E_{4}\), respectively, following their enforcement order during training. The four columns, \(E_{1}\), \(E_{2}\), \(E_{3}\), and \(E_{4}\), report the constraint satisfaction rates of the four constraints. Generation quality and constraint-compliance metrics are reported as percentages, whereas MAE is reported as a numeric value.

\vspace{1\baselineskip}

\begin{table*}[h]
\caption{Ablation on constraint sets without sequential enforcement (\textit{i.e.,} all constraints are activated simultaneously from the start of training). The diffusion models are trained on QM9~\cite{qm9}.}
\label{tab:ablation_set_qm9}
\centering
\small
\setlength{\tabcolsep}{6pt}
\renewcommand{\arraystretch}{1.25}
\resizebox{\textwidth}{!}{%
\begin{tabular}{l *{8}{c}}
\toprule
\multirow{2}{*}{\raisebox{-0.9ex}{\textsc{Constraint}}} &
\multicolumn{4}{c}{\textsc{Generation Quality}} &
\multicolumn{4}{c}{\textsc{Constraint Satisfaction}} \\
\cmidrule(lr){2-5}\cmidrule(l){6-9}
& \textsc{Valid} &
\textsc{Valid \& Unique} &
\textsc{Atom Stab} &
\textsc{Mol Stab} &
\(E_{1}\) &
\(E_{2}\) &
\(E_{3}\) &
\(E_{4}\) \\
\midrule
\(E_{1}\)                   & 90.9 & 98.4 & 94.8 & 87.0 & 94.8 & 75.4 & 77.2 & 80.5 \\
\(E_{2}\)                   & 92.9 & 98.6 & 94.9 & 86.2 & 78.4 & 96.6 & 81.2 & 84.1 \\
\(E_{3}\)                   & 93.2 & 98.5 & 94.0 & 87.1 & 79.9 & 83.5 & 95.3 & 85.3 \\
\(E_{4}\)                   & 93.5 & 98.6 & 95.3 & 88.8 & 81.7 & 84.5 & 86.2 & 96.9 \\
\(E_{1}, E_{3}\)            & 90.1 & 97.7 & 94.9 & 84.4 & 92.2 & 76.8 & 94.5 & 81.5 \\
\(E_{1}, E_{4}\)            & 91.3 & 97.6 & 92.7 & 83.7 & 92.8 & 78.2 & 82.7 & 95.9 \\
\(E_{1}, E_{2}\)            & 91.7 & 97.9 & 94.6 & 84.3 & 92.3 & 93.6 & 79.2 & 82.6 \\
\(E_{2}, E_{3}\)            & 92.4 & 98.0 & 94.4 & 85.0 & 79.6 & 93.6 & 94.1 & 84.5 \\
\(E_{2}, E_{4}\)            & 92.7 & 98.1 & 94.8 & 85.9 & 81.1 & 94.0 & 83.5 & 95.6 \\
\(E_{3}, E_{4}\)            & 92.5 & 97.9 & 94.5 & 85.7 & 80.5 & 83.9 & 94.2 & 95.8 \\
\(E_{1}, E_{2}, E_{3}\)     & 89.3 & 97.0 & 92.2 & 79.5 & 90.2 & 92.4 & 92.8 & 83.2 \\
\(E_{1}, E_{2}, E_{4}\)     & 89.6 & 97.1 & 92.5 & 80.2 & 90.5 & 92.5 & 84.1 & 93.8 \\
\(E_{2}, E_{3}, E_{4}\)     & 90.6 & 97.3 & 93.0 & 81.8 & 81.8 & 92.7 & 92.9 & 94.0 \\
\(E_{1}, E_{3}, E_{4}\)     & 88.9 & 96.9 & 91.8 & 79.2 & 90.6 & 75.2 & 93.3 & 94.7 \\
\(E_{1}, E_{2}, E_{3}, E_{4}\) & 89.4 & 96.5 & 92.4 & 78.8 & 89.6 & 93.2 & 92.4 & 94.1 \\
\bottomrule
\end{tabular}%
}
\end{table*}

\vspace{1\baselineskip}

\begin{table*}[h]
\caption{Ablation on constraint sets without sequential enforcement (\textit{i.e.,} all constraints are activated simultaneously from the start of training). The diffusion models are trained on GEOM-Drugs~\cite{geom-drug}.}
\label{tab:ablation_set_geom_drugs}
\centering
\small
\setlength{\tabcolsep}{6pt}
\renewcommand{\arraystretch}{1.25}
\resizebox{\textwidth}{!}{%
\begin{tabular}{l *{8}{c}}
\toprule
\multirow{2}{*}{\raisebox{-0.9ex}{\textsc{Constraint}}} &
\multicolumn{4}{c}{\textsc{Generation Quality}} &
\multicolumn{4}{c}{\textsc{Constraint Satisfaction}} \\
\cmidrule(lr){2-5}\cmidrule(l){6-9}
& \textsc{Valid} &
\textsc{Valid \& Unique} &
\textsc{Atom Stab} &
\textsc{Mol Stab} &
\(E_{1}\) &
\(E_{2}\) &
\(E_{3}\) &
\(E_{4}\) \\
\midrule
\(E_{1}\)                  & 91.3 & 98.5 & 83.2 & 68.5 & 91.5 & 70.7 & 72.1 & 74.3 \\
\(E_{2}\)                  & 92.6 & 98.6 & 83.6 & 69.2 & 72.9 & 92.9 & 74.6 & 76.8 \\
\(E_{3}\)                  & 92.8 & 98.5 & 83.0 & 70.0 & 74.1 & 76.8 & 91.7 & 78.2 \\
\(E_{4}\)                  & 93.3 & 98.6 & 84.3 & 72.0 & 75.2 & 77.7 & 79.5 & 93.3 \\
\(E_{1}, E_{3}\)           & 90.2 & 97.8 & 82.8 & 66.0 & 89.3 & 72.2 & 90.5 & 75.3 \\
\(E_{1}, E_{4}\)           & 90.7 & 97.8 & 82.1 & 65.4 & 89.9 & 73.2 & 76.2 & 91.8 \\
\(E_{1}, E_{2}\)           & 91.1 & 97.9 & 83.1 & 65.7 & 88.8 & 90.1 & 73.6 & 76.0 \\
\(E_{2}, E_{3}\)           & 91.8 & 98.0 & 83.2 & 67.2 & 74.3 & 90.2 & 90.3 & 77.6 \\
\(E_{2}, E_{4}\)           & 92.3 & 98.1 & 83.8 & 68.8 & 75.5 & 90.7 & 77.2 & 92.0 \\
\(E_{3}, E_{4}\)           & 92.1 & 98.0 & 83.5 & 68.5 & 74.9 & 77.8 & 90.4 & 92.2 \\
\(E_{1}, E_{2}, E_{3}\)    & 88.8 & 97.1 & 81.1 & 61.7 & 86.9 & 88.9 & 89.2 & 76.8 \\
\(E_{1}, E_{2}, E_{4}\)    & 89.2 & 97.2 & 81.7 & 63.0 & 87.2 & 89.2 & 77.9 & 90.2 \\
\(E_{2}, E_{3}, E_{4}\)    & 90.1 & 97.3 & 82.2 & 64.7 & 76.2 & 89.4 & 89.4 & 90.4 \\
\(E_{1}, E_{3}, E_{4}\)    & 88.5 & 97.1 & 80.7 & 61.3 & 87.2 & 70.6 & 89.4 & 90.7 \\
\(E_{1}, E_{2}, E_{3}, E_{4}\) & 89.0 & 96.9 & 81.3 & 60.8 & 86.3 & 89.3 & 88.5 & 90.0 \\
\bottomrule
\end{tabular}%
}
\end{table*}

\begin{table*}[h]
\caption{Ablation on constraint enforcement order (sequential constraint activation). The diffusion models are trained on QM9~\cite{qm9}.}
\label{tab:constraint_order_ablation}
\centering
\small
\setlength{\tabcolsep}{5pt}
\renewcommand{\arraystretch}{1.4}
\resizebox{\textwidth}{!}{%
\begin{tabular}{l *{8}{c}}
\toprule
\multirow{2}{*}{\raisebox{-0.9ex}{\textsc{Constraint}}} &
\multicolumn{4}{c}{\textsc{Generation Quality}} &
\multicolumn{4}{c}{\textsc{Constraint Satisfaction}} \\
\cmidrule(lr){2-5}\cmidrule(l){6-9}
& \textsc{Valid} &
\textsc{Valid \& Unique} &
\textsc{Atom Stab} &
\textsc{Mol Stab} &
\(E_{1}\) &
\(E_{2}\) &
\(E_{3}\) &
\(E_{4}\) \\
\midrule
\(E_{3}/E_{1}\)             & 86.5 & 97.9 & 86.2 & 75.2 & 88.1 & 78.4 & 92.1 & 82.6 \\
\(E_{4}/E_{1}\)             & 82.7 & 97.3 & 84.6 & 69.6 & 89.8 & 77.2 & 80.4 & 92.7 \\
\(E_{3}/E_{2}\)             & 90.2 & 98.8 & 89.8 & 81.3 & 82.9 & 95.5 & 91.9 & 84.9 \\
\(E_{4}/E_{2}\)             & 87.3 & 98.2 & 87.2 & 77.4 & 82.5 & 95.2 & 83.8 & 93.9 \\
\(E_{1}/E_{2}\)             & 84.2 & 97.5 & 85.4 & 71.2 & 90.2 & 90.3 & 79.2 & 82.0 \\
\(E_{1}/E_{3}\)             & 85.9 & 97.7 & 86.0 & 73.6 & 89.5 & 77.8 & 91.5 & 82.2 \\
\(E_{1}/E_{4}\)             & 83.4 & 97.4 & 85.1 & 70.3 & 90.6 & 76.6 & 80.2 & 91.9 \\
\(E_{2}/E_{3}\)             & 89.0 & 98.6 & 88.9 & 79.7 & 81.7 & 94.8 & 91.4 & 84.2 \\
\(E_{2}/E_{4}\)             & 86.3 & 98.1 & 86.7 & 76.2 & 81.6 & 94.4 & 82.7 & 93.1 \\
\(E_{3}/E_{4}\)             & 87.8 & 98.2 & 88.1 & 78.0 & 82.1 & 83.3 & 91.7 & 92.7 \\
\(E_{4}/E_{3}\)             & 89.3 & 98.7 & 89.3 & 80.4 & 82.9 & 84.4 & 91.6 & 94.4 \\
\(E_{2}/E_{1}\)             & 84.7 & 97.7 & 85.7 & 72.5 & 88.5 & 90.6 & 79.0 & 81.8 \\
\(E_{3}/E_{2}/E_{1}\)       & 88.5 & 98.5 & 88.7 & 78.4 & 88.1 & 92.3 & 91.4 & 83.8 \\
\(E_{4}/E_{2}/E_{1}\)       & 87.2 & 98.3 & 88.0 & 76.6 & 88.2 & 92.3 & 82.6 & 93.2 \\
\(E_{4}/E_{3}/E_{1}\)       & 88.0 & 98.3 & 88.5 & 77.3 & 88.5 & 83.5 & 91.4 & 93.2 \\
\(E_{4}/E_{3}/E_{2}\)       & 90.2 & 98.8 & 90.3 & 81.9 & 83.7 & 92.1 & 92.3 & 94.5 \\
\(E_{3}/E_{4}/E_{2}\)       & 89.5 & 98.7 & 89.6 & 80.7 & 83.0 & 91.8 & 91.7 & 93.6 \\
\(E_{3}/E_{1}/E_{4}\)       & 80.3 & 96.6 & 84.3 & 66.8 & 86.4 & 81.3 & 90.3 & 88.7 \\
\(E_{3}/E_{1}/E_{2}\)       & 82.6 & 97.0 & 85.1 & 70.5 & 86.3 & 88.0 & 90.6 & 83.3 \\
\(E_{4}/E_{1}/E_{3}\)       & 84.3 & 97.5 & 85.9 & 72.4 & 87.3 & 82.5 & 89.8 & 92.0 \\
\(E_{4}/E_{1}/E_{2}\)       & 83.6 & 97.3 & 85.5 & 71.1 & 87.2 & 87.9 & 81.7 & 91.9 \\
\(E_{4}/E_{3}/E_{1}/E_{2}\) & 89.9 & 99.0 & 91.0 & 83.0 & 89.5 & 91.4 & 92.4 & 94.6 \\
\(E_{3}/E_{2}/E_{1}/E_{4}\) & 84.7 & 97.6 & 87.0 & 72.8 & 86.9 & 90.1 & 91.1 & 88.5 \\
\(E_{2}/E_{4}/E_{1}/E_{3}\) & 83.1 & 97.3 & 86.3 & 70.9 & 85.2 & 90.3 & 87.6 & 90.6 \\
\(E_{2}/E_{4}/E_{3}/E_{1}\) & 87.5 & 98.2 & 88.8 & 77.4 & 86.9 & 91.3 & 90.4 & 92.1 \\
\(E_{2}/E_{3}/E_{1}/E_{4}\) & 83.9 & 97.4 & 86.7 & 71.9 & 85.6 & 90.4 & 89.9 & 87.7 \\
\(E_{2}/E_{3}/E_{4}/E_{1}\) & 88.9 & 98.4 & 89.6 & 79.8 & 86.5 & 91.3 & 91.3 & 92.5 \\
\(E_{2}/E_{1}/E_{4}/E_{3}\) & 78.2 & 96.3 & 84.2 & 64.5 & 84.5 & 87.0 & 85.5 & 87.8 \\
\(E_{1}/E_{3}/E_{4}/E_{2}\) & 79.2 & 96.4 & 84.7 & 65.7 & 85.3 & 84.7 & 88.9 & 88.9 \\
\(E_{3}/E_{4}/E_{2}/E_{1}\) & 90.6 & 99.0 & 90.9 & 82.9 & 87.7 & 91.7 & 92.0 & 94.0 \\
\(E_{3}/E_{4}/E_{1}/E_{2}\) & 87.3 & 98.1 & 89.0 & 77.9 & 87.5 & 89.4 & 91.3 & 92.0 \\
\(E_{1}/E_{3}/E_{2}/E_{4}\) & 74.9 & 95.9 & 83.8 & 62.0 & 84.9 & 84.3 & 88.9 & 86.6 \\
\(E_{1}/E_{2}/E_{4}/E_{3}\) & 76.3 & 96.0 & 84.1 & 63.1 & 85.2 & 86.7 & 85.1 & 87.9 \\
\(E_{4}/E_{1}/E_{2}/E_{3}\) & 85.7 & 97.8 & 87.3 & 74.8 & 87.6 & 89.3 & 88.4 & 92.8 \\
\(E_{1}/E_{2}/E_{3}/E_{4}\) & 96.9 & 97.2 & 98.1 & 95.6 & 96.1 & 97.2 & 96.8 & 97.7 \\
\(E_{4}/E_{2}/E_{1}/E_{3}\) & 88.1 & 98.4 & 89.1 & 78.5 & 87.5 & 91.7 & 89.2 & 93.1 \\
\(E_{4}/E_{2}/E_{3}/E_{1}\) & 89.7 & 98.7 & 90.0 & 81.2 & 88.5 & 92.0 & 91.5 & 93.7 \\
\(E_{4}/E_{1}/E_{3}/E_{2}\) & 87.7 & 98.8 & 88.9 & 79.5 & 87.5 & 89.2 & 91.0 & 93.7 \\
\(E_{3}/E_{2}/E_{4}/E_{1}\) & 90.0 & 99.1 & 90.7 & 81.9 & 89.3 & 91.9 & 92.2 & 94.0 \\
\(E_{3}/E_{1}/E_{2}/E_{4}\) & 74.6 & 95.8 & 83.6 & 61.7 & 85.3 & 84.1 & 89.1 & 86.5 \\
\bottomrule
\end{tabular}%
}
\end{table*}

\begin{table*}[h]
\caption{Ablation on constraint weights for training diffusion models on QM9~\cite{qm9}. The weight column follows the sequence \(\lambda_c(e)\), \text{ }\(\lambda_b(e)\), \text{ }\(\lambda_a(e)\), \text{ } and \(\lambda_v(e)\). Values in parentheses give the corresponding loss weights for \(E_{1}, E_{2}, E_{3},\) and \(E_{4}\) (\textit{e.g.,} \((4,3,2,1)\) indicates \(\lambda_c(e)=4\), \text{ } \(\lambda_b(e)=3\), \text{ } \(\lambda_a(e)=2\), \text{ } and \(\lambda_v(e)=1\)).}
\label{tab:ablation_weight_training}
\centering
\small
\setlength{\tabcolsep}{6pt}
\renewcommand{\arraystretch}{1.4}
\resizebox{1.0\textwidth}{!}{%
\begin{tabular}{l c c c c c c c c}
\toprule
\multirow{2}{*}{\raisebox{-0.9ex}{\textsc{Weight}}} &
\multicolumn{4}{c}{\textsc{Generation Quality}} &
\multicolumn{4}{c}{\textsc{Constraint Satisfaction}} \\
\cmidrule(lr){2-5}\cmidrule(lr){6-9}
& \textsc{Valid} &
\textsc{Valid \& Unique} &
\textsc{Atom Stab} &
\textsc{Mol Stab} &
\(E_{1}\) &
\(E_{2}\) &
\(E_{3}\) &
\(E_{4}\) \\
\midrule
(3, 3, 2, 2)         & 93.2 & 99.3 & 97.1 & 89.1 & 92.6 & 94.7 & 93.2 & 93.2 \\
(4, 2.5, 2.5, 1)     & 92.4 & 99.1 & 96.6 & 89.3 & 94.6 & 92.8 & 94.4 & 89.9 \\
(3, 2.5, 2, 1.5)     & 92.9 & 99.2 & 96.9 & 86.9 & 93.0 & 93.5 & 92.3 & 92.1 \\
(5, 3, 1.5, 0.5)     & 91.9 & 98.6 & 96.1 & 84.7 & 96.4 & 95.2 & 89.2 & 86.2 \\
(5, 2.5, 1.5, 1)     & 91.6 & 98.7 & 95.9 & 84.0 & 96.5 & 94.2 & 88.9 & 88.1 \\
(4.5, 3.5, 1.5, 0.5) & 91.2 & 98.5 & 95.8 & 82.5 & 95.9 & 96.1 & 88.4 & 86.0 \\
(5, 3, 1, 1)         & 90.9 & 98.6 & 95.4 & 80.9 & 96.5 & 95.1 & 86.9 & 88.0 \\
(3, 3.5, 2.5, 1)     & 86.7 & 97.8 & 93.4 & 74.9 & 88.9 & 96.3 & 95.5 & 88.1 \\
(3, 3, 3, 1)         & 52.5 & 93.2 & 78.6 & 49.4 & 70.2 & 84.2 & 85.0 & 68.9 \\
(3.5, 3, 2.5, 1)     & 82.2 & 97.5 & 91.6 & 70.4 & 91.4 & 95.2 & 94.6 & 86.8 \\
(3, 2.5, 3, 1.5)     & 47.9 & 92.7 & 74.9 & 44.2 & 66.8 & 79.2 & 83.5 & 79.6 \\
(4, 2.5, 2, 1.5)     & 88.9 & 98.2 & 94.4 & 76.6 & 93.4 & 92.8 & 90.9 & 93.8 \\
(3.5, 2.5, 2, 2)     & 49.4 & 93.0 & 76.2 & 45.2 & 72.5 & 76.3 & 78.1 & 85.4 \\
(4, 2, 2, 2)         & 44.8 & 92.6 & 72.5 & 40.8 & 74.2 & 70.4 & 76.4 & 84.8 \\
(3.5, 3, 1.5, 2)     & 56.3 & 94.2 & 81.3 & 52.1 & 78.6 & 88.5 & 74.3 & 85.0 \\
(4.5, 2.5, 2, 1)     & 94.2 & 99.4 & 97.2 & 90.0 & 96.1 & 93.4 & 93.7 & 93.6 \\
(4.5, 2, 2.5, 1)     & 89.3 & 98.3 & 94.5 & 77.1 & 94.9 & 89.2 & 94.8 & 88.3 \\
(4, 3.5, 1.5, 1)     & 93.0 & 99.1 & 96.7 & 88.0 & 94.6 & 96.4 & 89.0 & 89.7 \\
(3.5, 3.5, 2, 1)     & 93.6 & 99.3 & 96.9 & 88.7 & 92.9 & 96.6 & 93.5 & 91.3 \\
(3.5, 2, 2.5, 2)     & 46.2 & 92.5 & 73.9 & 42.0 & 71.7 & 70.7 & 80.1 & 86.4 \\
(4, 2, 3, 1)         & 50.7 & 93.3 & 77.0 & 46.8 & 72.4 & 73.5 & 85.3 & 70.7 \\
(4, 3, 2, 1)         & 96.9 & 97.2 & 98.1 & 95.6 & 96.1 & 97.2 & 96.8 & 97.7 \\
(2.5, 3.5, 2.5, 1.5) & 80.4 & 97.1 & 90.3 & 68.6 & 84.3 & 96.2 & 95.0 & 87.5 \\
(2.5, 3, 3, 1.5)     & 53.6 & 93.8 & 79.7 & 50.3 & 71.3 & 85.2 & 86.1 & 79.3 \\
(5, 2, 2, 1)         & 88.3 & 98.1 & 94.1 & 75.8 & 96.5 & 89.3 & 90.6 & 89.9 \\
(4.5, 3, 1, 1.5)     & 88.7 & 98.2 & 94.4 & 78.1 & 95.6 & 95.5 & 85.2 & 90.4 \\
(3, 4, 2, 1)         & 89.5 & 98.3 & 94.6 & 78.9 & 87.5 & 96.8 & 93.3 & 89.0 \\
(2, 4, 2.5, 1.5)     & 73.5 & 96.0 & 88.3 & 61.9 & 79.1 & 96.5 & 94.4 & 85.6 \\
(2.5, 2.5, 3, 2)     & 49.3 & 93.0 & 75.8 & 44.9 & 70.2 & 77.6 & 84.1 & 84.3 \\
(1.5, 4, 3, 1.5)     & 45.2 & 92.4 & 73.0 & 41.1 & 64.1 & 91.3 & 84.4 & 76.8 \\
(1, 4.5, 3, 1.5)     & 42.9 & 92.0 & 70.9 & 39.2 & 60.2 & 93.5 & 84.3 & 75.5 \\
\bottomrule
\end{tabular}%
}
\end{table*}

\begin{table*}[h]
\caption{MAE ($\downarrow$) under different neuro-symbolic usage (training only vs.\ sampling only) for quantum-property-conditioned generation. The diffusion models are trained on QM9~\cite{qm9}.}
\label{tab:ablation_mae}
\centering
\small
\setlength{\tabcolsep}{7pt}
\renewcommand{\arraystretch}{1.4}
\resizebox{1.0\textwidth}{!}{%
\begin{tabular}{l c c c c c c}
\toprule
\textsc{Use During} &
$C_v$ \(\left(\frac{\text{cal}}{\text{mol}}\,\text{K}\right)\) &
$\mu$ (D) &
$\alpha$ (Bohr$^3$) &
$\Delta \varepsilon$ (meV) &
$\varepsilon_{\mathrm{HOMO}}$ (meV) &
$\varepsilon_{\mathrm{LUMO}}$ (meV) \\
\midrule
Training    & 0.612 & 0.676 & 1.62 & 348 & 232 & 260 \\
Sampling    & 0.658	& 0.724	& 1.76	& 382	& 248	& 278 \\
Both        & 0.569 & 0.618 & 1.49 & 314 & 219 & 246 \\
\bottomrule
\end{tabular}%
}
\end{table*}

\begin{table*}[h]
\caption{Ablation on constraint weights during sampling on QM9~\cite{qm9}. The weight column follows the sequence \(\lambda_c(e)\), \text{ }\(\lambda_b(e)\), \text{ }\(\lambda_a(e)\), \text{ } and \(\lambda_v(e)\). Values in parentheses give the corresponding loss weights for \(E_{1}, E_{2}, E_{3},\) and \(E_{4}\) (\textit{e.g.,} \((4,3,2,1)\) indicates \(\lambda_c(e)=4\), \text{ } \(\lambda_b(e)=3\), \text{ } \(\lambda_a(e)=2\), \text{ } and \(\lambda_v(e)=1\)).}
\label{tab:ablation_weight_samling}
\centering
\small
\setlength{\tabcolsep}{6pt}
\renewcommand{\arraystretch}{1.4}
\resizebox{1.0\textwidth}{!}{%
\begin{tabular}{l c c c c c c c c}
\toprule
\multirow{2}{*}{\raisebox{-0.9ex}{\textsc{Weight}}} &
\multicolumn{4}{c}{\textsc{Generation Quality}} &
\multicolumn{4}{c}{\textsc{Constraint Satisfaction}} \\
\cmidrule(lr){2-5}\cmidrule(lr){6-9}
& \textsc{Valid} &
\textsc{Valid \& Unique} &
\textsc{Atom Stab} &
\textsc{Mol Stab} &
\(E_{1}\) &
\(E_{2}\) &
\(E_{3}\) &
\(E_{4}\) \\
\midrule
(4, 3, 2, 1)         & 93.8 & 97.6 & 97.9 & 89.4 & 96.2 & 95.5 & 91.6 & 85.8 \\
(3, 3, 2, 2)         & 93.7 & 97.6 & 97.9 & 87.2 & 94.8 & 95.0 & 92.0 & 90.7 \\
(4, 2.5, 2.5, 1)     & 93.6 & 97.6 & 97.8 & 88.1 & 96.4 & 93.4 & 94.2 & 85.4 \\
(3, 2.5, 2, 1.5)     & 93.8 & 97.7 & 98.0 & 88.6 & 94.4 & 93.1 & 91.3 & 89.5 \\
(5, 3, 1.5, 0.5)     & 92.9 & 97.5 & 97.5 & 87.5 & 97.3 & 95.2 & 84.4 & 82.4 \\
(5, 2.5, 1.5, 1)     & 92.9 & 97.5 & 97.5 & 87.3 & 97.4 & 93.6 & 84.2 & 85.2 \\
(4.5, 3.5, 1.5, 0.5) & 92.7 & 97.5 & 97.4 & 86.2 & 97.1 & 96.4 & 83.8 & 81.9 \\
(5, 3, 1, 1)         & 92.7 & 97.5 & 97.4 & 87.0 & 97.3 & 95.2 & 82.7 & 85.9 \\
(3, 3.5, 2.5, 1)     & 93.1 & 97.6 & 97.7 & 86.8 & 90.1 & 96.6 & 95.4 & 84.1 \\
(3, 3, 3, 1)         & 93.0 & 97.5 & 97.6 & 88.5 & 89.7 & 95.8 & 96.4 & 83.8 \\
(3.5, 3, 2.5, 1)     & 93.3 & 97.6 & 97.7 & 87.9 & 93.8 & 95.1 & 94.9 & 84.3 \\
(3, 2.5, 3, 1.5)     & 92.6 & 97.4 & 97.2 & 85.8 & 88.9 & 93.0 & 96.3 & 92.1 \\
(4, 2.5, 2, 1.5)     & 93.1 & 97.5 & 97.6 & 85.7 & 95.7 & 92.8 & 90.9 & 92.8 \\
(3.5, 2.5, 2, 2)     & 92.4 & 97.4 & 97.1 & 87.4 & 87.8 & 91.9 & 89.8 & 95.8 \\
(4, 2, 2, 2)         & 92.1 & 97.3 & 96.9 & 86.1 & 95.1 & 84.2 & 90.2 & 96.0 \\
(3.5, 3, 1.5, 2)     & 92.5 & 97.4 & 97.2 & 84.6 & 89.2 & 95.5 & 86.9 & 95.4 \\
\bottomrule
\end{tabular}%
}
\end{table*}

\begin{table*}[h]
\caption{Ablation on constraint enforcement schedules across sampling steps on QM9~\cite{qm9}.}
\label{tab:ablation_enforcement_samling}
\centering
\small
\setlength{\tabcolsep}{6pt}
\renewcommand{\arraystretch}{1.4}
\resizebox{1.0\textwidth}{!}{%
\begin{tabular}{l c c c c c c c c}
\toprule
\multirow{2}{*}{\raisebox{-0.16cm}{\textsc{Enforcement}}} &
\multicolumn{4}{c}{\textsc{Generation Quality}} &
\multicolumn{4}{c}{\textsc{Constraint Satisfaction}} \\
\cmidrule(lr){2-5}\cmidrule(lr){6-9}
&
\textsc{Valid} &
\textsc{Valid \& Unique} &
\textsc{Atom Stab} &
\textsc{Mol Stab} &
\(E_{1}\) &
\(E_{2}\) &
\(E_{3}\) &
\(E_{4}\) \\
\midrule
Every 2 steps          & 95.5 & 96.7 & 97.9 & 90.3 & 94.5 & 95.8 & 94.6 & 93.8 \\
Every 10 steps         & 93.7 & 96.6 & 96.1 & 84.1 & 91.8 & 93.7 & 92.4 & 91.3 \\
Every 100 steps        & 90.1 & 98.5 & 94.3 & 82.0 & 83.3 & 87.5 & 85.6 & 84.8 \\
First 50 steps    & 92.0 & 97.3 & 96.3 & 80.0 & 79.1 & 84.3 & 82.6 & 81.5 \\
First 250 steps   & 93.1 & 97.6 & 97.4 & 83.2 & 87.6 & 91.2 & 89.5 & 88.4 \\
First 500 steps   & 93.9 & 98.7 & 98.0 & 84.7 & 90.5 & 93.2 & 91.8 & 90.9 \\
Last 50 steps     & 93.2 & 96.6 & 97.3 & 82.6 & 86.4 & 90.3 & 88.7 & 88.0 \\
Last 250 steps    & 94.1 & 96.7 & 98.1 & 85.0 & 93.1 & 95.5 & 94.2 & 93.4 \\
Last 500 steps    & 94.3 & 97.7 & 98.3 & 85.5 & 93.9 & 96.1 & 94.9 & 94.3 \\
\bottomrule
\end{tabular}%
}
\end{table*}

\begin{table*}[h]
\caption{Comparison of symbolic constraints usage (training only vs.\ sampling only) on two datasets.}
\label{tab:ablation_ns_setting}
\centering
\small
\setlength{\tabcolsep}{6pt}
\renewcommand{\arraystretch}{1.4}
\resizebox{1.0\textwidth}{!}{%
\begin{tabular}{l l c c c c c c c c}
\toprule
\multirow{2}{*}{\raisebox{-0.7ex}{\textsc{{\normalsize Dataset}}}} &
\multirow{2}{*}{\raisebox{-0.7ex}{\textsc{{\normalsize Use During}}}} &
\multicolumn{4}{c}{\textsc{Generation Quality}} &
\multicolumn{4}{c}{\textsc{Constraint Satisfaction}} \\
\cmidrule(lr){3-6}\cmidrule(lr){7-10}
& &
\textsc{Valid} &
\textsc{Valid \& Unique} &
\textsc{Atom Stab} &
\textsc{Mol Stab} &
\(E_{1}\) &
\(E_{2}\) &
\(E_{3}\) &
\(E_{4}\) \\
\midrule
\multirow{2}{*}{\raisebox{-3.2ex}{QM9~\cite{qm9}}} &
Training &
94.5 & 95.9 & 96.7 & 91.9 & 94.1 & 95.4 & 93.5 & 95.0 \\
&
Sampling &
91.5 & 96.5 & 93.9 & 88.3 & 90.9 & 92.8 & 91.2 & 92.9 \\ &
Both & 96.9 & 97.2 & 98.1 & 95.6 & 96.1 & 97.2 & 96.8 & 97.7 \\
\midrule
\multirow{2}{*}{\raisebox{-3.2ex}{GEOM-Drugs~\cite{geom-drug}}} &
Training &
95.9 & 97.5 & 86.5 & 75.8 & 87.8 & 91.7 & 90.1 & 91.6 \\
&
Sampling &
94.9 & 95.1 & 85.1 & 72.5 & 86.9 & 90.8 & 88.3 & 91.7 \\ &
Both & 97.2 & 98.8 & 87.9 & 79.4 & 92.6 & 95.8 & 93.6 & 96.2 \\
\bottomrule
\end{tabular}%
}
\end{table*}

\clearpage
\begin{table}[h]
\centering
\small
\setlength{\tabcolsep}{8pt}
\renewcommand{\arraystretch}{1.6}
\caption{Ablation study of the weighting strategy on two datasets. Our adaptive weighting has two phases: Phase~1 (\textit{i.e.,} linear ramping) linearly ramps up loss weights, and Phase~2 (\textit{i.e.,} energy adaptive) increases weights based on the degree of constraint satisfaction (see Section~\ref{sec:method} for details).}
\vspace{0.15cm}
\label{tab:ablation_weighting_mechanism}
\resizebox{\linewidth}{!}{%
\begin{tabular}{llcccccccc}
\toprule
\multirow{2}{*}{\raisebox{-0.9ex}{\textsc{Dataset}}} &
\multirow{2}{*}{\raisebox{-0.9ex}{\textsc{Weighting}}} &
\multicolumn{4}{c}{\textsc{Generation Quality}} &
\multicolumn{4}{c}{\textsc{Constraint Satisfaction}} \\
\cmidrule(lr){3-6}\cmidrule(lr){7-10}
& & \textsc{Valid} &
\textsc{Valid \& Unique} &
\textsc{Atom Stab} &
\textsc{Mol Stab} &
\(E_{1}\) & \(E_{2}\) & \(E_{3}\) & \(E_{4}\) \\
\midrule
\multirow{3}{*}{\raisebox{-4.0ex}{QM9~\cite{qm9}}} &
Neither (Non-adaptive)          & 88.5 & 96.8 & 89.2 & 84.7 & 75.3 & 81.1 & 78.5 & 85.1 \\
& Linear Ramping Only  & 91.9 & 97.0 & 93.5 & 87.6 & 81.3 & 86.5 & 86.2 & 90.3 \\
& Energy Adaptive Only & 94.3 & 97.1 & 94.7 & 90.8 & 86.4 & 91.7 & 89.2 & 93.6 \\ &
Both & 96.9 & 97.2 & 98.1 & 95.6 & 96.1 & 97.2 & 96.8 & 97.7 \\
\midrule
\multirow{3}{*}{\raisebox{-4.0ex}{GEOM-Drugs~\cite{geom-drug}}} &
Neither (Non-adaptive)         & 85.2 & 96.3 & 78.6 & 62.4 & 75.2 & 81.2 & 76.5 & 77.9 \\
& Linear Ramping Only  & 88.7 & 96.4 & 81.3 & 66.2 & 78.5 & 86.8 & 80.6 & 83.7 \\
& Energy Adaptive Only & 91.5 & 96.7 & 85.3 & 72.6 & 82.3 & 92.6 & 88.3 & 91.2 \\ &
Both & 97.2 & 98.8 & 87.9 & 79.4 & 92.6 & 95.8 & 93.6 & 96.2 \\
\bottomrule
\end{tabular}%
}
\end{table}

\begin{figure*}[h]
    \centering
    \includegraphics[width=1.0\textwidth]{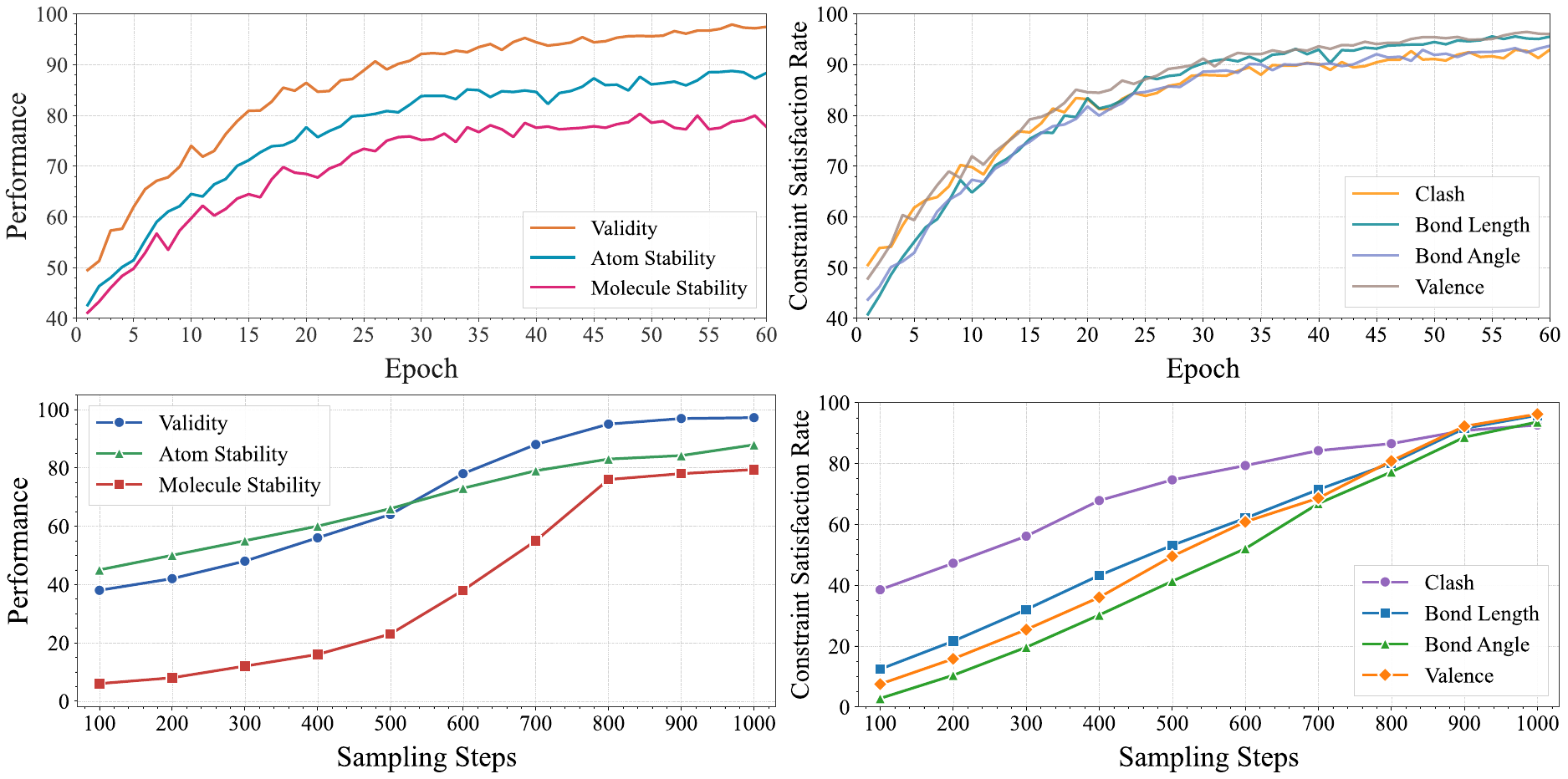}
    \caption{Ablation on sampling steps and training epochs on GEOM-Drugs~\cite{geom-drug}. We save diffusion model checkpoints every epoch. Increasing either sampling steps or training epochs consistently improves generation quality and constraint compliance.}
    \label{fig:gd_step_epoch}
\end{figure*}

\vspace{1\baselineskip}

In summary, our ablation studies include:

\begin{enumerate}
    \item sequential versus non-sequential enforcement (\textit{i.e.}, activating all constraints simultaneously from the start of training) of symbolic constraint subsets (Table~\ref{tab:ablation_set_qm9} and Table~\ref{tab:ablation_set_geom_drugs});
    
    \item training curricula and schedules (\textit{e.g.}, constraint activation order in Table~\ref{tab:constraint_order_ablation} and epoch schedules in Figure~\ref{fig:gd_step_epoch});

    \item constraint weighting schemes (Table~\ref{tab:ablation_weight_training}, Table~\ref{tab:ablation_weight_samling}, and Table~\ref{tab:ablation_weighting_mechanism});

    \item applying neuro-symbolic constraints during training versus sampling only (Table~\ref{tab:ablation_mae} and Table~\ref{tab:ablation_ns_setting});
    
    \item sampling-time strategies (\textit{e.g.}, when and how constraints are enforced during denoising in Table~\ref{tab:ablation_enforcement_samling}, and the denoising-step schedule in Figure~\ref{fig:gd_step_epoch}).
\end{enumerate}

Altogether, the extensive ablation studies validate that the proposed training strategies, including sequential constraint activation and adaptive weighting, effectively improve generation quality, constraint compliance, and physicochemical plausibility in diffusion-based 3D molecule generation. We also show that our framework supports diffusion model fine-tuning under the same symbolic constraints, with an ablation over fine-tuning epochs. Further details are provided in Section~\ref{sec:fine-tuning}.

\clearpage
\section{Generalizing Our Training Strategy to Multiple Diffusion Backbones}
\label{generalization_to_backbones}

As shown in Table~\ref{tab:generalization_backbones}, the proposed training strategy yields consistently high generation quality and constraint satisfaction across all five backbones on both datasets, confirming its generalizability. On QM9~\cite{qm9}, all backbones exceed 95\% validity and 92\% on every constraint metric. On the more challenging Geom-Drugs~\cite{geom-drug}, constraint satisfaction remains above 87\% despite a drop in molecule stability, suggesting that the symbolic constraints provide a reliable quality floor regardless of molecular complexity. Among the backbones, UAE-3D~\cite{uae-3d} and SLDM~\cite{sldm} benefit most, indicating that stronger base architectures better leverage the introduced physicochemical priors.

\vspace{1\baselineskip}

\begin{table}[h]
\centering
\small
\setlength{\tabcolsep}{7pt}
\renewcommand{\arraystretch}{1.4}
\caption{Generalization of our training strategy to different diffusion backbones on two datasets.}
\vspace{0.1cm}
\label{tab:generalization_backbones}
\resizebox{\linewidth}{!}{%
\begin{tabular}{llcccccccc}
\toprule
\multirow{2}{*}{\raisebox{-0.9ex}{\textsc{Dataset}}} &
\multirow{2}{*}{\raisebox{-0.9ex}{\textsc{Method}}} &
\multicolumn{4}{c}{\textsc{Generation Quality}} &
\multicolumn{4}{c}{\textsc{Constraint Satisfaction}} \\
\cmidrule(lr){3-6}\cmidrule(lr){7-10}
& & \textsc{Valid} &
\textsc{Valid \& Unique} &
\textsc{Atom Stab} &
\textsc{Mol Stab} &
\(E_{1}\) & \(E_{2}\) & \(E_{3}\) & \(E_{4}\) \\
\midrule
\multirow{5}{*}{\textsc{QM9~\cite{qm9}}}
& GCDM~\cite{gcdm}   & 95.8 & 97.6 & 96.6 & 95.1 & 92.7 & 95.9 & 94.4 & 95.7 \\
& UniGEM~\cite{unigem} & 96.4 & 96.7 & 97.9 & 96.9 & 94.4 & 97.2 & 96.8 & 97.0 \\
& UAE-3D~\cite{uae-3d} & 97.1 & 96.8 & 98.3 & 98.8 & 95.7 & 98.1 & 97.5 & 97.9 \\
& RADM~\cite{radm}   & 96.1 & 96.3 & 97.1 & 95.7 & 93.5 & 96.1 & 95.1 & 96.2 \\
& SLDM~\cite{sldm}   & 96.7 & 96.6 & 98.1 & 96.3 & 94.9 & 97.6 & 96.9 & 97.4 \\
\midrule
\multirow{5}{*}{\textsc{Geom-Drugs~\cite{geom-drug}}}
& GCDM~\cite{gcdm}  & 94.7 & 95.4 & 90.6 & 70.4 & 87.1 & 90.8 & 89.5 & 91.3 \\
& UniGEM~\cite{unigem} & 96.5 & 94.6 & 91.3 & 72.3 & 90.0 & 92.3 & 93.1 & 95.7 \\
& UAE-3D~\cite{uae-3d} & 97.3 & 95.8 & 92.2 & 78.0 & 93.9 & 93.7 & 94.4 & 96.8 \\
& RADM ~\cite{radm}  & 95.2 & 95.4 & 89.9 & 74.3 & 88.5 & 93.3 & 92.6 & 92.5 \\
& SLDM~\cite{sldm}   & 96.8 & 96.7 & 90.6 & 78.7 & 92.0 & 93.1 & 93.6 & 96.1 \\
\bottomrule
\end{tabular}%
}
\end{table}

\vspace{2\baselineskip}
\section{Out-of-Distribution Generalizability of SensDiff}
\label{ood}

To assess out-of-distribution generalizability, we conduct cross-dataset evaluations by training on one dataset and evaluating on another (QM9~\cite{qm9} $\leftrightarrow$ GEOM-Drugs~\cite{geom-drug}), where the two datasets differ significantly in molecular property and complexity. As shown in Tables~\ref{tab:ood1} and~\ref{tab:ood2}, SensDiff achieves the best performance across all metrics in both transfer directions, obtaining 87.9\% and 92.9\% validity with 69.2\% and 82.0\% molecule stability, respectively, surpassing the strongest baselines by a substantial margin. The constraint compliance metrics ($E_1$ to $E_4$) exhibit a similarly dominant trend, confirming the structural soundness of generated molecules even under severe distribution shift. We attribute this strong transferability to the neuro-symbolic nature of our approach. Rather than merely memorizing data-specific patterns, the symbolic constraints encode fundamental physical and chemical principles, such as bond length rules, valence rules, and steric constraints, that are universally valid regardless of the molecular domain. By internalizing these scientific priors during training and enforcing them during sampling, SensDiff acquires an effective understanding of molecular geometry that naturally generalizes to unseen datasets, whereas purely data-driven baselines degrade more severely when the test distribution departs from the training one.

To evaluate the ability to generate rare (out-of-distribution) molecules, we select molecules with extreme LogP or QED values from the tail of the training distribution and task each model to generate molecules targeting these underrepresented property regimes. As shown in Table~\ref{tab:ood_logp_qed}, SensDiff consistently achieves the best generation quality, reaching 84.2\% validity, 88.1\% atom stability, and 68.6\% molecule stability, outperforming the strongest baseline, UAE-3D~\cite{uae-3d}, by 2.6\%, 2.9\%, and 2.2\%, respectively. The constraint compliance metrics ($E_1$ to $E_4$) further suggest this merit, with SensDiff surpassing all baselines by a clear margin. These improvements stem from the inherent difficulty of generating out-of-distribution molecules. As the target drifts toward low-density regions of the data distribution, purely data-driven methods lack sufficient training signal and tend to produce implausible structures. Our neuro-symbolic-guided sampling addresses this by enforcing explicit structural constraints during generation, helping preserve molecular validity even when the model extrapolates beyond well-represented property regions.

\begin{table}[h]
\centering
\small
\setlength{\tabcolsep}{7pt}
\renewcommand{\arraystretch}{1.3}
\caption{Out-of-distribution performance comparison. We train the model on GEOM-Drugs~\cite{geom-drug} and evaluate on QM9~\cite{qm9} to assess its cross-data generalizability.}
\vspace{0.15cm}
\label{tab:ood1}
\resizebox{\linewidth}{!}{%
\begin{tabular}{lcccccccc}
\toprule
\multirow{2}{*}{\raisebox{-0.9ex}{\centering\textsc{Model}}} &
\multicolumn{4}{c}{\textsc{Generation Quality}} &
\multicolumn{4}{c}{\textsc{Constraint Satisfaction}} \\
\cmidrule(lr){2-5}\cmidrule(lr){6-9}
& \textsc{Valid} &
\textsc{Valid \& Unique} &
\textsc{Atom Stab} &
\textsc{Mol Stab} &
\(E_{1}\) & \(E_{2}\) & \(E_{3}\) & \(E_{4}\) \\
\midrule
EDM~\cite{edm}              & 58.4 & 98.6 & 72.2 & 41.5 & 57.6 & 64.8 & 62.1 & 67.5 \\
EEGSDE~\cite{eegsde}        & 61.9 & 98.8 & 74.6 & 44.9 & 60.2 & 67.4 & 64.3 & 69.8 \\
GEOLDM~\cite{geoldm}        & 66.5 & 98.9 & 78.2 & 50.7 & 64.7 & 71.9 & 68.9 & 74.3 \\
EquiFM~\cite{equifm}        & 71.2 & 99.0 & 81.6 & 56.6 & 69.2 & 76.2 & 73.5 & 78.6 \\
GeoBFN~\cite{geobfn}        & 74.9 & 99.1 & 84.2 & 61.3 & 72.8 & 79.5 & 76.9 & 81.9 \\
GCDM~\cite{gcdm}            & 77.5 & 99.2 & 85.9 & 64.9 & 75.5 & 82.1 & 79.6 & 84.4 \\
RADM~\cite{radm}            & 79.7 & 99.2 & 87.3 & 67.2 & 77.3 & 84.3 & 81.7 & 86.1 \\
SLDM~\cite{sldm}            & 82.7 & 99.3 & 89.5 & 71.8 & 81.3 & 87.9 & 85.6 & 89.7 \\
UAE-3D~\cite{uae-3d}        & 84.2 & 99.3 & 90.3 & 73.6 & 82.9 & 89.2 & 87.1 & 90.9 \\
UniGEM~\cite{unigem}        & 83.4 & 99.3 & 89.9 & 72.8 & 82.2 & 88.7 & 86.5 & 90.1 \\
\midrule
\rowcolor[RGB]{225, 230, 245} SensDiff (Ours)             & \textbf{92.9} & \textbf{99.4} & \textbf{93.3} & \textbf{82.0} & \textbf{89.6} & \textbf{93.3} & \textbf{92.2} & \textbf{95.7} \\
\bottomrule
\end{tabular}%
}
\end{table}

\begin{table}[h]
\centering
\small
\setlength{\tabcolsep}{7pt}
\renewcommand{\arraystretch}{1.3}
\caption{Out-of-distribution performance comparison. We train the model on QM9~\cite{qm9} and evaluate on GEOM-Drugs~\cite{geom-drug} to assess its cross-data generalizability.}
\vspace{0.15cm}
\label{tab:ood2}
\resizebox{\linewidth}{!}{%
\begin{tabular}{lcccccccc}
\toprule
\multirow{2}{*}{\raisebox{-0.9ex}{\centering\textsc{Model}}} &
\multicolumn{4}{c}{\textsc{Generation Quality}} &
\multicolumn{4}{c}{\textsc{Constraint Satisfaction}} \\
\cmidrule(lr){2-5}\cmidrule(lr){6-9}
& \textsc{Valid} &
\textsc{Valid \& Unique} &
\textsc{Atom Stab} &
\textsc{Mol Stab} &
\(E_{1}\) & \(E_{2}\) & \(E_{3}\) & \(E_{4}\) \\
\midrule
EDM~\cite{edm}           & 46.3 & 98.8 & 61.8 & 18.5 & 48.6 & 56.2 & 53.8 & 58.3 \\
EEGSDE~\cite{eegsde}     & 49.7 & 98.9 & 64.2 & 20.9 & 51.2 & 58.9 & 56.1 & 60.7 \\
GEOLDM~\cite{geoldm}     & 54.6 & 99.0 & 68.0 & 25.8 & 55.9 & 63.2 & 60.5 & 65.1 \\
EquiFM~\cite{equifm}     & 59.9 & 99.1 & 70.8 & 31.5 & 60.7 & 68.0 & 65.2 & 69.8 \\
GeoBFN~\cite{geobfn}     & 64.1 & 99.2 & 71.3 & 36.9 & 65.0 & 72.2 & 69.4 & 73.9 \\
GCDM~\cite{gcdm}         & 67.5 & 99.2 & 73.9 & 41.3 & 68.2 & 75.4 & 72.7 & 76.9 \\
RADM~\cite{radm}         & 70.4 & 99.3 & 76.0 & 45.2 & 71.1 & 78.1 & 75.9 & 79.7 \\
SLDM~\cite{sldm}         & 75.3 & 99.3 & 79.2 & 52.7 & 76.1 & 82.7 & 80.4 & 84.3 \\
UAE-3D~\cite{uae-3d}     & 77.2 & 99.4 & 80.6 & 55.8 & 77.9 & 84.3 & 82.1 & 85.9 \\
UniGEM~\cite{unigem}     & 76.2 & 99.3 & 80.0 & 54.4 & 77.1 & 83.6 & 81.3 & 85.1 \\
\midrule
\rowcolor[RGB]{225, 230, 245} SensDiff (Ours)          & \textbf{87.9} & \textbf{99.4} & \textbf{85.5} & \textbf{69.2} & \textbf{89.6} & \textbf{90.2} & \textbf{91.1} & \textbf{94.5} \\
\bottomrule
\end{tabular}%
}
\end{table}

\begin{table}[h]
\centering
\small
\setlength{\tabcolsep}{7pt}
\renewcommand{\arraystretch}{1.3}
\caption{Performance comparison on generating rare molecular samples with extreme (high) LogP and QED values from the tail of the training distribution.}
\vspace{6pt}
\label{tab:ood_logp_qed}
\resizebox{\linewidth}{!}{%
\begin{tabular}{lcccccccccc}
\toprule
\multirow{2}{*}{\raisebox{-0.9ex}{\centering\textsc{Model}}} &
\multicolumn{2}{c}{\textsc{Descriptor}} &
\multicolumn{3}{c}{\textsc{Generation Quality}} &
\multicolumn{4}{c}{\textsc{Constraint Satisfaction}} \\
\cmidrule(lr){2-3}\cmidrule(lr){4-6}\cmidrule(lr){7-10}
& \textsc{LogP} & \textsc{QED}
& \textsc{Valid} & \textsc{Atom Stab} & \textsc{Mol Stab}
& \(E_{1}\) & \(E_{2}\) & \(E_{3}\) & \(E_{4}\) \\
\midrule
GEOLDM~\cite{geoldm}  & 3.86 & 0.708 & 64.1 & 69.2 & 43.8 & 65.2 & 74.2 & 71.8 & 77.4 \\
GCDM~\cite{gcdm}      & 3.94 & 0.731 & 68.4 & 73.6 & 51.2 & 69.4 & 79.2 & 77.1 & 82.6 \\
RADM~\cite{radm}      & 4.08 & 0.748 & 72.1 & 76.4 & 54.8 & 73.2 & 82.1 & 79.6 & 84.2 \\
SLDM~\cite{sldm}      & 4.19 & 0.764 & 74.8 & 78.6 & 59.2 & 75.8 & 84.6 & 82.1 & 86.6 \\
UniGEM~\cite{unigem}  & 4.22 & 0.771 & 75.6 & 79.4 & 60.4 & 76.4 & 85.2 & 82.8 & 87.4 \\
UAE-3D~\cite{uae-3d}  & 4.48 & 0.792 & 81.6 & 85.2 & 66.4 & 82.2 & 90.4 & 88.8 & 92.6 \\
\midrule
\rowcolor[RGB]{225, 230, 245}  SensDiff (Ours)       & 4.33 & 0.779 & \textbf{84.2} & \textbf{88.1} & \textbf{68.6} & \textbf{84.8} & \textbf{93.6} & \textbf{91.2} & \textbf{96.4} \\
\bottomrule
\end{tabular}%
}
\end{table}

\clearpage
\section{Additional Evaluation Metrics for 3D Molecule Generation}
\label{more_metrics}

In this section, we consider two additional evaluation metrics: Tanimoto similarity and constraint satisfaction rate (CSR). Tanimoto similarity measures structural similarity by comparing molecular fingerprints of generated samples to those from the prior distribution~\cite{tanimoto, butina1999unsupervised}. The constraint satisfaction rate is the fraction of generated molecules that satisfy the prescribed rules. 

\begin{figure*}[h]
    \centering
    \includegraphics[width=1.0\textwidth]{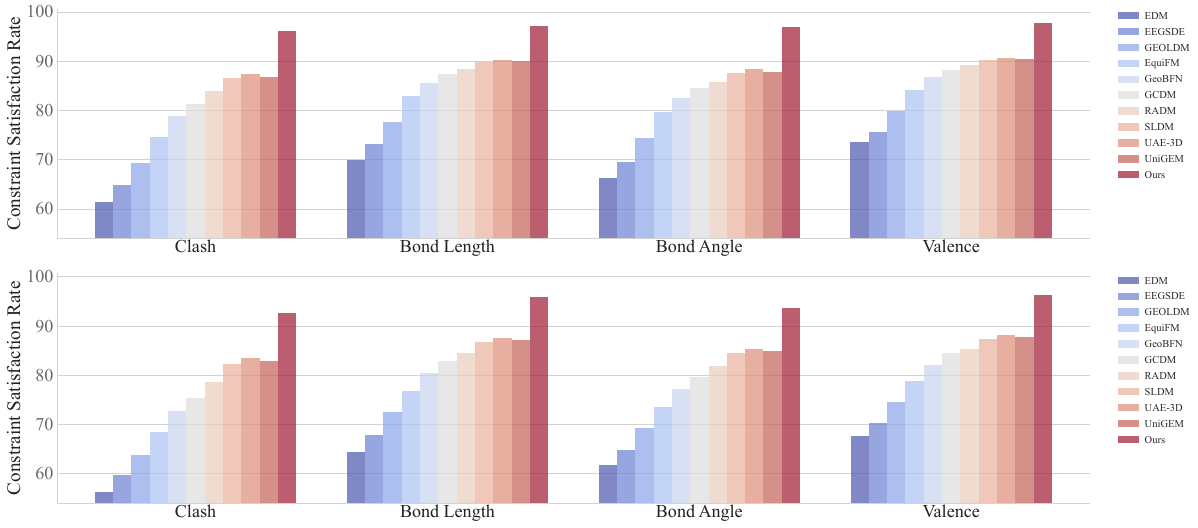}
    \caption{Constraint satisfaction rate on QM9~\cite{qm9} (\textbf{upper}) and GEOM-Drugs~\cite{geom-drug} (\textbf{bottom}).}
    \label{fig:csr_appendix}
\end{figure*}

As shown in Table~\ref{tab:tanimoto_sim}, SensDiff achieves the highest Tanimoto similarity on both datasets, outperforming the strongest baseline by 10.7\% and 13.7\%, respectively, indicating that SensDiff generates molecules whose fingerprints more closely match the training distribution. The consistent gains across both datasets suggest that the constraints not only enforce structural validity but also steer generation toward more chemically realistic substructures. As shown in Figure~\ref{fig:csr_appendix}, SensDiff consistently achieves the highest constraint satisfaction rate across all four symbolic constraints on both QM9 and Geom-Drugs, surpassing the strongest baseline in every category. Notably, the improvement is most pronounced on the clash and bond angle constraints, which are typically harder to satisfy, indicating that our training strategy particularly excels at capturing complex geometric dependencies.

\vspace{1\baselineskip}

\begin{table}[h]
\centering
\tiny
\setlength{\tabcolsep}{5pt}
\renewcommand{\arraystretch}{1.3}
\caption{Tanimoto similarity between generated samples and the training distribution on two datasets.}
\vspace{0.15cm}
\label{tab:tanimoto_sim}
\resizebox{0.9\linewidth}{!}{%
\begin{tabular}{lclc}
\toprule
\multicolumn{2}{c}{\textsc{QM9~\cite{qm9}}} & \multicolumn{2}{c}{\textsc{GEOM-Drugs~\cite{geom-drug}}} \\
\cmidrule(lr){1-2}\cmidrule(lr){3-4}
\textsc{Model} & \textsc{Tanimoto Similarity (\%, $\uparrow$)} & \textsc{Model} & \textsc{Tanimoto Similarity (\%, $\uparrow$)} \\
\midrule
EDM~\cite{edm}           & 56.2 & EDM~\cite{edm}           & 24.2 \\
EEGSDE~\cite{eegsde}     & 60.8 & EEGSDE~\cite{eegsde}     & 27.3 \\
GEOLDM~\cite{geoldm}     & 63.6 & GEOLDM~\cite{geoldm}     & 32.7 \\
EquiFM~\cite{equifm}     & 65.8 & EquiFM~\cite{equifm}     & 36.1 \\
GeoBFN~\cite{geobfn}     & 68.1 & GeoBFN~\cite{geobfn}     & 39.6 \\
GCDM~\cite{gcdm}         & 69.8 & GCDM~\cite{gcdm}         & 42.2 \\
RADM~\cite{radm}         & 71.2 & RADM~\cite{radm}         & 45.2 \\
SLDM~\cite{sldm}         & 73.4 & SLDM~\cite{sldm}         & 49.7 \\
UAE-3D~\cite{uae-3d}     & 74.1 & UAE-3D~\cite{uae-3d}     & 51.4 \\
UniGEM~\cite{unigem}     & 73.7 & UniGEM~\cite{unigem}     & 50.3 \\
\midrule
\rowcolor[RGB]{225, 230, 245}
SensDiff (Ours) & \textbf{84.8} & SensDiff (Ours) & \textbf{65.1} \\
\bottomrule
\end{tabular}%
}
\end{table}

\vspace{2\baselineskip}

\clearpage
\section{Analysis of Failure Cases}
\label{fail_case_appendix}

\begin{figure*}[t]
    \centering
    \includegraphics[width=1.0\textwidth]{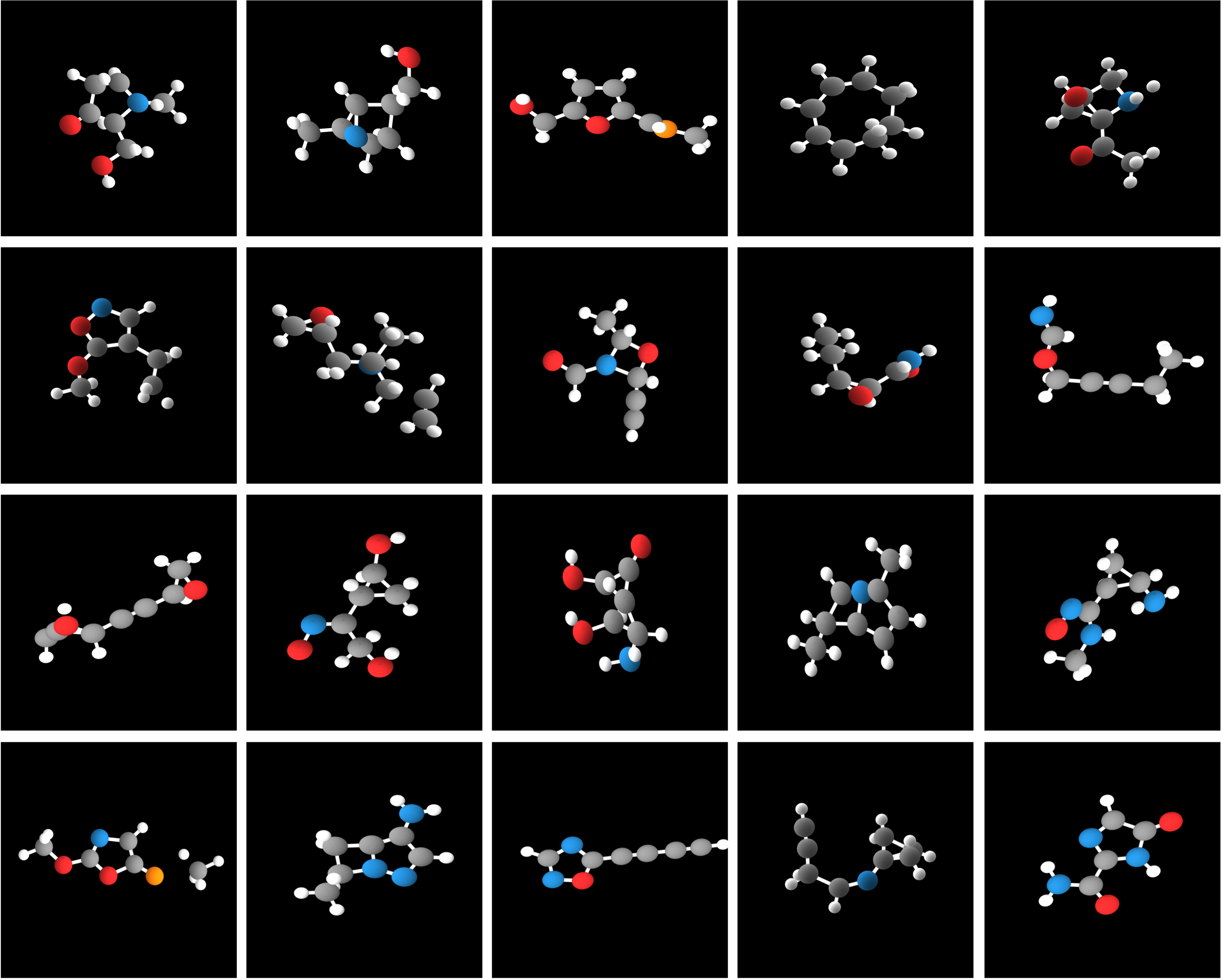}
    \caption{Randomly selected failure cases. In each row, the five molecules violate \(E_{\text{clash}}\), \(E_{\text{bond\_len}}\), \(E_{\text{angle}}\), and \(E_{\text{val}}\), respectively.}
    \label{fig:failure_case}
    \vspace{0.8cm}
\end{figure*}

A key advantage of symbolic constrained molecule generation is that it makes model failures interpretable. In purely implicit molecule generators, failures typically appear as low validity or unrealistic 3D structures, but it is difficult to pinpoint which physicochemical regularities were not learned. In SensDiff, the symbolic energies \(E_{\text{clash}}\) (\textit{i.e.}, steric non-overlap), \(E_{\text{bond\_len}}\) (\textit{i.e.}, bond length), \(E_{\text{angle}}\) (\textit{i.e.}, bond angle), and \(E_{\text{val}}\) (\textit{i.e.}, valence validity) provide human-understandable diagnostics via Equation~\ref{eq:clash_energy}--\ref{eq:val_energy}. This helps us view failures not merely as invalid molecules, but as evidence that the model has not yet internalized a specific physicochemical rules. Accordingly, we conduct a qualitative diagnostic study of representative failure cases from earlier versions of our model and weaker training strategies. We use these findings to inform the final constraint design, the overall objective in Equation~\ref{eq:total_loss}, and the sequential curriculum in Equation~\ref{eq:lambda_schedule}. We inspected representative generated molecules together with their recovered local geometry and decoded topology, and grouped them by the violated symbolic rules. We reuse the same rule vocabulary that defines \(\mathcal L_{\text{symbolic}}\) in Equation~\ref{eq:symbolic_loss_final} as the analysis interface, forming a closed loop between interpretability and model design. This enables fine-grained diagnosis by distinguishing failures due to non-bonded repulsion, local bonded geometry, topology restoration, or interactions among constraints, rather than treating all invalid generations as a single error class. We visualize randomly selected failure cases for each of the four constraints in Figure~\ref{fig:failure_case}.

The most consequential failure mode is the isolated violation of \(E_{\text{clash}}\). In these samples, the generated structure exhibits clear steric collapse. Non-bonded atoms become too close, local clusters are compressed, and some coordinates even overlap. Specifically, this suggests that the model has not yet learned the basic excluded-volume regularity induced by non-bonded repulsion and van der Waals separation~\cite{bondi1964van}. Such failures typically arise in the early stage, high-noise portion of the reverse process, before the latent state enters a molecule-like geometric manifold. This failure is upstream, since once atomic placement is implausible, later constraints no longer operate on a chemically meaningful configuration. Based on this, our analysis indicates that \(E_{\text{clash}}\) should serve as the first stabilizer of the reverse process rather than one constraint among many, motivating why we activate \(E_{\text{clash}}\) first and assign it the largest weight in Equation~\ref{eq:symbolic_loss_final}.

This upstream role becomes clearer in coupled failure modes involving \(E_{\text{clash}}\). When \(E_{\text{clash}}\) and \(E_{\text{bond\_len}}\) fail together, the typical issue is not that the bond-length prior is inherently weak. Rather, steric collapse has already corrupted the local configuration, pushing bonded distances outside their empirical regime. Therefore, violations of \(E_{\text{bond\_len}}\) are often secondary to violations of \(E_{\text{clash}}\). A similar dependency appears when \(E_{\text{clash}}\) and \(E_{\text{angle}}\) fail together, where abnormal bond angles usually reflect unstable atomic placement rather than a purely angular defect. This is chemically consistent with the fact that realistic bond lengths and angular preferences are well defined only after atoms occupy a physically plausible local neighborhood~\cite{cordero2008covalent, bader1988physical}. The same logic applies to co-failures of \(E_{\text{clash}}\) and \(E_{\text{val}}\): once the geometry collapses, bond inference becomes unreliable and valence assignments naturally break. In summary, many seemingly distinct failures can be traced back to the same upstream geometric defect, motivating our curriculum that establishes spatial non-overlap before introducing higher-order physicochemical rules.

Isolated violations of \(E_{\text{angle}}\), as well as joint violations of \(E_{\text{bond\_len}}\) and \(E_{\text{angle}}\), point to a different issue. Bond angles capture chemically meaningful coordination preferences, but they are intrinsically weaker and more ambiguous than steric exclusion or bond length. An angle only constrains the directional relation among three atoms, does not uniquely determine a 3D structure, and the same angle can correspond to multiple 3D geometries, which is especially problematic in noisy diffusion states. Moreover, \(E_{\text{angle}}\) relies on the bond posterior to decide which atom triplets are relevant, so uncertainty in inferred connectivity can propagate into the angular term itself. As a result, its gradients become informative only after local connectivity is sufficiently reliable. These failure cases showcase that activating angular rules too early or too strongly yields weak guidance or conflicts with more fundamental geometric objectives. This explains why we place \(E_{\text{angle}}\) after \(E_{\text{bond\_len}}\), and keep its weight below those of steric exclusion and bond length in Equation~\ref{eq:symbolic_loss_final}. Physicochemically, angle realism is a vital refinement signal, but it is not an effective first-order guidance for generation.

\begin{table}[t]
\centering
\caption{Subset-wise evaluation on strained and atypical motifs in GEOM-Drugs~\cite{geom-drug}. Sanitization is the fraction of generated molecules that pass RDKit sanitization. Ring BL Err is the mean absolute bond-length error computed only on bonds inside the target motif. Ring BA Err is the mean absolute bond-angle error computed only on angles inside the target motif.}
\vspace{0.2cm}
\label{tab:failure_subset_eval}
\setlength{\tabcolsep}{9pt}
\renewcommand{\arraystretch}{1.25}
\resizebox{0.96\linewidth}{!}{
\begin{tabular}{l l c c c c}
\toprule
Motif & Metric & SensDiff & GEOLDM~\cite{geoldm} & SLDM~\cite{sldm} & UAE-3D~\cite{uae-3d} \\
\midrule
\multirow{4}{*}{Cyclopropane}
& Sanitization (\(\uparrow\))      & 0.95  & 0.89  & 0.88  & 0.91 \\
& Mol Stab (\(\uparrow\))          & 0.90  & 0.81  & 0.80  & 0.83 \\
& Ring BL Err (\(\downarrow\))     & 0.024 & 0.038 & 0.042 & 0.035 \\
& Ring BA Err (deg) (\(\downarrow\)) & 2.8   & 5.2   & 5.8   & 4.5 \\

\midrule
\multirow{4}{*}{Cyclobutane}
& Sanitization (\(\uparrow\))      & 0.92  & 0.84  & 0.83  & 0.87 \\
& Mol Stab (\(\uparrow\))          & 0.87  & 0.76  & 0.74  & 0.79 \\
& Ring BL Err (\(\downarrow\))     & 0.028 & 0.045 & 0.050 & 0.041 \\
& Ring BA Err (deg) (\(\downarrow\)) & 3.5   & 6.8   & 7.2   & 5.5 \\

\midrule
\multirow{4}{*}{\(\beta\)-Lactam}
& Sanitization (\(\uparrow\))      & 0.90  & 0.80  & 0.78  & 0.83 \\
& Mol Stab (\(\uparrow\))          & 0.82  & 0.71  & 0.68  & 0.74 \\
& Ring BL Err (\(\downarrow\))     & 0.030 & 0.052 & 0.058 & 0.048 \\
& Ring BA Err (deg) (\(\downarrow\)) & 4.2   & 7.5   & 8.0   & 6.2 \\

\midrule
\multirow{4}{*}{Bridged ring}
& Sanitization (\(\uparrow\))      & 0.91  & 0.82  & 0.80  & 0.85 \\
& Mol Stab (\(\uparrow\))          & 0.84  & 0.73  & 0.71  & 0.76 \\
& Ring BL Err (\(\downarrow\))     & 0.025 & 0.042 & 0.048 & 0.038 \\
& Ring BA Err (deg) (\(\downarrow\)) & 3.2   & 6.0   & 6.5   & 5.0 \\
\bottomrule
\end{tabular}
}
\end{table}

\begin{table}[t]
\centering
\caption{Generation-frequency fidelity for strained and atypical motifs on GEOM-Drugs~\cite{geom-drug}. GEOM-Drugs Reference is the empirical frequency of each motif in the dataset. The model columns report the corresponding frequency among generated molecules. Valid is the overall validity of generated molecules under the standard validity metric used in the main text.}
\vspace{0.2cm}
\label{tab:failure_frequency_fidelity}
\setlength{\tabcolsep}{9pt}
\renewcommand{\arraystretch}{1.25}
\resizebox{0.96\linewidth}{!}{
\begin{tabular}{l c c c c c}
\toprule
Motif & GEOM-Drugs Reference~\cite{geom-drug} & SensDiff & GEOLDM~\cite{geoldm} & SLDM~\cite{sldm} & UAE-3D~\cite{uae-3d} \\
\midrule
Cyclopropane & 4.8\%  & 4.5\%  & 3.8\%  & 4.1\%  & 4.3\% \\
Cyclobutane  & 2.5\%  & 2.3\%  & 1.9\%  & 2.1\%  & 2.2\% \\
\(\beta\)-Lactam & 0.75\% & 0.68\% & 0.42\% & 0.55\% & 0.60\% \\
Bridged ring & 3.4\%  & 3.1\%  & 2.5\%  & 2.8\%  & 2.9\% \\
Valid         & --     & 93\%   & 84\%   & 83\%   & 87\% \\
\bottomrule
\end{tabular}
}
\end{table}

\begin{table}[t]
\centering
\caption{Inspection of motif-specific geometric statistics in the symbolic rule tables. Here \(\mu\) and \(\sigma\) denote the mean and standard deviation of the corresponding bond-length or bond-angle statistic used by the rule table. The comparison shows that cyclopropane is assigned a distinct geometric prior instead of being matched to a generic acyclic alkane.}
\vspace{0.2cm}
\label{tab:failure_rule_stats}
\setlength{\tabcolsep}{9pt}
\renewcommand{\arraystretch}{1.3}
\resizebox{0.85\linewidth}{!}{
\begin{tabular}{l c c c}
\toprule
Feature & Cyclopropane & Acyclic alkane & Difference \\
\midrule
Mean of C--C bond length (\(\mu\))   & 1.510 \AA   & 1.540 \AA   & $-$0.030 \AA \\
Standard deviation of C--C bond length (\(\sigma\)) & 0.035 \AA   & 0.028 \AA   & -- \\
Mean of C--C--C bond angle (\(\mu\)) & 60.0\(^\circ\)  & 109.5\(^\circ\) & $-$49.5\(^\circ\) \\
Standard deviation of C--C--C bond angle (\(\sigma\)) & 3.2\(^\circ\)   & 4.5\(^\circ\)   & -- \\
\bottomrule
\end{tabular}
}
\end{table}

Isolated violations of \(E_{\text{val}}\) are also informative. In these failure cases, the continuous geometry can appear partially plausible, but the molecule becomes chemically invalid after mapping back to a discrete graph. Some atoms exhibit invalid valence, recovered bond orders are inconsistent, or the resulting topology is illegal. This pattern suggests a failure of geometry–topology alignment rather than pure geometric collapse. Therefore, the model has not yet learned a stable correspondence between local structure and admissible bonding patterns, which is precisely the regime targeted by valence-based reasoning and bond–valence consistency. Crucially, this does not suggest enforcing valence first, since valence validity is downstream of geometry and bond inference. Nevertheless, it does indicate that legality cannot be postponed to a purely terminal repair step. Once the reverse process drifts too far from a topologically admissible regime, post-hoc restoration becomes infeasible. This is why we place \(E_{\text{val}}\) last in the curriculum, yet still incorporate it into both training via Equation~\ref{eq:symbolic_loss_final} and Equation~\ref{eq:total_loss}, rather than treating it as an external validator.

Moreover, the most revealing compound failure is the simultaneous violation of \(E_{\text{clash}}\), \(E_{\text{angle}}\), and \(E_{\text{val}}\). We observed this pattern most often in harder settings, especially when the model was required to satisfy multiple structural objectives at the same. These samples suggest that the issue is not merely weak constraints but cross-constraint interference. Multiple rules activate before the structural hierarchy stabilizes, pulling the sample in competing directions and preventing effective satisfaction of any single rule. This was a key negative result in our preliminary study. It demonstrates that enabling all constraints from the start, or increasing all weights in lockstep, does not improve generation even when each constraint is chemically motivated. The challenge is not only which rules to impose, but when they become physicochemically meaningful during denoising. This observation directly supports the adaptive stage-wise weighting schedule in Equation~\ref{eq:lambda_schedule}.

Taken together, all these failure cases reveal a clear hierarchy of physicochemical dependencies. \(E_{\text{clash}}\) governs whether 3D molecule generation first enters a physically admissible spatial regime. \(E_{\text{bond\_len}}\) and \(E_{\text{angle}}\) then refine local bonded geometry once that regime has stabilized. \(E_{\text{val}}\) finally enforces consistency between continuous structure and chemically legal topology. This is the core logic implemented in our sequential training curriculum. Motivated by this, the order \(E_{\text{clash}} \rightarrow E_{\text{bond\_len}} \rightarrow E_{\text{angle}} \rightarrow E_{\text{val}}\) should be viewed not as a heuristic training trick, but as the outcome of the diagnostic loop. The same idea also holds for the descending weight pattern in Equation~\ref{eq:symbolic_loss_final}. In particular, the most upstream and physically indispensable rule is emphasized most strongly, while more indirect and uncertainty-sensitive rules are introduced later and more gently. This analysis also clarifies why SensDiff leverages exactly these four constraints. Across failure cases, we repeatedly observed four missing pieces of internalized scientific structure: excluded-volume control, bond-length consistency, bond-angle consistency, and valence validity. Each deficit maps naturally to a differentiable symbolic energy that can be incorporated into \(\mathcal L_{\text{symbolic}}\) and ultimately \(\mathcal L_{\text{total}}\) in Equation~\ref{eq:total_loss}. Meanwhile, this study indicates that progress does not come from adding constraints indiscriminately. Overly aggressive or poorly sequential constraint enforcement often increases interference rather than validity. The resulting design principle is thus not more rules, but the right rules, activated when their underlying structural signal becomes reliable.

Empirically, a natural concern is whether symbolic constraints may unintentionally penalize chemically valid but geometrically unusual motifs, especially strained rings and bridged structures whose bond lengths and bond angles differ from standard acyclic patterns. To examine this issue, we perform a failure mode analysis on GEOM-Drugs~\cite{geom-drug} using three studies. First, we conduct a subset-wise evaluation on molecules containing cyclopropane, cyclobutane, \(\beta\)-lactam, and bridged-ring motifs in \Cref{tab:failure_subset_eval}. We report Sanitization, the fraction of generated molecules that pass RDKit sanitization, Mol Stab, the fraction that satisfies the molecular stability criterion used in the main text, Ring BL Err, the mean absolute bond-length error on bonds inside the target ring motif, and Ring BA Err, the mean absolute bond-angle error on angles inside the target ring motif in degrees. Second, we evaluate generation-frequency fidelity in \Cref{tab:failure_frequency_fidelity}, where the GEOM-Drugs Reference column gives the empirical frequency of each motif in the dataset and the model columns report the corresponding frequency among generated molecules. We also report Valid, the overall molecular validity under the same metric used throughout the paper. Third, we inspect the statistical rule tables in \Cref{tab:failure_rule_stats}, where \(\mu\) and \(\sigma\) denote the mean and standard deviation of the corresponding bond-length or bond-angle statistic, to verify that unusual motifs are assigned motif-specific geometric priors rather than generic acyclic ones. The results show that SensDiff does not suppress unusual but chemically valid motifs. In \Cref{tab:failure_subset_eval}, SensDiff achieves higher sanitization and molecular stability than GEOLDM, SLDM, and UAE-3D across all four motif categories, while also obtaining lower ring-specific bond-length and bond-angle errors. Moreover, \Cref{tab:failure_frequency_fidelity} shows that the generation frequency of these motifs remains close to the GEOM-Drugs reference distribution. As shown in \Cref{tab:failure_rule_stats}, this behavior is consistent with the context-dependent symbolic rules, which assign strained motifs distinct geometric priors instead of generic acyclic ones. Overall, the results indicate that the symbolic rules improve common geometric and valence-related failure modes while remaining compatible with strained and atypical molecular motifs.

In summary, this preliminary study highlights a concrete methodological advantage of neuro-symbolic generative modeling. The same symbolic formulation that improves the model also enables human understandable debugging. Rather than reporting only lower validity, we can diagnose whether a model fails to suppress early steric collapse, applies angular supervision before local bonding is reliable, or defers chemical legality until geometry and topology have already drifted apart. Such fine-grained diagnosis is difficult to obtain from purely implicit generative models. In this sense, the failure analysis is not merely an empirical add-on, but evidence that rule-level interpretability can directly drive model improvements. Based on this, the final SensDiff formulation in Equation~\ref{eq:symbolic_loss_final} and Equation~\ref{eq:total_loss}, and Equation~\ref{eq:lambda_schedule} can be viewed as the outcome of this analytical process.

\vspace{2\baselineskip}
\section{Extending the Framework to Diffusion Model Fine-Tuning}
\label{sec:fine-tuning}

\begin{figure*}[t]
    \centering
    \includegraphics[width=1.0\textwidth]{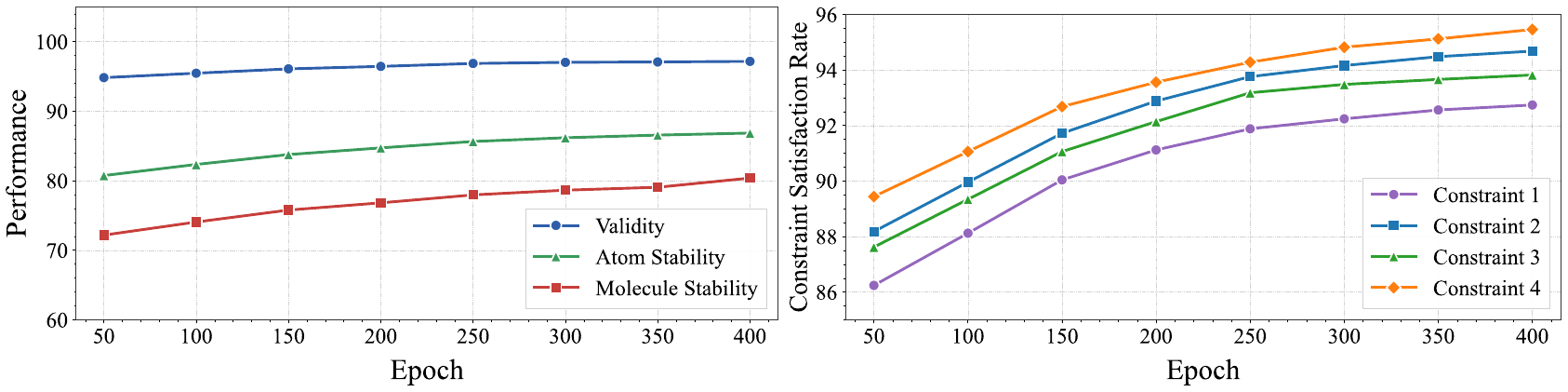}
    \caption{Fine-tuning the diffusion model with symbolic constraints on QM9~\cite{qm9}. (\textbf{Left}) Generation quality over fine-tuning epochs. (\textbf{Right}) Constraint satisfaction rate over fine-tuning epochs.}
    \label{fig:ft_qm9}
    \vspace{0.4cm}
\end{figure*}

\begin{figure*}[t]
    \centering
    \includegraphics[width=1.0\textwidth]{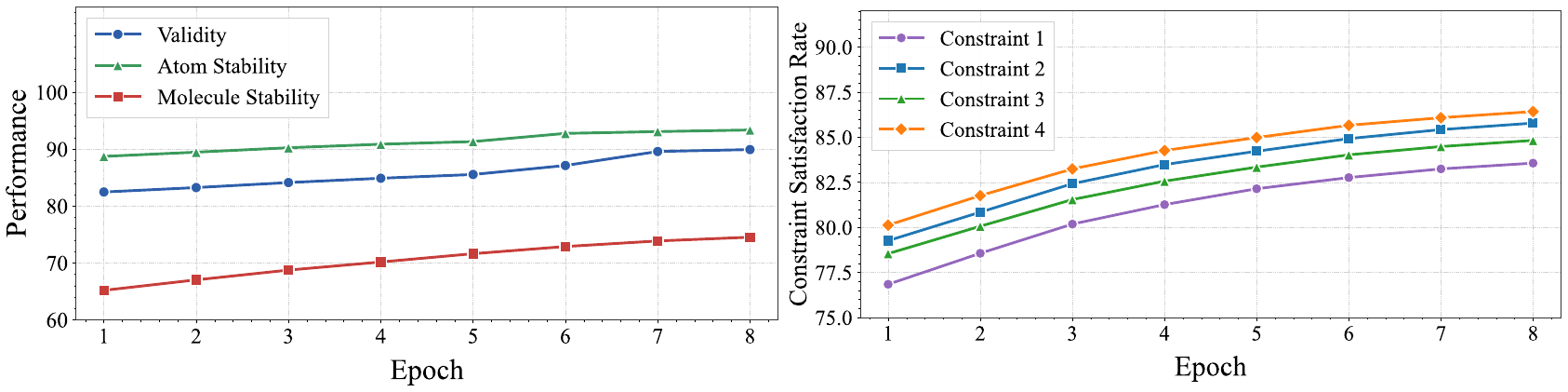}
    \caption{Fine-tuning the diffusion model with symbolic constraints on GEOM-Drugs~\cite{geom-drug}. (\textbf{Left}) Generation quality over fine-tuning epochs. (\textbf{Right}) Constraint satisfaction rate over fine-tuning epochs.}
    \label{fig:ft_gd}
\end{figure*}

\begin{table*}[t]
\centering
\caption{Timestep-wise reliability of intermediate denoising states on QM9~\cite{qm9}. As denoising proceeds, bond confidence increases, bond entropy decreases, and all symbolic constraints become progressively more active. The last column reports the fraction of samples whose effective symbolic weight is approximately zero, showing that very noisy steps are already naturally suppressed under our formulation.}
\vspace{0.1cm}
\label{tab:timestep_reliability}
\renewcommand{\arraystretch}{1.3}
\resizebox{0.9\linewidth}{!}{
\begin{tabular}{c|cc|cccc|c}
\toprule
$t$ & Bond Top-1 Confidence ($\uparrow$) & Bond Entropy ($\downarrow$) & $E_1$ & $E_2$ & $E_3$ & $E_4$ & Fraction of $\lambda \approx 0$ (\%, $\downarrow$) \\
\midrule
900 & 0.32 & 3.62 & 0.25 & 0.09 & 0.04 & 0.02 & 0.92 \\
700 & 0.52 & 2.68 & 0.52 & 0.30 & 0.19 & 0.10 & 0.72 \\
500 & 0.73 & 1.88 & 0.77 & 0.58 & 0.43 & 0.28 & 0.42 \\
300 & 0.87 & 1.28 & 0.92 & 0.82 & 0.68 & 0.53 & 0.16 \\
100 & 0.95 & 0.85 & 0.98 & 0.96 & 0.91 & 0.82 & 0.02 \\
50  & 0.97 & 0.72 & 0.99 & 0.98 & 0.95 & 0.90 & 0.01 \\
\bottomrule
\end{tabular}
}
\end{table*}

\begin{table*}[t]
\centering
\caption{Ablation of supervision schedules on QM9~\cite{qm9}. $V_0$ is the default epoch-only curriculum used in SensDiff. $V_1$ applies timestep-only weighting. $V_2$ and $V_3$ combine epoch-wise and timestep-wise scheduling, using either a shared threshold or per-constraint thresholds. The epoch-only curriculum remains the most effective and stable overall, while explicit timestep-aware strategies provide at most marginal gains over the default formulation.}
\vspace{0.1cm}
\label{tab:epoch_timestep_ablation}
\renewcommand{\arraystretch}{1.4}
\large
\resizebox{\linewidth}{!}{
\begin{tabular}{l|cccc|cccc|cc}
\toprule
Variant & Valid & Atom Stab & Mol Stab & V\&U & $E_1$ & $E_2$ & $E_3$ & $E_4$ & Loss fluctuation ($\downarrow$) & Convergence epoch ($\downarrow$) \\
\midrule
$V_0$ (SensDiff) & 96.9 & 98.1 & 95.6 & 97.2 & 0.965 & 0.958 & 0.949 & 0.957 & $\pm 12\%$ & 2400 \\
$V_1$ (timestep-only) & 93.2 & 94.5 & 91.0 & 94.1 & 0.946 & 0.938 & 0.928 & 0.935 & $\pm 32\%$ & 2800 \\
$V_2$ (joint shared $\tau=T/2$) & 96.4 & 97.6 & 95.2 & 96.8 & 0.961 & 0.954 & 0.945 & 0.953 & $\pm 16\%$ & 2500 \\
$V_3$ (joint per-constraint $\tau$) & 96.7 & 97.9 & 95.4 & 97.0 & 0.963 & 0.957 & 0.947 & 0.956 & $\pm 14\%$ & 2450 \\
\bottomrule
\end{tabular}
}
\end{table*}

In this section, we further extend our training strategy to fine-tune the model using the same symbolic constraints. Figure~\ref{fig:ft_qm9} and Figure~\ref{fig:ft_gd} report performance across fine-tuning epochs on QM9~\cite{qm9} and GEOM-Drugs~\cite{geom-drug}. On both datasets, all metrics improve steadily with fine-tuning, with the largest gains appearing in the early epochs (\textit{e.g.,} on GEOM-Drugs, validity increases from 82.5\% to 89.6\% within only 8 epochs). Notably, constraint compliance improves in tandem with generation quality, demonstrating that symbolic constrained fine-tuning improves structural validity and physicochemical plausibility simultaneously rather than trading one for the other. The rapid convergence further highlights the efficiency of our approach. Only a few fine-tuning epochs are needed to internalize symbolic constraints into a pretrained diffusion model, making it practical for large-scale 3D molecule generation. We attribute this efficiency to the dense supervision provided by symbolic constraints, which complements the data-driven objective by supplying explicit gradient feedback at each denoising step. Moreover, this improvement suggests that these constraints are aligned with the data distribution, acting as a regularizer instead of an adversarial objective. Essentially, neuro-symbolic fine-tuning distills scientific priors into the model parameters, yielding lasting gains with more efficient computation than large-scale pretraining.

\vspace{2\baselineskip}
\section{Additional Studies on Training Schemes}
\label{config}

In this section, we conduct a systematic experimental study of training schemes for improving diffusion models with symbolic supervision. Specifically, we compare epoch-wise and timestep-wise supervision strategies, analyze the reliability of intermediate denoising states across timesteps, and perform dataset-specific ablations on curriculum activation epochs and ramp length to evaluate their effects on generation quality, validity, stability, and constraint satisfaction.

\textbf{Epoch-wise versus timestep-wise supervision.}
An important question is whether symbolic supervision should be scheduled only across epochs or also across diffusion timesteps. Since reverse states at high noise levels are less reliable, timestep-aware supervision might weaken noisy symbolic signals early in denoising and strengthen them later when molecular structure is more formed. We therefore study whether explicit timestep-aware scheduling brings additional benefits beyond the epoch-wise curriculum used in SensDiff. Specifically, we first analyze the reliability of intermediate denoising states across timesteps in \Cref{tab:timestep_reliability}, using bond top-1 confidence, bond entropy, per-constraint activation strength, and the fraction of samples with near-zero effective symbolic weight. We then compare four supervision strategies on QM9 in \Cref{tab:epoch_timestep_ablation}: the default epoch-only curriculum, a timestep-only strategy, and two joint epoch+timestep strategies with either a shared or a per-constraint timestep threshold. In all cases, only the supervision schedule is changed. The results show that the current formulation already has clear implicit timestep adaptivity: early noisy states receive near-zero effective symbolic weights, while later denoising steps activate the constraints much more strongly. As a result, explicit timestep-aware scheduling brings only limited changes and does not improve the overall stability--quality trade-off over the default epoch-wise curriculum.

\textbf{Hyperparameter selection.}
Another practical question is how to choose the activation epochs for the staged symbolic curriculum, as well as the ramp length \(R_{\text{ramp}}\), which denotes the number of training epochs over which the weight of a newly introduced symbolic constraint is linearly increased from zero to its target value after that stage is activated. This design is important because activating constraints too early or too abruptly may inject unstable supervision when the denoising model is still inaccurate, whereas activating them too late may underuse symbolic guidance and limit the final improvement. Moreover, the appropriate schedule is expected to depend on the dataset, since the reliability of intermediate molecular structures and the convergence speed of training differ between QM9~\cite{qm9} and GEOM-Drugs~\cite{geom-drug}. We therefore perform dataset-specific ablations to understand how the curriculum hyperparameters affect generation quality, validity, stability, and symbolic satisfaction. Specifically, we first vary the activation epochs of the three curriculum stages on QM9~\cite{qm9} and GEOM-Drugs~\cite{geom-drug} in \Cref{tab:qm9_stage_epochs,tab:geom_stage_epochs}. This experiment isolates when each group of symbolic constraints should be introduced during training. We then study the joint effect of activation epochs and \(R_{\text{ramp}}\) in \Cref{tab:qm9_stage_ramp,tab:geom_stage_ramp}. This allows us to evaluate not only when new constraints should be activated, but also how smoothly their influence should increase afterward. In all cases, the model architecture, optimizer, and symbolic energies are kept fixed, and only the curriculum hyperparameters are varied. The results indicate that the selected schedules provide a favorable balance between training stability and generation quality. They also show that the optimal activation epochs are dataset-dependent rather than universal.

\section{Limitation and Future Work}
\label{limitation}

While SensDiff performs strongly across benchmarks, several limitations remain. The symbolic constraints in this work are manually specified from established physicochemical knowledge, and automatically learning such rules from data or scientific literature is an important direction for future work. Moreover, extending the framework to more diverse molecular distributions and larger chemical spaces would further test its robustness and scalability. In addition, although the sequential activation curriculum is chemically intuitive, its optimal ordering and schedule may vary across domains, calling for more adaptive curriculum design. We hope SensDiff provides a useful foundation for integrating domain knowledge into generative modeling.

\section{Broader Impacts}
\label{Broader Impacts}

Integrating symbolic domain knowledge into diffusion models can improve the scientific plausibility and structural consistency of generated molecules, which may be valuable for early-stage drug discovery and materials design. Because SensDiff operates at sampling time without retraining, it also lowers the barrier to incorporating updated chemical knowledge into existing pipelines and makes the framework easier to adapt across evolving application settings. However, depending on how rules and thresholds are specified, the constraints may bias generation toward over-regularized, undesirable, or potentially toxic molecules. We hope this work encourages further exploration of neuro-symbolic methods for scientific discovery.

\begin{table}[!t]
\centering
\caption{Ablation of stage activation epochs on QM9~\cite{qm9}.}
\vspace{0.1cm}
\label{tab:qm9_stage_epochs}
\renewcommand{\arraystretch}{1.25}
\resizebox{0.96\linewidth}{!}{
\begin{tabular}{c c c|c c c c|c c c c}
\toprule
Stage~1 & Stage~2 & Stage~3 & Valid & V\&U & Atom Stab & Mol Stab & $E_1$ & $E_2$ & $E_3$ & $E_4$ \\
\midrule
400 & 1120 & 1760 & 91.5 & 92.8 & 92.7 & 89.2 & 93.6 & 92.9 & 89.2 & 92.8 \\
400 & 1680 & 2640 & 92.4 & 94.7 & 94.6 & 91.1 & 94.1 & 93.4 & 91.5 & 90.3 \\
600 & 1120 & 2640 & 92.2 & 93.5 & 91.4 & 90.9 & 95.0 & 91.3 & 92.4 & 94.2 \\
600 & 1680 & 1760 & 93.6 & 93.9 & 93.9 & 92.3 & 96.2 & 92.5 & 91.6 & 93.4 \\
\bottomrule
\end{tabular}
}
\renewcommand{\arraystretch}{1.0}
\vspace{0.1cm}
\end{table}

\begin{table}[!t]
\centering
\caption{Joint ablation of stage activation epochs and ramp length \(R_{\text{ramp}}\) on QM9~\cite{qm9}.}
\vspace{0.1cm}
\label{tab:qm9_stage_ramp}
\renewcommand{\arraystretch}{1.25}
\resizebox{0.96\linewidth}{!}{
\begin{tabular}{c c c|c | c c c c|c c c c}
\toprule
Stage~1 & Stage~2 & Stage~3 & $R_{\text{ramp}}$ & Valid & V\&U & Atom Stab & Mol Stab & $E_1$ & $E_2$ & $E_3$ & $E_4$ \\
\midrule
375 & 1050 & 1650 & 50  & 90.7 & 88.0 & 91.9 & 91.4 & 91.6 & 91.9 & 90.2 & 90.8 \\
375 & 1050 & 2750 & 200 & 89.9 & 87.2 & 92.1 & 90.6 & 91.2 & 92.5 & 89.6 & 90.4 \\
375 & 1750 & 1650 & 100 & 92.6 & 89.9 & 93.8 & 93.3 & 92.6 & 91.9 & 92.0 & 90.8 \\
375 & 1750 & 2750 & 50  & 93.8 & 89.1 & 93.0 & 92.5 & 92.2 & 91.5 & 93.6 & 91.1 \\
625 & 1050 & 1650 & 200 & 88.9 & 87.2 & 93.1 & 89.6 & 92.8 & 88.9 & 89.6 & 90.4 \\
625 & 1750 & 1650 & 100 & 92.1 & 90.5 & 95.4 & 91.8 & 93.9 & 94.2 & 91.8 & 92.1 \\
625 & 1750 & 2750 & 100 & 90.1 & 90.4 & 92.3 & 94.8 & 92.3 & 92.4 & 93.0 & 92.6 \\
625 & 1750 & 2750 & 50  & 91.3 & 89.6 & 93.5 & 93.1 & 94.5 & 91.8 & 87.9 & 91.7 \\
\bottomrule
\end{tabular}
}
\renewcommand{\arraystretch}{1.0}
\vspace{0.1cm}
\end{table}

\begin{table}[!t]
\centering
\caption{Ablation of stage activation epochs on GEOM-Drugs~\cite{geom-drug}.}
\vspace{0.1cm}
\label{tab:geom_stage_epochs}
\renewcommand{\arraystretch}{1.25}
\resizebox{0.96\linewidth}{!}{
\begin{tabular}{c c c|c c c c|c c c c}
\toprule
Stage~1 & Stage~2 & Stage~3 & Valid & V\&U & Atom Stab & Mol Stab & $E_1$ & $E_2$ & $E_3$ & $E_4$ \\
\midrule
8  & 22 & 34 & 90.2 & 89.8 & 79.9 & 71.4 & 89.6 & 86.3 & 83.6 & 85.5 \\
8  & 32 & 50 & 89.7 & 91.3 & 80.4 & 72.9 & 90.8 & 89.2 & 89.8 & 88.4 \\
12 & 22 & 50 & 92.3 & 91.0 & 82.0 & 72.6 & 89.1 & 85.2 & 86.3 & 86.6 \\
12 & 32 & 34 & 93.1 & 91.7 & 80.8 & 73.2 & 92.3 & 90.3 & 86.2 & 87.7 \\
\bottomrule
\end{tabular}
}
\renewcommand{\arraystretch}{1.0}
\vspace{0.1cm}
\end{table}

\begin{table}[!t]
\centering
\caption{Joint ablation of stage activation epochs and ramp length \(R_{\text{ramp}}\) on GEOM-Drugs~\cite{geom-drug}.}
\vspace{0.1cm}
\label{tab:geom_stage_ramp}
\renewcommand{\arraystretch}{1.25}
\resizebox{0.96\linewidth}{!}{
\begin{tabular}{c c c|c | c c c c|c c c c}
\toprule
Stage~1 & Stage~2 & Stage~3 & $R_{\text{ramp}}$ & Valid & V\&U & Atom Stab & Mol Stab & $E_1$ & $E_2$ & $E_3$ & $E_4$ \\
\midrule
8  & 20 & 32 & 5  & 85.9 & 84.5 & 78.6 & 68.1 & 83.7 & 87.8 & 80.0 & 84.3 \\
8  & 20 & 53 & 20 & 84.6 & 83.2 & 77.2 & 65.7 & 85.4 & 82.7 & 81.1 & 82.0 \\
8  & 34 & 32 & 10 & 89.1 & 87.6 & 80.7 & 67.3 & 87.7 & 84.1 & 83.0 & 84.2 \\
8  & 34 & 53 & 5  & 86.7 & 86.3 & 75.4 & 69.9 & 84.9 & 83.5 & 81.3 & 83.6 \\
13 & 20 & 32 & 20 & 84.6 & 83.1 & 76.3 & 64.7 & 88.9 & 82.0 & 83.6 & 85.7 \\
13 & 20 & 53 & 10 & 90.0 & 88.6 & 81.6 & 70.2 & 86.7 & 84.6 & 85.1 & 85.7 \\
13 & 34 & 32 & 10 & 91.0 & 88.5 & 80.7 & 69.0 & 86.0 & 84.3 & 84.5 & 85.1 \\
13 & 34 & 53 & 5  & 87.6 & 87.2 & 79.3 & 67.9 & 85.6 & 84.1 & 83.5 & 84.4 \\
\bottomrule
\end{tabular}
}
\renewcommand{\arraystretch}{1.0}
\vspace{0.3cm}
\end{table}

\end{document}